\documentclass[runningheads]{llncs}
\usepackage[T1]{fontenc}
\usepackage{graphicx}
\usepackage{appendix}
\usepackage{hyperref}
\usepackage{colortbl}
\usepackage{amsmath} 
\usepackage{amssymb} 
\usepackage{cleveref}
\usepackage{float}
\usepackage{subcaption}
\usepackage{caption}
\usepackage[dvipsnames]{xcolor}
\usepackage{fixltx2e} %
\usepackage{multirow}

\begin{document}
\title{MediAug: Exploring Visual Augmentation in Medical Imaging}

\author{Xuyin Qi$^{12*}$, Zeyu Zhang$^{13*\dag}$, Canxuan Gang$^{4*}$, Hao Zhang$^5$, Lei Zhang$^5$, Zhiwei Zhang$^6$, Yang Zhao$^{1\ddag}$\\
~~~~\\
$^{1}$La Trobe~~$^{2}$AIML~~$^{3}$ANU~~$^{4}$UNSW~~$^5$UCAS~~$^6$PSU\\
~~~~\\
\scriptsize$^{*}$Equal contribution. $^{\dag}$Project lead. $^{\ddag}$Corresponding author: y.zhao2@latrobe.edu.au.}

\authorrunning{Xuyin Qi, Zeyu Zhang, and Canxuan Gang et al.}

\institute{}

\maketitle              %

\begin{abstract}
Data augmentation is essential in medical imaging for improving classification accuracy, lesion detection, and organ segmentation under limited data conditions. However, two significant challenges remain. First, a pronounced domain gap between natural photographs and medical images can distort critical disease features. Second, augmentation studies in medical imaging are fragmented and limited to single tasks or architectures, leaving the benefits of advanced mix based strategies unclear. To address these challenges, we propose a unified evaluation framework with six mix based augmentation methods integrated with both convolutional and transformer backbones on brain tumour MRI and eye disease fundus datasets. Our contributions are threefold. (1) We introduce \textbf{MediAug}, a comprehensive and reproducible benchmark for advanced data augmentation in medical imaging. (2) We systematically evaluate MixUp, YOCO, CropMix, CutMix, AugMix, and SnapMix with ResNet‑50 and ViT-B backbones. (3) We demonstrate through extensive experiments that MixUp yields the greatest improvement on the brain tumor classification task for ResNet-50 with \textbf{79.19\%} accuracy and SnapMix yields the greatest improvement for ViT-B with \textbf{99.44\%} accuracy, and that YOCO yields the greatest improvement on the eye disease classification task for ResNet-50 with \textbf{91.60\%} accuracy and CutMix yields the greatest improvement for ViT-B with \textbf{97.94\%} accuracy. Code will be available at \url{https://github.com/AIGeeksGroup/MediAug}.

\end{abstract}

\section{Introduction}
Data augmentation (DA) expands limited datasets with synthetic variations that preserve labels such as flips, rotations and elastic deformations, and it has become indispensable for deep learning \cite{shorten2019survey}. In medical image analysis where scans annotated by experts are scarce, class distributions are skewed and diagnostic cues are subtle, DA directly translates into more reliable screening, triage and treatment planning systems. Two significant challenges remain. First, there is a pronounced domain gap between natural photographs, which motivated most modern DA methods, and medical images whose low contrast, high noise and dense semantic content can make policies designed for ImageNet distort or remove critical disease features, leaving these methods underexplored in clinical AI \cite{ma2022transferability}. Second, DA studies that focus on medical imaging are often fragmented and limited to a single task or model, so it remains unclear which advanced strategies based on mixing images truly improve performance and why \cite{islam2024systematic}.

To address these challenges, we investigate whether the state of the art mix based DA policies, originally developed for natural images, can boost medical classifiers. We introduce \textbf{MediAug}, a unified pipeline that applies six prominent techniques, including MixUp \cite{zhang2017mixup}, YOCO \cite{han2022you}, CropMix \cite{han2022cropmix}, CutMix \cite{yun2019cutmix}, AugMix \cite{hendrycks2019augmix} and SnapMix \cite{huang2021snapmix}, to two clinically relevant datasets, namely the brain tumour MRI dataset \cite{sartaj_bhuvaji_ankita_kadam_prajakta_bhumkar_sameer_dedge_swati_kanchan_2020} and the eye disease fundus dataset \cite{eye_diseases_classification_dataset}. We evaluate both ResNet-50 \cite{he2016deep} and ViT-B \cite{dosovitskiy2020image} backbones by conducting comprehensive experiments that reveal how augmentation and architecture choices affect performance.

\begin{itemize}
  \item We introduce \textbf{MediAug}, a comprehensive and reproducible study of advanced DA strategies for medical imaging that offers a unified reference and practical guidance for the research community.
  \item Our study provides a systematic evaluation of MixUp, CutMix, SnapMix, AugMix, CropMix and YOCO across convolutional and transformer backbones, highlighting their respective strengths and limitations.
  \item We conducted comprehensive experiments on the brain tumour and eye disease datasets, achieving \textbf{79.19\%} with MixUp on ResNet-50 and \textbf{99.44\%} with SnapMix on ViT-B for brain tumour classification and \textbf{91.60\%} with YOCO on ResNet-50 and \textbf{97.94\%} with CutMix on ViT-B for eye disease classification, indicating that MixUp on ResNet-50 and SnapMix on ViT-B are optimal for brain tumours and YOCO on ResNet-50 and CutMix on ViT-B are optimal for eye diseases.
\end{itemize}

\section{Related Work}

Visual data augmentation has long been a cornerstone in computer vision, in natural images, random crops \cite{krizhevsky2012imagenet} mitigate overfitting, flips \cite{perez2017effectiveness} improve generalization, rotations \cite{shorten2019survey} increase feature robustness, and color jitter \cite{shorten2019survey} expands appearance diversity. Noise injection and elastic deformations \cite{perez2017effectiveness} further enhance resilience to imaging artifacts. In medical imaging, DA enhances image classification \cite{luo2025pathohr,qi2025medconv,wu2024xlip,zhang2024jointvit}, object detection \cite{zhao2025peddet,zhang2024meddet,cai2024medical,cai2024msdet,zhao2024landmark}, and semantic segmentation \cite{zhu2025doei,zhang2025gamed,tan2024segkan,ge2024esa,tan2024segstitch,zhang2024segreg,wu2023bhsd,zhang2023thinthick} performance, where expert annotated scans are scarce and manual labeling costs are high, these techniques reduce reliance on large annotated datasets \cite{litjens2017survey}, improve the accuracy of automated screening and triage systems \cite{tajbakhsh2016convolutional}, support semi supervised and transfer learning workflows \cite{cheplygina2019not} when annotations are limited, assist clinicians by lowering diagnostic costs \cite{esteva2017dermatologist,zhang2024deep}, accelerate clinical throughput \cite{rajpurkar2017chexnet,hiwase2025can}, and enable more consistent segmentation of lesions and organs in complex modalities such as MRI and CT \cite{litjens2017survey,qi2025projectedex}. Beyond these traditional methods, advanced mix based approaches generate richer training samples by semantically combining multiple images, for example MixUp blends pairs of images and their labels to smooth decision boundaries \cite{zhang2017mixup}, YOCO applies independent augmentations to subregions to enhance local and global diversity \cite{han2022you}, CropMix merges crops at multiple scales to capture multi resolution features \cite{han2022cropmix}, CutMix replaces patches between images to preserve spatial context \cite{yun2019cutmix}, AugMix ensembles diverse augmentation chains to improve model calibration and uncertainty estimation \cite{hendrycks2019augmix}, and SnapMix leverages class activation maps to guide semantic mixing and improve fine grained classification \cite{huang2021snapmix}. These strategies have demonstrated significant performance gains on natural image benchmarks and offer promising directions for robust medical image analysis.

\section{Method}

\subsection{Overview}

\begin{figure}[t]
\centering
\includegraphics[width=\textwidth]{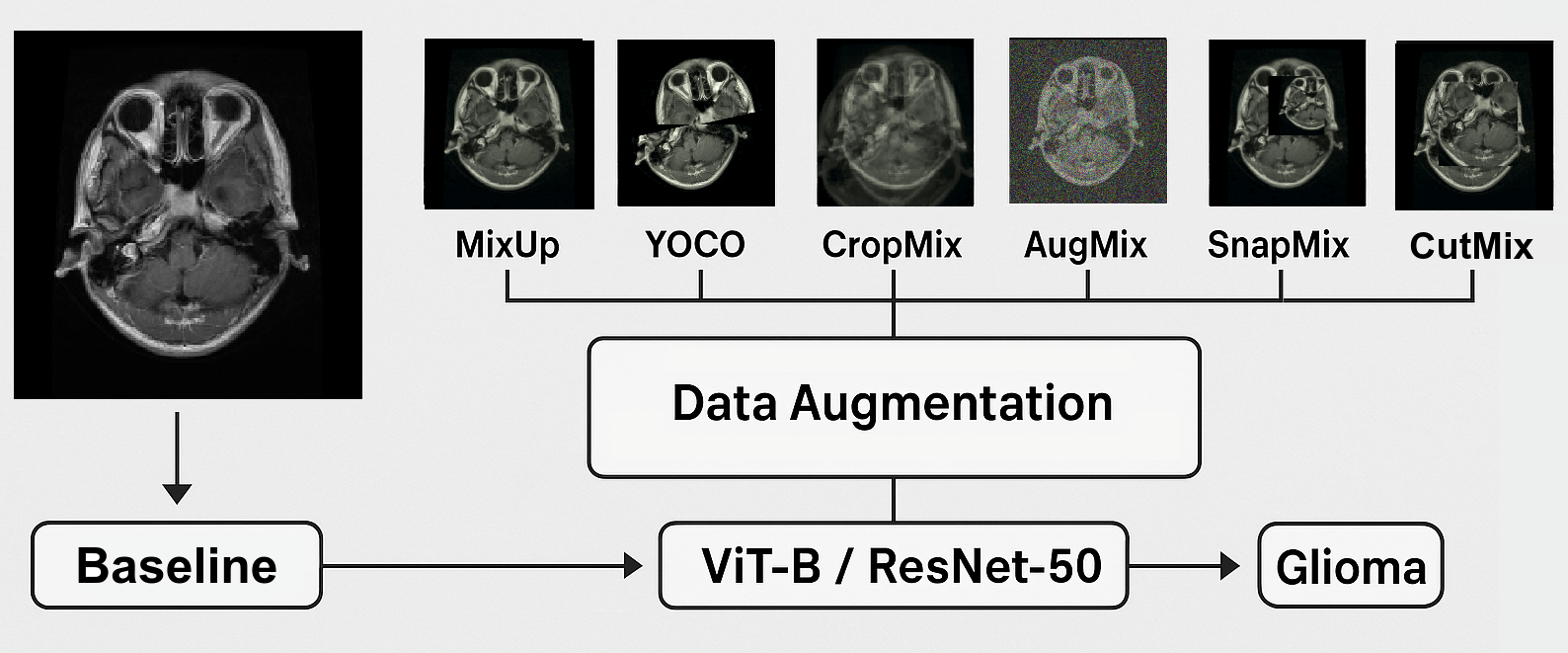}
\caption{\textit{Architecture of MediAug}: We enhance medical representation learning via advanced visual augmentation.}
\label{fig:arch}
\end{figure}

Our method enhances medical representation learning by applying advanced visual data augmentation strategies, as illustrated in \cref{fig:arch}. To begin with, we create augmented versions of each input image with six popular techniques: MixUp \cite{zhang2017mixup}, YOCO \cite{han2022you}, CropMix \cite{han2022cropmix}, CutMix \cite{yun2019cutmix}, AugMix \cite{hendrycks2019augmix}, and SnapMix \cite{huang2021snapmix}, together with the original image as a baseline. These images are then processed by two backbone networks: Vit-B \cite{dosovitskiy2020image} pretrained with JointViT \cite{zhang2024jointvit} and ResNet-50 \cite{he2016deep} pretrained with MedConv \cite{qi2025medconv}. A classification head outputs disease labels, enabling the backbones to learn robust and transferable medical representations from the augmented data.

\begin{figure}[t]
\centering
\includegraphics[width=\textwidth]{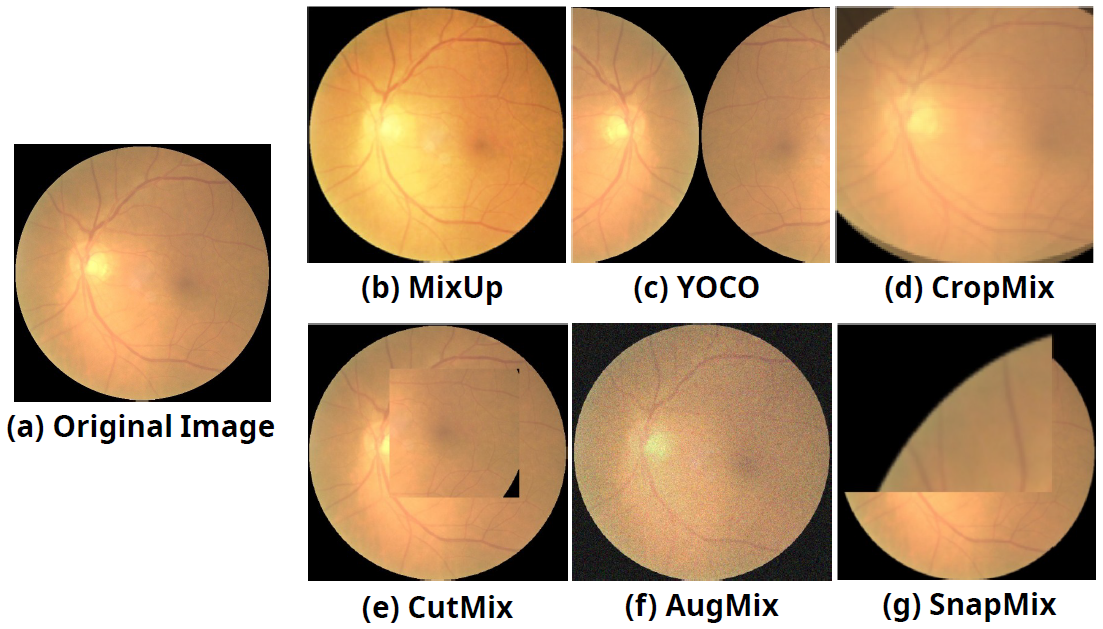}
\caption{\textit{Augmentation in Eye diseases classification dataset} \cite{eye_diseases_classification_dataset}: (a) An image of a cataract. (b) \textbf{\textit{MixUp}} \cite{zhang2017mixup}: Original image and contrast-enhanced image is mixed with a mixing parameter $\lambda = 0.42$. (c) \textbf{\textit{YOCO}} \cite{han2022you}: splits, flips, and recombines the image for augmentation. (d) \textbf{\textit{CropMix}} \cite{han2022cropmix}: combines three 25\% cropped views using MixUp \cite{zhang2017mixup} for augmentation. (e) \textbf{\textit{CutMix}} \cite{yun2019cutmix}: augments by relocating a 1/4 cropped region within the image. (f) \textbf{\textit{AugMix}} \cite{hendrycks2019augmix}: AugMix blends blurring, sharpening, and color adjustments with the original. (g) \textbf{\textit{SnapMix}} \cite{huang2021snapmix}: SnapMix blends interpolated regions, weighted by saliency maps, with the original.}
\label{fig:eye}
\end{figure}

\subsection{MixUp}
\noindent Mixup \cite{zhang2017mixup} is a data augmentation technique that improves the generalization ability of neural networks by interpolating images and labels. Let \( I_a \) and \( I_b \) represent two randomly selected input images from the training dataset, and \( y_a \) and \( y_b \) be their corresponding one-hot encoded labels. A mixing coefficient \( \lambda \) is sampled from a Beta distribution, denoted as \( \lambda \sim \text{Beta}(\alpha, \alpha) \), where \( \alpha > 0 \) controls the strength of interpolation. The mixed image \( \tilde{I} \) and label \( \tilde{y} \) are computed as:

\[
\tilde{I} = \lambda I_a + (1 - \lambda) I_b, \quad \tilde{y} = \lambda y_a + (1 - \lambda) y_b.
\]

\noindent\( I_a \) and \( I_b \) are two randomly selected input images, while \( y_a \) and \( y_b \) are their corresponding one-hot encoded labels. The parameter \( \lambda \) represents the mixing coefficient, sampled from a Beta distribution with parameter \( \alpha > 0 \), which determines the interpolation ratio between \( I_a \) and \( I_b \). The interpolated image \( \tilde{I} \) and label \( \tilde{y} \) are obtained by combining \( I_a \) and \( I_b \), as well as \( y_a \) and \( y_b \), respectively, using \( \lambda \) and \( 1 - \lambda \) as weights.

\noindent During training, the loss function is defined as:
\[
\mathcal{L} = \frac{1}{n} \sum_{i=1}^n \ell(f(\tilde{I}_i), \tilde{y}_i),
\]
\noindent where \( f \) denotes the neural network model, \( \ell \) represents the loss function (e.g., cross-entropy), \( n \) is the batch size, and \( (\tilde{I}_i, \tilde{y}_i) \) are the mixed image and label for the \( i \)-th training example.

\begin{figure}[t]
\centering
\includegraphics[width=\textwidth]{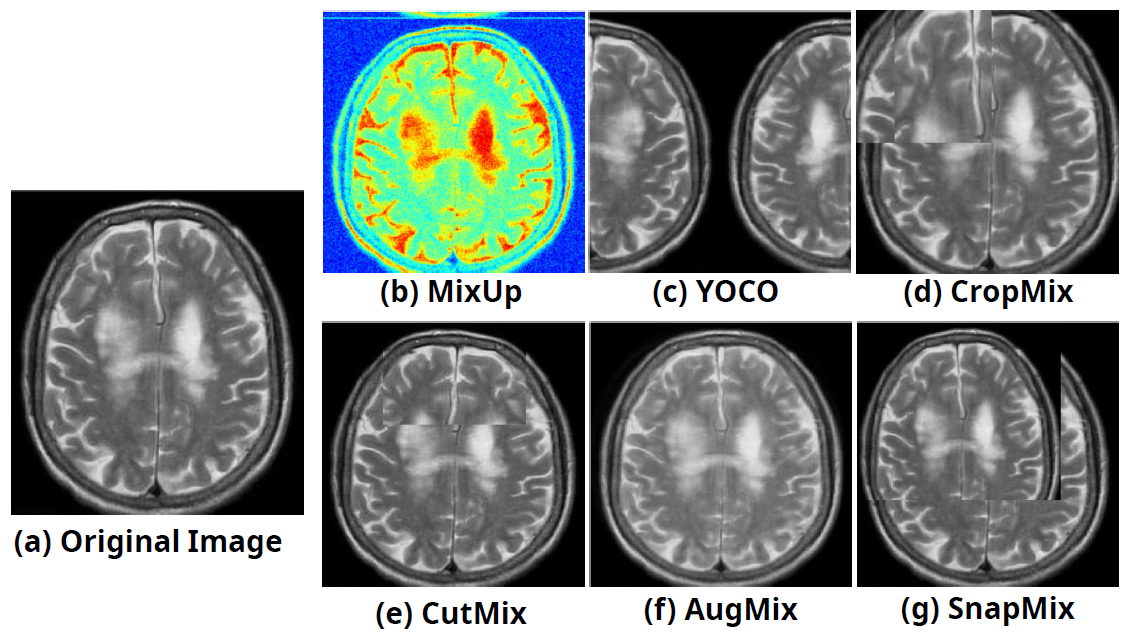}
\caption{\textit{Augmentation in Brain Tumor Classification (MRI) dataset} \cite{sartaj_bhuvaji_ankita_kadam_prajakta_bhumkar_sameer_dedge_swati_kanchan_2020}: (a) An image of a no\_tumor in Testing. (b) \textbf{\textit{MixUp}} \cite{zhang2017mixup}: Contrast-adjusted and mixed ($\lambda=0.30$), in grayscale and pseudo-color. (c) \textbf{\textit{YOCO}} \cite{han2022you}: The image was resized, contrast-enhanced, flipped, and concatenated. (d) \textbf{\textit{CropMix}} \cite{han2022cropmix}: It from a contrast-enhanced, randomly cropped image using CutMix. (e) \textbf{\textit{CutMix}} \cite{yun2019cutmix}: Original image flipped and mixed using random CutMix enhancement ($\lambda=0.70$). (f) \textbf{\textit{AugMix}} \cite{hendrycks2019augmix}: It enhances the image with flipping and brightness adjustment. (g) \textbf{\textit{SnapMix}} \cite{huang2021snapmix}: It enhances by blending resized cropped regions with the original image.}
\label{fig_brain}
\end{figure}

\noindent In medical imaging, Mixup, as illustrated in panels (b) of \cref{fig:eye,fig_brain}, is particularly effective in addressing challenges such as limited labeled data and class imbalance. By interpolating images and labels, Mixup enhances the dataset while maintaining semantic consistency. For example, in lesion classification tasks, Mixup blends regions of interest across different images, helping the model focus on localized features. This technique alleviates overfitting, improves robustness to label noise, and reduces sensitivity to adversarial examples, making it highly valuable for medical image analysis.

\subsection{YOCO}
\noindent The YOCO method \cite{han2022you}, as illustrated in panels (c) of \cref{fig:eye,fig_brain}, is designed for processing medical images \( X \in \mathbb{R}^{C \times H \times W} \), where \( C \) represents the number of image channels (e.g., grayscale or RGB), \( H \) is the height, and \( W \) is the width of the image. The goal of YOCO is to apply data augmentation in a way that enhances both local and global diversity while maintaining the structural integrity of the image. A data augmentation function \( a(\cdot): \mathbb{R}^{C \times H \times W} \to \mathbb{R}^{C \times H \times W} \) transforms the input image \( X \) into an augmented image \( X' = a(X) \).

\noindent Unlike traditional augmentation methods, YOCO first splits the input image into sub-regions. To achieve this, the image is randomly split into two parts along either the height or width dimension. Specifically, if the random variable \( p \sim U(0, 1) \) (a uniformly distributed random number between 0 and 1) satisfies \( 0 < p \leq 0.5 \), the image is split along the height, resulting in two sub-images \( [X_1, X_2] = \text{cutH}(X) \). Alternatively, if \( 0.5 < p \leq 1 \), the image is split along the width, resulting in \( [X_1, X_2] = \text{cutW}(X) \). The dimensions of these sub-images depend on the splitting direction:
- For height splitting, \( X_1, X_2 \in \mathbb{R}^{C \times \frac{H}{2} \times W} \),
- For width splitting, \( X_1, X_2 \in \mathbb{R}^{C \times H \times \frac{W}{2}} \).

\noindent After splitting, independent augmentation functions \( a_1(\cdot) \) and \( a_2(\cdot) \) are applied to the two sub-images \( X_1 \) and \( X_2 \), respectively. These augmentations may involve operations such as rotation, flipping, or intensity adjustment. The final augmented image \( X' \) is then reconstructed by concatenating the two augmented sub-images along the original splitting axis:
\[
X' = \text{concat}[a_1(X_1), a_2(X_2)].
\]

\noindent To generalize this approach for more complex scenarios, YOCO can split the image into \( M+1 \) parts along the height and \( N+1 \) parts along the width, producing \( (M+1) \times (N+1) \) sub-images. This is represented as:
\[
[X_{i,j}] = \text{cut}^{M,N}(X), \quad i=1, \dots, M+1, \ j=1, \dots, N+1,
\]
\noindent where \( X_{i,j} \) denotes the sub-image located in the \( i \)-th row and \( j \)-th column of the grid. Each sub-image \( X_{i,j} \) is independently augmented using a corresponding augmentation function \( a_{i,j}(\cdot) \). The final augmented image is then reconstructed by concatenating all augmented sub-images:
\[
X' = \text{concat}_{i=1,j=1}^{M+1,N+1} [a_{i,j}(X_{i,j})].
\]

\noindent By augmenting sub-regions while preserving global structure, YOCO boosts data diversity, aiding detection of localized features in medical imaging and improving model robustness.

\begin{figure}[t]
    \centering
    \begin{minipage}{0.48\textwidth}
        \centering
        \includegraphics[width=\linewidth]{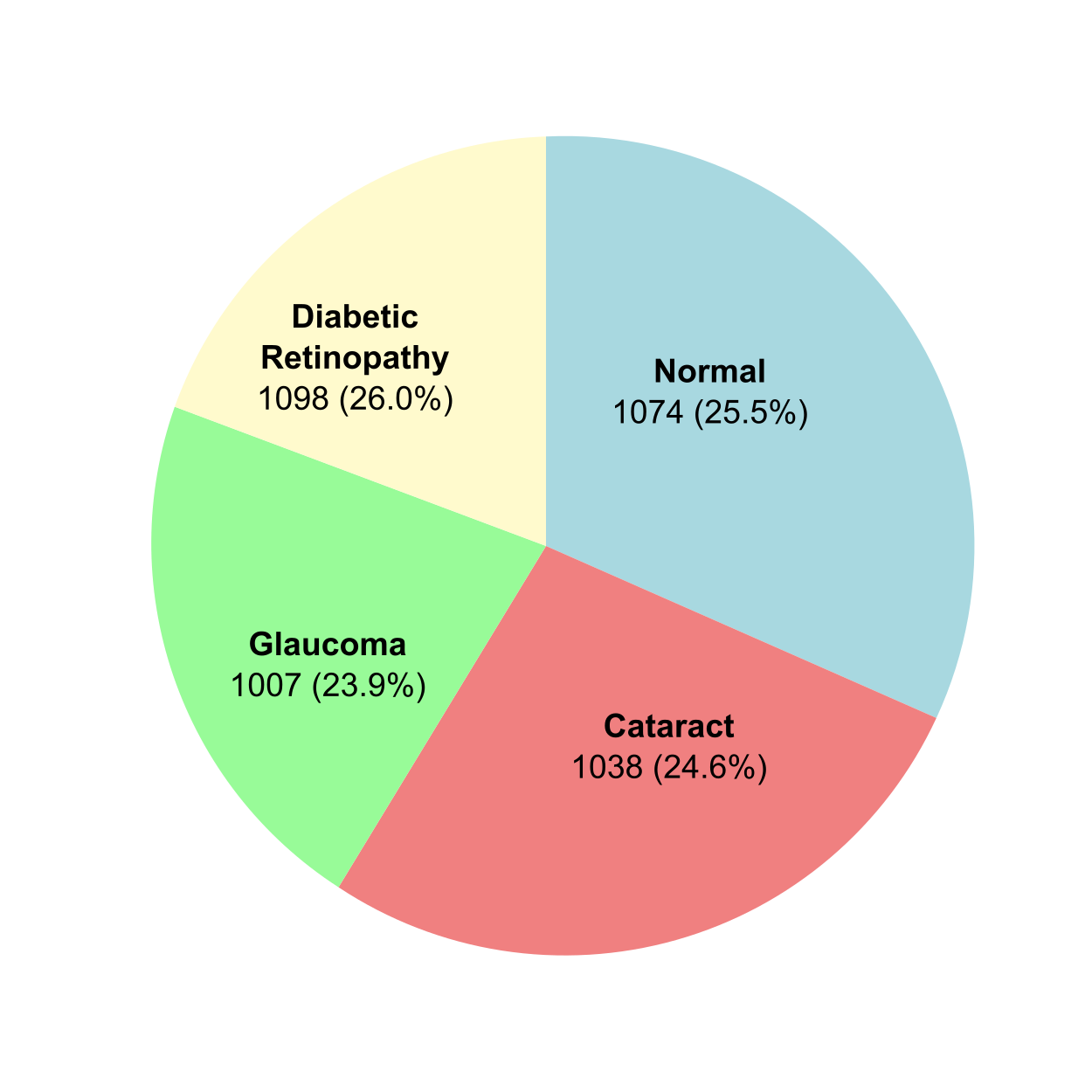}
        \caption{\small Eye disease classification dataset \cite{eye_diseases_classification_dataset} with four categories (23.9\%-26.0\%).}
        \label{fig:eye-disease-pie}
    \end{minipage}
    \hfill
    \begin{minipage}{0.48\textwidth}
        \centering
        \includegraphics[width=\linewidth]{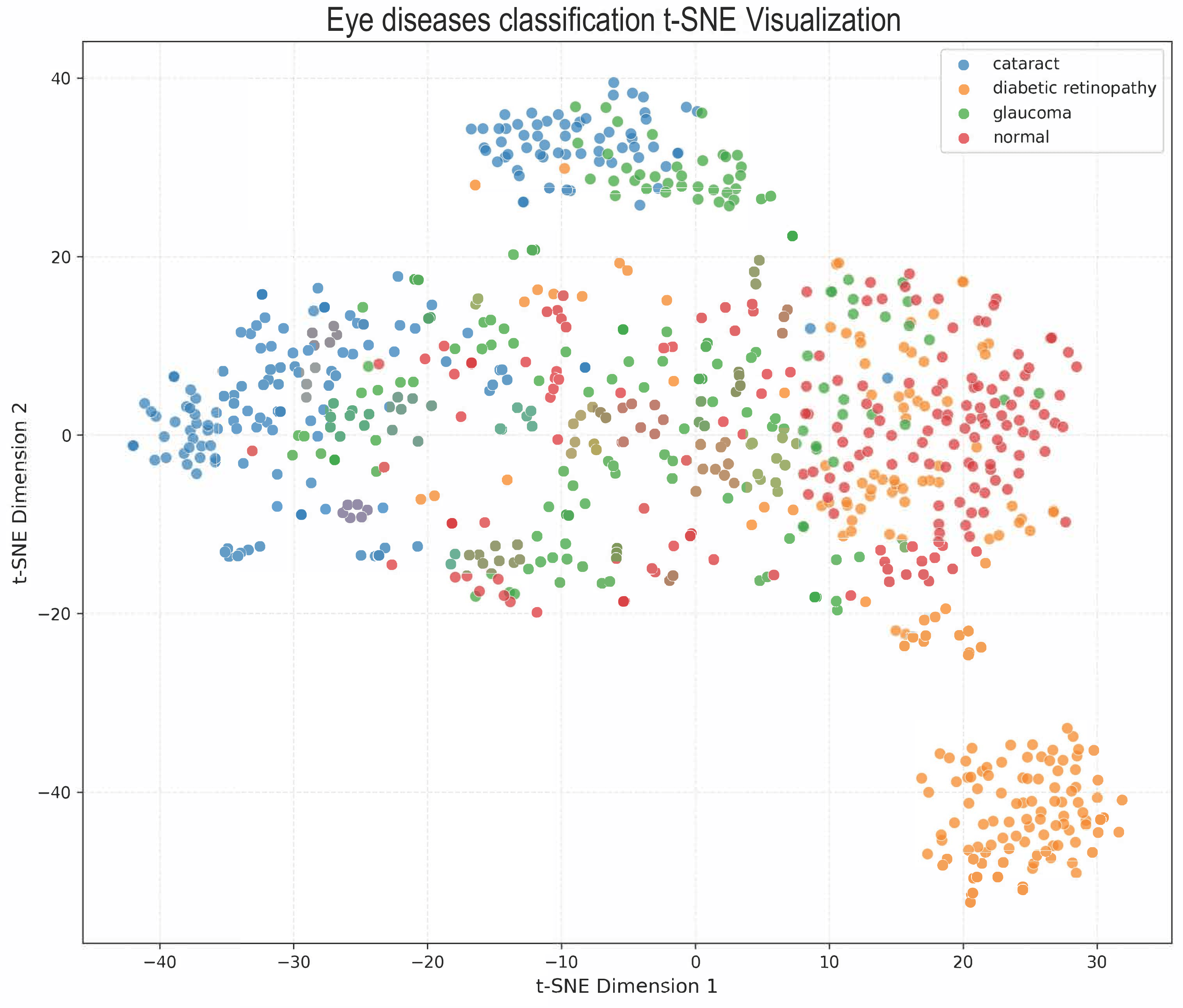}
        \caption{\small t-SNE visualization of eye diseases \cite{eye_diseases_classification_dataset}, showing clustering patterns across cataract, diabetic retinopathy, glaucoma, and normal cases.}
        \label{fig:eye-disease-tsne}
    \end{minipage}
\end{figure}

\subsection{CropMix}
\noindent CropMix \cite{han2022cropmix} is a data augmentation method that improves the generalization ability of neural networks by combining multi-scale random crops to capture diverse features and suppress labeling errors. In medical imaging, this method is particularly effective in addressing challenges such as limited labeled data and complex lesion structures. By generating multiple cropped views of the original image, CropMix enhances data diversity while preserving semantic consistency.

\noindent Let \( I \) represent the original input medical image, and \( I_i \) denote a cropped view of \( I \), obtained through random resized cropping (RRC) with crop scale \( s_i \), where \( s_i \in [s_{\text{min}}, s_{\text{max}}] \). The cropped views \( I_i \) are generated to capture both fine-grained and coarse-grained information from the image. To combine these views, a mixing coefficient \( \lambda \) is sampled from a Beta distribution, denoted as \( \lambda \sim \text{Beta}(\alpha, \alpha) \), where \( \alpha > 0 \) controls the interpolation strength. The mixed image \( \tilde{I} \) is computed using the following formula:

\[
\tilde{I} = \lambda I_1 + (1 - \lambda) I_2,
\]

\noindent where \( I_1 \) and \( I_2 \) are two randomly selected cropped views of \( I \), and \( \lambda \) determines the mixing ratio. Since all cropped views are derived from the same image, the label \( y \) of the original image \( I \) remains unchanged, ensuring semantic consistency:

\[
\tilde{y} = y.
\]

\noindent In medical imaging applications, CropMix, as illustrated in panels (d) of \cref{fig:eye,fig_brain}, leverages the multi-scale information captured from the cropped views to enhance the model's ability to learn localized lesion features while retaining global anatomical structures. For example, in lesion classification tasks, the combination of fine-grained and coarse-grained information helps the model focus on both detailed lesion characteristics and broader contextual cues. By interpolating cropped views and preserving labels, CropMix effectively reduces sensitivity to label noise, alleviates overfitting, and improves robustness, making it highly valuable for medical image analysis.

\begin{figure}[t]
    \centering
    \begin{minipage}{0.48\textwidth}
        \centering
        \includegraphics[width=\linewidth]{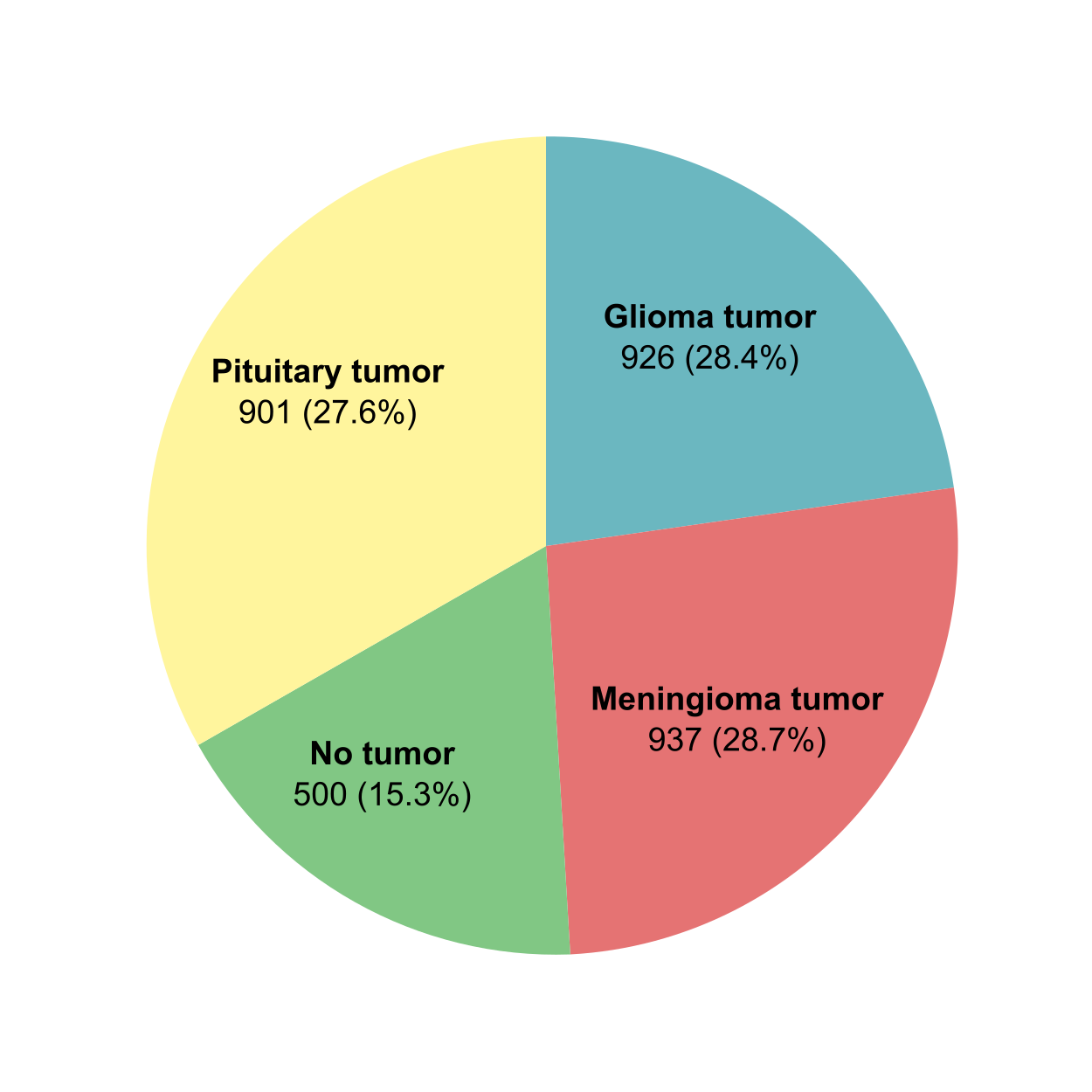}
        \caption{\small Brain MRI dataset \cite{sartaj_bhuvaji_ankita_kadam_prajakta_bhumkar_sameer_dedge_swati_kanchan_2020} showing Meningioma tumor 28.7\%, Glioma 28.4\%, Pituitary 27.6\%, No tumor 15.3\%.}
        \label{fig:brain-tumor-pie}
    \end{minipage}
    \hfill
    \begin{minipage}{0.48\textwidth}
        \centering
        \includegraphics[width=\linewidth]{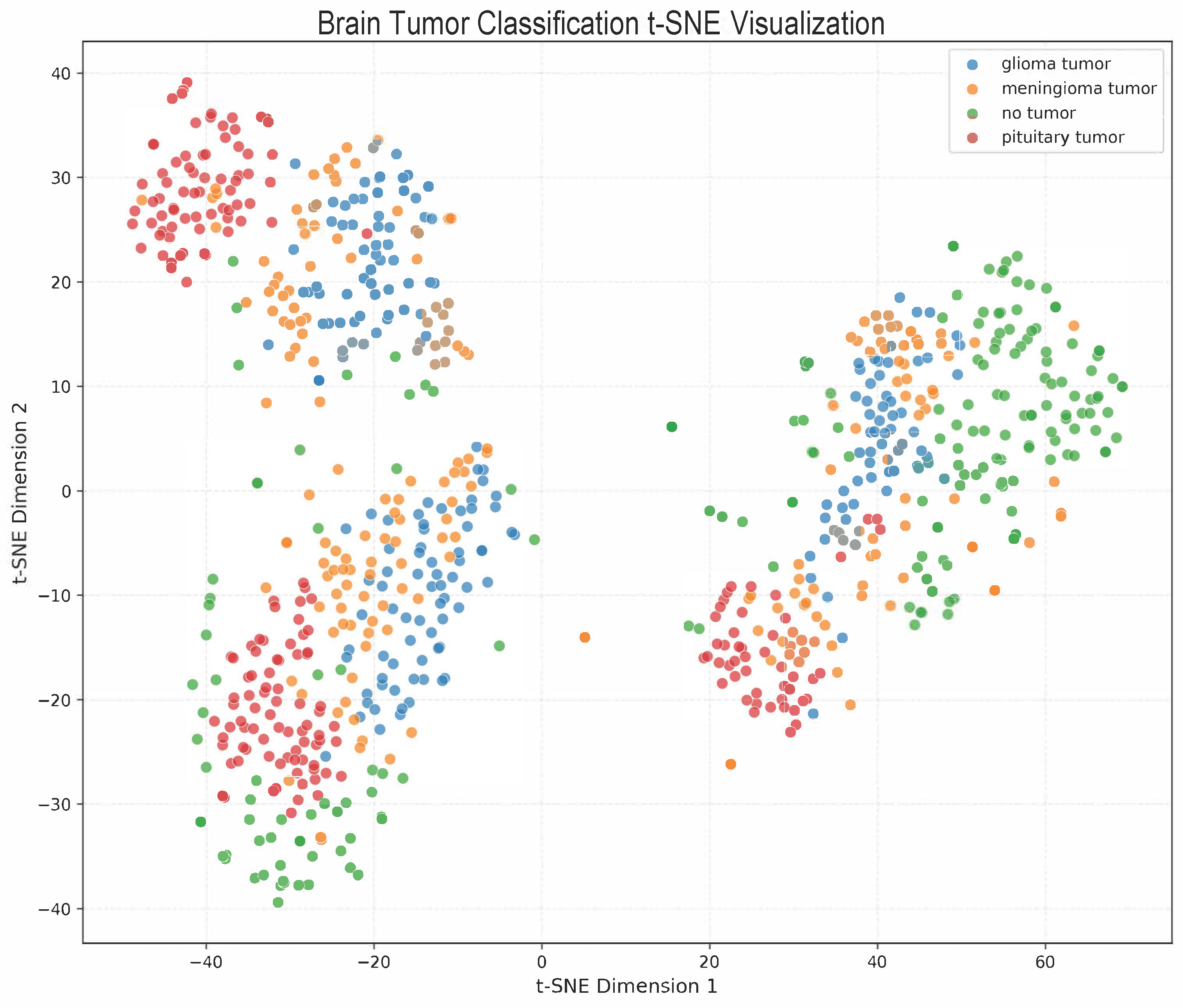}
        \caption{\small t-SNE visualization of brain tumor categories \cite{sartaj_bhuvaji_ankita_kadam_prajakta_bhumkar_sameer_dedge_swati_kanchan_2020} showing clustering of glioma, meningioma, pituitary, and non-tumor cases.}
        \label{fig:brain-tumor-tsne}
    \end{minipage}
\end{figure}

\subsection{CutMix}
\noindent CutMix \cite{yun2019cutmix} is a data augmentation technique that improves the generalization ability of neural networks by combining regions from different images and interpolating their labels. Let \( I_A \) and \( I_B \) represent two randomly selected input images from the training dataset, and \( y_A \) and \( y_B \) be their corresponding one-hot encoded labels. A binary mask \( M \) is used to define the region in \( I_A \) that is replaced with a patch from \( I_B \). The mixing ratio \( \lambda \) is sampled from a Beta distribution, denoted as \( \lambda \sim \text{Beta}(\alpha, \alpha) \), where \( \alpha > 0 \) controls the interpolation strength. The mixed image \( \tilde{I} \) and label \( \tilde{y} \) are computed as:

\[
\tilde{I} = M \odot I_A + (1 - M) \odot I_B, \quad \tilde{y} = \lambda y_A + (1 - \lambda) y_B.
\]

\noindent\( I_A \) and \( I_B \) are two randomly selected input images, while \( y_A \) and \( y_B \) are their corresponding one-hot encoded labels. The binary mask \( M \) defines the region in \( I_A \) that is replaced with pixels from \( I_B \), and \( \lambda \) represents the mixing ratio, which is proportional to the area of \( M \). The mixed image \( \tilde{I} \) is obtained by combining \( I_A \) and \( I_B \) based on \( M \), while the mixed label \( \tilde{y} \) interpolates \( y_A \) and \( y_B \) using \( \lambda \) and \( 1 - \lambda \) as weights.

\noindent The binary mask \( M \) is determined by sampling the bounding box coordinates \( r_x \) and \( r_y \), and dimensions \( r_w \) and \( r_h \). The coordinates \( r_x \) and \( r_y \) are sampled uniformly:
\[
r_x \sim \text{Unif}(0, W), \quad r_y \sim \text{Unif}(0, H),
\]
\noindent where \( W \) and \( H \) are the width and height of the image. The dimensions \( r_w \) and \( r_h \) are computed as:
\[
r_w = W \sqrt{1 - \lambda}, \quad r_h = H \sqrt{1 - \lambda}.
\]

\noindent During training, the loss function is defined as:
\[
\mathcal{L} = \frac{1}{n} \sum_{i=1}^n \ell(f(\tilde{I}_i), \tilde{y}_i),
\]
\noindent where \( f \) denotes the neural network model, \( \ell \) represents the loss function (e.g., cross-entropy), \( n \) is the batch size, and \( (\tilde{I}_i, \tilde{y}_i) \) are the mixed image and label for the \( i \)-th training example.

\noindent In medical imaging, CutMix, as illustrated in panels (e) of \cref{fig:eye,fig_brain}, is particularly effective in addressing challenges such as limited labeled data and class imbalance. By replacing regions between images and interpolating labels, CutMix enhances the dataset while maintaining spatial and semantic consistency. For example, in lesion detection tasks, CutMix combines regions of interest across different images, helping the model focus on localized features. This technique alleviates overfitting, improves robustness to label noise, and reduces sensitivity to adversarial examples, making it highly valuable for medical image analysis.

\begin{table}[t]
    \centering
    \caption{Comparative performance on \textbf{brain tumor classification dataset} \cite{sartaj_bhuvaji_ankita_kadam_prajakta_bhumkar_sameer_dedge_swati_kanchan_2020}.}
    \resizebox{\linewidth}{!}{%
    \begin{tabular}{l|l|c|c|c|c|c|c|c}
        \hline
        \textbf{Method} & \textbf{Backbone} & \textbf{Accuracy} & \textbf{Precision} & \textbf{Recall} & \textbf{Sensitivity} & \textbf{Specificity} & \textbf{F1 Score} & \textbf{ROC AUC} \\
        \hline
        \multirow{2}{*}{Baseline}  & ResNet-50 & 76.4 & 83.18 & 76.14 & 75.26 & 91.69 & 72.22 & 92.34 \\
                                   & ViT-B    & 85.20 & 84.78 & 82.58 & 81.55 & 93.94 & 82.55 & 96.61 \\
        \hline
        \multirow{2}{*}{AugMix}    & ResNet-50 & 76.65\textsubscript{\textcolor{Green}{+0.25}} & 82.35\textsubscript{\textcolor{Red}{-0.83}} & 76.65\textsubscript{\textcolor{Green}{+0.51}} & 75.94\textsubscript{\textcolor{Green}{+0.68}} & 91.92\textsubscript{\textcolor{Green}{+0.23}} & 73.72\textsubscript{\textcolor{Green}{+1.5}} & 94.11\textsubscript{\textcolor{Green}{+1.77}} \\
                                   & ViT-B    & 97.51\textsubscript{\textcolor{Green}{+12.31}} & 84.63\textsubscript{\textcolor{red}{-0.15}} & 84.32\textsubscript{\textcolor{Green}{+1.74}} & 83.39\textsubscript{\textcolor{Green}{+1.84}} & 94.53\textsubscript{\textcolor{Green}{+0.59}} & 84.32\textsubscript{\textcolor{Green}{+1.77}} & 95.39\textsubscript{\textcolor{red}{-1.22}} \\
        \hline
        \multirow{2}{*}{CropMix}   & ResNet-50 & 73.35\textsubscript{\textcolor{Red}{-3.05}} & 78.58\textsubscript{\textcolor{Red}{-4.60}} & 73.35\textsubscript{\textcolor{Red}{-2.79}} & 71.20\textsubscript{\textcolor{Red}{-4.06}} & 90.78\textsubscript{\textcolor{Red}{-0.91}} & 
        70.85\textsubscript{\textcolor{Red}{-1.37}} &
        91.47\textsubscript{\textcolor{Red}{-0.87}}   \\
                                   & ViT-B    & 99.05\textsubscript{\textcolor{Green}{+13.85}} & 86.10\textsubscript{\textcolor{Green}{+1.32}} & 86.06\textsubscript{\textcolor{Green}{+3.48}} & 85.40\textsubscript{\textcolor{Green}{+3.85}} & 95.16\textsubscript{\textcolor{Green}{+1.22}} & 86.00\textsubscript{\textcolor{Green}{+3.45}} & 96.80\textsubscript{\textcolor{Green}{+0.19}} \\
        \hline
        \multirow{2}{*}{CutMix}    
        & ResNet-50 & 74.37\textsubscript{\textcolor{red}{-2.03}} & 81.82\textsubscript{\textcolor{red}{-1.36}} & 74.37\textsubscript{\textcolor{red}{-1.77}} & 72.63\textsubscript{\textcolor{red}{-2.63}} & 91.04\textsubscript{\textcolor{red}{-0.65}} & 72.73\textsubscript{\textcolor{Green}{+0.51}} & 91.87\textsubscript{\textcolor{red}{-0.47}} \\
        & ViT-B    & 97.61\textsubscript{\textcolor{Green}{+12.41}} & 87.57\textsubscript{\textcolor{Green}{+2.79}} & 87.28\textsubscript{\textcolor{Green}{+4.70}} & 86.63\textsubscript{\textcolor{Green}{+5.08}} & 95.59\textsubscript{\textcolor{Green}{+1.65}} & 87.29\textsubscript{\textcolor{Green}{+4.74}} & 96.41\textsubscript{\textcolor{red}{-0.20}} \\
        \hline
        \multirow{2}{*}{MixUp}     
        & \cellcolor{yellow!20}ResNet-50 & \cellcolor{yellow!20}79.19\textsubscript{\textcolor{Green}{+2.79}} & \cellcolor{yellow!20}84.50\textsubscript{\textcolor{Green}{
        +1.32}} & \cellcolor{yellow!20}74.36\textsubscript{\textcolor{Red}{-1.78}} & \cellcolor{yellow!20}79.19\textsubscript{\textcolor{Green}{+3.93}} & \cellcolor{yellow!20}78.53\textsubscript{\textcolor{Red}{-13.16}} & \cellcolor{yellow!20}92.77\textsubscript{\textcolor{Green}{+20.55}} & \cellcolor{yellow!20}76.55\textsubscript{\textcolor{Red}{-15.79}} \\
        & ViT-B    & 98.52\textsubscript{\textcolor{Green}{+13.32}} & 89.05\textsubscript{\textcolor{Green}{+4.27}} & 89.02\textsubscript{\textcolor{Green}{+6.44}} & 89.14\textsubscript{\textcolor{Green}{+7.59}} & 96.27\textsubscript{\textcolor{Green}{+2.33}} & 88.98\textsubscript{\textcolor{Green}{+6.43}} & 97.42\textsubscript{\textcolor{Green}{+0.81}} \\
        \hline
        \multirow{2}{*}{SnapMix}   
        & ResNet-50 & 78.68\textsubscript{\textcolor{Green}{+2.28}} & 84.21\textsubscript{\textcolor{Green}{+1.03}} & 78.68\textsubscript{\textcolor{Green}{+2.54}} & 77.25\textsubscript{\textcolor{Green}{+1.99}} & 92.58\textsubscript{\textcolor{Green}{+0.89}} & 76.72\textsubscript{\textcolor{Green}{+4.50}} & 93.70\textsubscript{\textcolor{Green}{+1.36}} \\
        & \cellcolor{blue!20}ViT-B  & \cellcolor{blue!20}99.44\textsubscript{\textcolor{Green}{+14.24}} & \cellcolor{blue!20}90.68\textsubscript{\textcolor{Green}{+5.90}}& \cellcolor{blue!20}90.59\textsubscript{\textcolor{Green}{+8.01}} & \cellcolor{blue!20}90.83\textsubscript{\textcolor{Green}{+9.28}} & \cellcolor{blue!20}96.72\textsubscript{\textcolor{Green}{+2.78}} & \cellcolor{blue!20}90.62\textsubscript{\textcolor{Green}{+8.07}} & \cellcolor{blue!20}97.82\textsubscript{\textcolor{Green}{+1.21}} \\
        \hline
        \multirow{2}{*}{YOCO}      
        & ResNet-50 & 
        74.37\textsubscript{\textcolor{Red}{-2.03}} & 
        79.73\textsubscript{\textcolor{Red}{-3.45}} & 
        74.37\textsubscript{\textcolor{Red}{-1.77}} & 
        72.78\textsubscript{\textcolor{Red}{-2.48}} & 
        91.11\textsubscript{\textcolor{Red}{-0.58}} & 
        71.92\textsubscript{\textcolor{Red}{-0.30}} & 
        91.79\textsubscript{\textcolor{Red}{-0.55}} \\

        & ViT-B    & 95.02\textsubscript{\textcolor{Green}{+9.82}} & 87.64\textsubscript{\textcolor{Green}{+2.86}} & 87.63\textsubscript{\textcolor{Green}{+5.05}} & 87.92\textsubscript{\textcolor{Green}{+6.37}} & 95.75\textsubscript{\textcolor{Green}{+1.81}} & 87.54\textsubscript{\textcolor{Green}{+4.99}} & 97.21\textsubscript{\textcolor{Green}{+0.60}} \\
        \hline
    \end{tabular}}
    \label{tab:compare_brain}
\end{table}

\subsection{AugMix}

\noindent AugMix \cite{hendrycks2019augmix} is a data augmentation method that enhances robustness and uncertainty estimation by mixing diverse augmentations while enforcing consistency. In medical imaging, it effectively addresses challenges such as data corruption and unseen perturbations, ensuring reliable predictions in critical tasks.

\noindent Let \( x_{\text{orig}} \) represent the original input medical image, and \( O \) denote the set of augmentation operations (e.g., rotation, translation, posterization). AugMix generates augmented images by combining multiple augmentation chains, where \( k \) is the number of chains, and \( w_i \) is the mixing weight for the \( i \)-th chain, sampled from a Dirichlet distribution: \( w \sim \text{Dirichlet}(\alpha, \dots, \alpha) \). The augmented image \( x_{\text{aug}} \) is computed as:

\[
x_{\text{aug}} = \sum_{i=1}^{k} w_i \cdot \text{chain}_i(x_{\text{orig}}),
\]

\noindent where \( \text{chain}_i(\cdot) \) represents an augmentation chain composed of multiple operations. The final AugMix image \( x_{\text{augmix}} \) is interpolated between \( x_{\text{orig}} \) and \( x_{\text{aug}} \), with interpolation weight \( m \) sampled from a Beta distribution: \( m \sim \text{Beta}(\alpha, \alpha) \). The interpolation is defined as:

\[
x_{\text{augmix}} = m \cdot x_{\text{orig}} + (1 - m) \cdot x_{\text{aug}}.
\]

\noindent To enforce consistency across embeddings, AugMix introduces the Jensen-Shannon divergence consistency loss \( \mathcal{L}_{\text{JS}} \). This loss measures the divergence between the model’s predicted distributions for the original image \( x_{\text{orig}} \) and two independently generated AugMix images \( x_{\text{augmix1}} \) and \( x_{\text{augmix2}} \). The loss is defined as:

\[
\mathcal{L}_{\text{JS}} = \text{JS}(p(y|x_{\text{orig}}), p(y|x_{\text{augmix1}}), p(y|x_{\text{augmix2}})),
\]

\noindent where \( p(y|x) \) is the model’s predicted distribution for image \( x \), and \( \text{JS}(\cdot) \) represents the Jensen-Shannon divergence.

\noindent In medical imaging, AugMix, as illustrated in panels (f) of \cref{fig:eye,fig_brain}, is particularly effective in improving robustness against unseen perturbations and enhancing uncertainty estimation. The augmented image \( x_{\text{aug}} \) combines diverse augmentation chains to introduce variability while preserving semantic consistency. The AugMix image \( x_{\text{augmix}} \) interpolates between the original and augmented images, ensuring the final image remains meaningful. The Jensen-Shannon divergence loss \( \mathcal{L}_{\text{JS}} \) enforces consistency across embeddings, improving calibration and reducing sensitivity to data corruption. This makes AugMix a valuable technique for medical imaging tasks such as disease classification, lesion segmentation, and anomaly detection.

\begin{table}[t]
    \centering
    \caption{Comparative performance on \textbf{eye diseases classification dataset} \cite{eye_diseases_classification_dataset}.}
    \resizebox{\linewidth}{!}{%
    \begin{tabular}{l|l|c|c|c|c|c|c|c}
        \hline
        \textbf{Method} & \textbf{Backbone} & \textbf{Accuracy} & \textbf{Precision} & \textbf{Recall} & \textbf{Sensitivity} & \textbf{Specificity} & \textbf{F1 Score} & \textbf{ROC AUC} \\
        \hline
        \multirow{2}{*}{Baseline}  & ResNet-50 & 90.77 & 87.95 & 85.78 & 85.79 & 85.98 & 85.98 & 96.12 \\
                                   & ViT-B    & 80.36 & 80.61 & 78.81 & 78.90 & 92.99 & 79.17 & 95.45 \\
        \hline
        \multirow{2}{*}{AugMix}    & ResNet-50 & 83.31\textsubscript{\textcolor{Red}{-7.46}} & 
        84.55\textsubscript{\textcolor{Red}{-3.40}} & 
        83.31\textsubscript{\textcolor{Red}{-2.47}} & 
        83.33\textsubscript{\textcolor{Red}{-2.46}} & 
        94.46\textsubscript{\textcolor{Green}{+8.48}} & 
        83.53\textsubscript{\textcolor{Red}{-2.45}} & 
        95.85\textsubscript{\textcolor{Red}{-0.27}} \\

                                   & ViT-B    & 93.71\textsubscript{\textcolor{Green}{+13.35}} & 82.13\textsubscript{\textcolor{Green}{+1.52}} & 81.63\textsubscript{\textcolor{Green}{+2.82}} & 81.51\textsubscript{\textcolor{Green}{+2.61}} & 93.91\textsubscript{\textcolor{Green}{+0.92}} & 81.51\textsubscript{\textcolor{Green}{+2.34}} & 94.70\textsubscript{\textcolor{red}{-0.75}} \\
        \hline
        \multirow{2}{*}{CropMix}   & ResNet-50 & 73.25\textsubscript{\textcolor{Red}{-17.52}} & 
        77.15\textsubscript{\textcolor{Red}{-10.80}} & 
        73.25\textsubscript{\textcolor{Red}{-12.53}} & 
        73.57\textsubscript{\textcolor{Red}{-12.22}} & 
        91.13\textsubscript{\textcolor{Green}{+5.15}} & 
        72.90\textsubscript{\textcolor{Red}{-13.08}} & 
        92.24\textsubscript{\textcolor{Red}{-3.88}} \\
                                   & ViT-B    & 97.32\textsubscript{\textcolor{Green}{+16.96}} & 83.91\textsubscript{\textcolor{Green}{+3.30}} & 83.56\textsubscript{\textcolor{Green}{+4.75}} & 83.55\textsubscript{\textcolor{Green}{+4.65}} & 94.54\textsubscript{\textcolor{Green}{+1.55}} & 83.66\textsubscript{\textcolor{Green}{+4.49}} & 95.13\textsubscript{\textcolor{red}{-0.32}} \\
        \hline
        \multirow{2}{*}{CutMix}    
        & ResNet-50 & 73.25\textsubscript{\textcolor{red}{-17.52}} & 77.15\textsubscript{\textcolor{red}{-10.80}} & 73.25\textsubscript{\textcolor{red}{-12.53}} & 73.57\textsubscript{\textcolor{red}{-12.22}} & 91.13\textsubscript{\textcolor{Green}{+5.15}} & 72.90\textsubscript{\textcolor{red}{-13.08}} & 92.24\textsubscript{\textcolor{red}{-3.88}} \\
        & \cellcolor{blue!20}ViT-B    & \cellcolor{blue!20}97.94\textsubscript{\textcolor{Green}{+17.58}} & \cellcolor{blue!20}81.08\textsubscript{\textcolor{Green}{+0.47}} & \cellcolor{blue!20}81.04\textsubscript{\textcolor{Green}{+2.23}} & \cellcolor{blue!20}80.87\textsubscript{\textcolor{Green}{+1.97}} & \cellcolor{blue!20}93.69\textsubscript{\textcolor{Green}{+0.70}} & \cellcolor{blue!20}80.96\textsubscript{\textcolor{Green}{+1.79}} & \cellcolor{blue!20}94.74\textsubscript{\textcolor{red}{-0.71}} \\
        \hline
        \multirow{2}{*}{MixUp}     & ResNet-50 & 88.99\textsubscript{\textcolor{Red}{-1.78}} & 
        89.50\textsubscript{\textcolor{Green}{+1.55}} & 
        88.99\textsubscript{\textcolor{Green}{+3.21}} & 
        88.68\textsubscript{\textcolor{Green}{+2.89}} & 
        96.33\textsubscript{\textcolor{Green}{+10.35}} & 
        88.75\textsubscript{\textcolor{Green}{+2.77}} & 
        97.69\textsubscript{\textcolor{Green}{+1.57}} \\
                                   & ViT-B    & 90.55\textsubscript{\textcolor{Red}{-0.22}} & 83.88\textsubscript{\textcolor{Green}{+3.27}} & 83.26\textsubscript{\textcolor{Green}{+4.45}} & 83.14\textsubscript{\textcolor{Green}{+4.24}} & 94.44\textsubscript{\textcolor{Green}{+1.45}} & 83.47\textsubscript{\textcolor{Green}{+4.30}} & 95.96\textsubscript{\textcolor{Green}{+0.51}} \\
        \hline
        \multirow{2}{*}{SnapMix}   & ResNet-50 & 87.69\textsubscript{\textcolor{Red}{-3.08}} & 
        88.81\textsubscript{\textcolor{Green}{+0.86}} & 
        87.69\textsubscript{\textcolor{Green}{+1.91}} & 
        87.32\textsubscript{\textcolor{Green}{+1.53}} & 
        95.89\textsubscript{\textcolor{Green}{+9.91}} & 
        87.59\textsubscript{\textcolor{Green}{+1.61}} & 
        96.51\textsubscript{\textcolor{Green}{+0.39}} \\

                                   & ViT-B    & 93.93\textsubscript{\textcolor{Green}{+3.16}} & 81.14\textsubscript{\textcolor{Green}{+0.53}} & 81.04\textsubscript{\textcolor{Green}{+2.23}} & 80.76\textsubscript{\textcolor{Green}{+1.86}} & 93.69\textsubscript{\textcolor{Green}{+0.70}} & 80.99\textsubscript{\textcolor{Green}{+1.82}} & 94.20\textsubscript{\textcolor{red}{-1.25}} \\
        \hline
        \multirow{2}{*}{YOCO}      & \cellcolor{yellow!20}ResNet-50 & \cellcolor{yellow!20}91.60\textsubscript{\textcolor{Green}{+0.83}} & \cellcolor{yellow!20}91.78\textsubscript{\textcolor{Green}{+3.83}} & \cellcolor{yellow!20}91.60\textsubscript{\textcolor{Green}{+5.82}} & \cellcolor{yellow!20}91.37\textsubscript{\textcolor{Green}{+5.58}} & \cellcolor{yellow!20}97.20\textsubscript{\textcolor{Green}{+11.22}} & \cellcolor{yellow!20}91.51\textsubscript{\textcolor{Green}{+5.53}} & \cellcolor{yellow!20}97.89\textsubscript{\textcolor{Green}{+1.77}} \\
                                   & ViT-B    & 97.72\textsubscript{\textcolor{Green}{+17.36}} & 87.65\textsubscript{\textcolor{Green}{+7.04}} & 87.56\textsubscript{\textcolor{Green}{+8.75}} & 87.36\textsubscript{\textcolor{Green}{+8.46}} & 95.86\textsubscript{\textcolor{Green}{+2.87}} & 87.52\textsubscript{\textcolor{Green}{+8.35}} & 97.27\textsubscript{\textcolor{Green}{+1.82}} \\
        \hline
    \end{tabular}}
    \label{tab:compare_eye}
\end{table}

\begin{figure}[t]
\centering

\begin{minipage}[b]{0.48\textwidth}
    \centering
    \includegraphics[width=\textwidth]{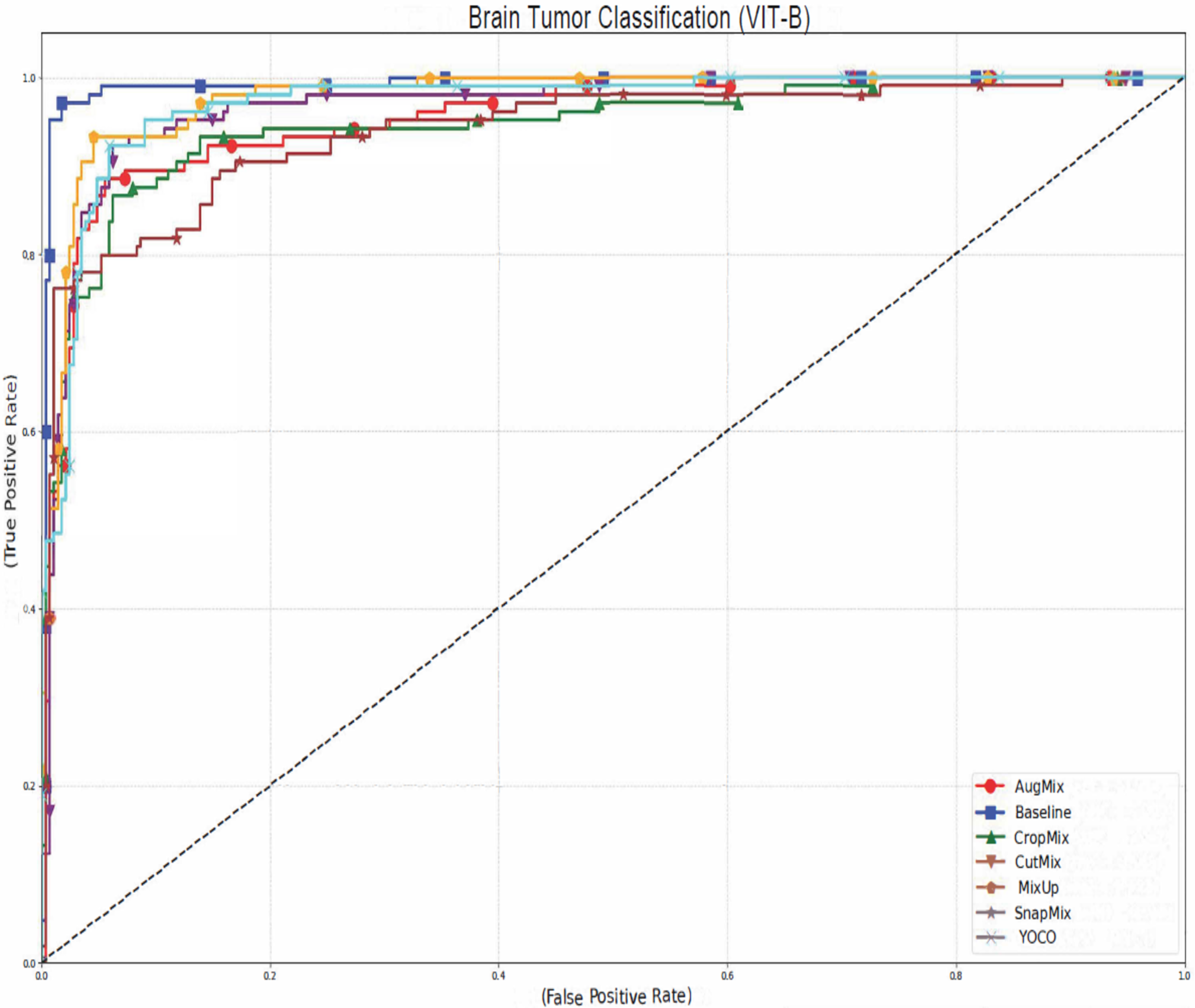}
    \caption{ROC curves with ViT-B on \textbf{brain tumor dataset} \cite{sartaj_bhuvaji_ankita_kadam_prajakta_bhumkar_sameer_dedge_swati_kanchan_2020}.}
    \label{Brain-VIT-B_roc}
\end{minipage}
\hfill
\begin{minipage}[b]{0.48\textwidth}
    \centering
    \includegraphics[width=\textwidth]{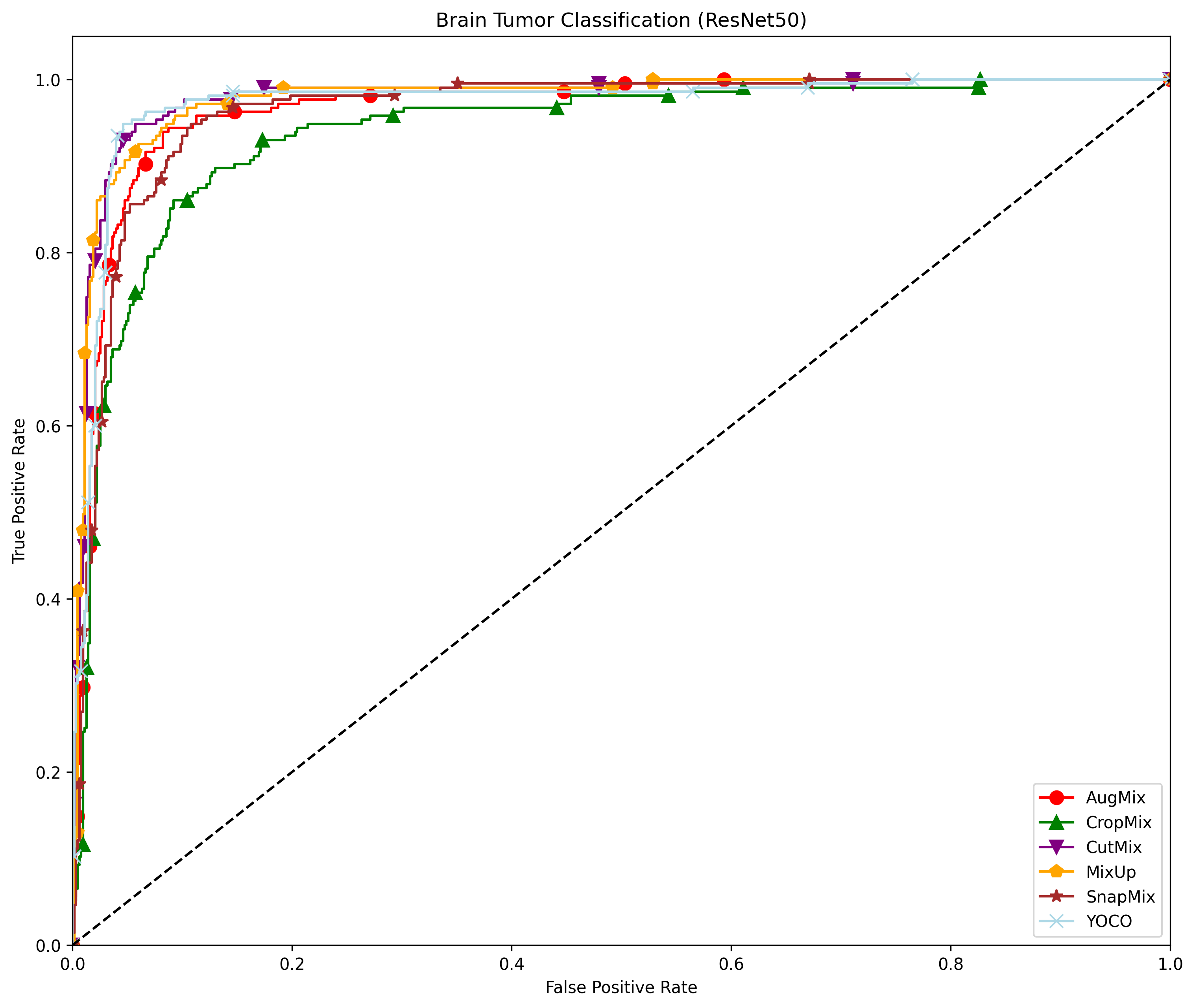}
    \caption{ROC curves with ResNet-50 on \textbf{brain tumor dataset} \cite{sartaj_bhuvaji_ankita_kadam_prajakta_bhumkar_sameer_dedge_swati_kanchan_2020}.}
    \label{Brain-resnet_roc}
\end{minipage}

\end{figure}

\subsection{SnapMix}
\noindent SnapMix \cite{huang2021snapmix} is a data augmentation method that leverages class activation maps (CAM) to ensure semantic consistency when mixing images and labels. In medical imaging, SnapMix enhances model performance by focusing on localized features such as lesions, mitigating overfitting, and improving robustness in fine-grained tasks like disease classification and lesion detection.

\noindent Let \( I_i \) denote an input medical image with label \( y_i \) (e.g., disease class or lesion type). The class activation map \( \text{CAM}_i \) for \( I_i \) is normalized to sum to 1, producing a semantic percent map \( \text{SPM}_i \) that represents pixel-level semantic relevance. A region \( R_i \) is cropped from \( I_i \), and its semantic ratio \( \text{SR}_i \) is computed as:

\[
\text{SR}_i = \sum_{p \in R_i} \text{SPM}_i(p),
\]

\noindent where \( p \) represents a pixel in the cropped region \( R_i \). The semantic ratio \( \text{SR}_i \) quantifies the semantic importance of the region \( R_i \) based on the values in \( \text{SPM}_i \).

\noindent To create a mixed image \( \tilde{I} \), SnapMix combines cropped regions \( R_a \) and \( R_b \) from two input images \( I_a \) and \( I_b \):

\[
\tilde{I} = R_a + R_b.
\]

\noindent The mixed label \( \tilde{y} \) for \( \tilde{I} \) is computed by weighting the original labels \( y_a \) and \( y_b \) of \( I_a \) and \( I_b \) according to their semantic ratios \( \text{SR}_a \) and \( \text{SR}_b \):

\[
\tilde{y} = \frac{\text{SR}_a}{\text{SR}_a + \text{SR}_b} \cdot y_a + \frac{\text{SR}_b}{\text{SR}_a + \text{SR}_b} \cdot y_b.
\]

\noindent This ensures that the mixed label \( \tilde{y} \) accurately reflects the semantic contributions of \( R_a \) and \( R_b \) to the mixed image \( \tilde{I} \).

\noindent In medical imaging, SnapMix, as illustrated in panels (g) of \cref{fig:eye,fig_brain}, is particularly effective for tasks involving fine-grained details, such as lesion segmentation and disease classification. By using CAMs to guide the mixing process, SnapMix ensures that the mixed images retain meaningful semantic information. The semantic ratio \( \text{SR}_i \) captures the importance of specific regions, allowing the method to focus on areas of high diagnostic relevance. This approach mitigates overfitting, enhances model robustness to data variability, and improves generalization to unseen cases, making SnapMix a valuable tool in medical image analysis.

\begin{figure}[t]
\centering

\begin{minipage}[b]{0.48\textwidth}
    \centering
    \includegraphics[width=\textwidth]{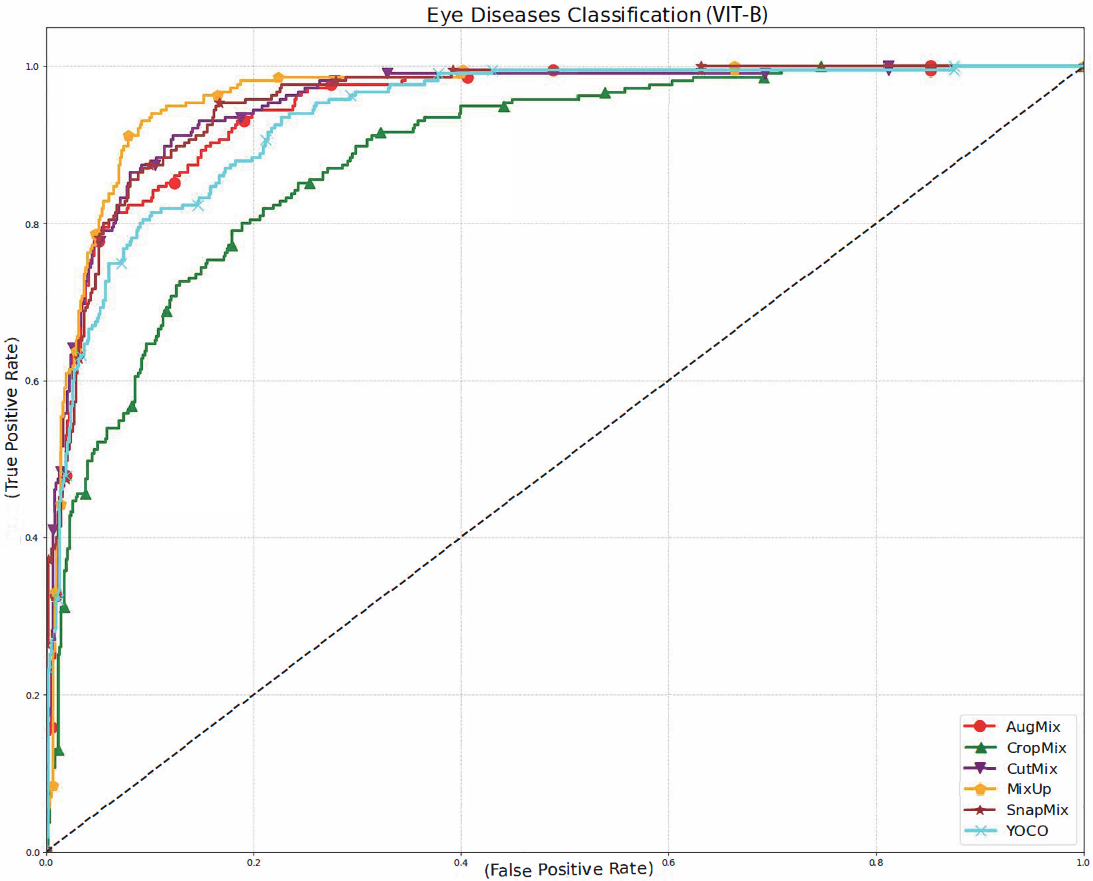}
    \caption{ROC curves with ViT-B on \textbf{eye diseases dataset} \cite{eye_diseases_classification_dataset}.}
    \label{Eye-VIT-B_roc}
\end{minipage}
\hfill
\begin{minipage}[b]{0.48\textwidth}
    \centering
    \includegraphics[width=\textwidth]{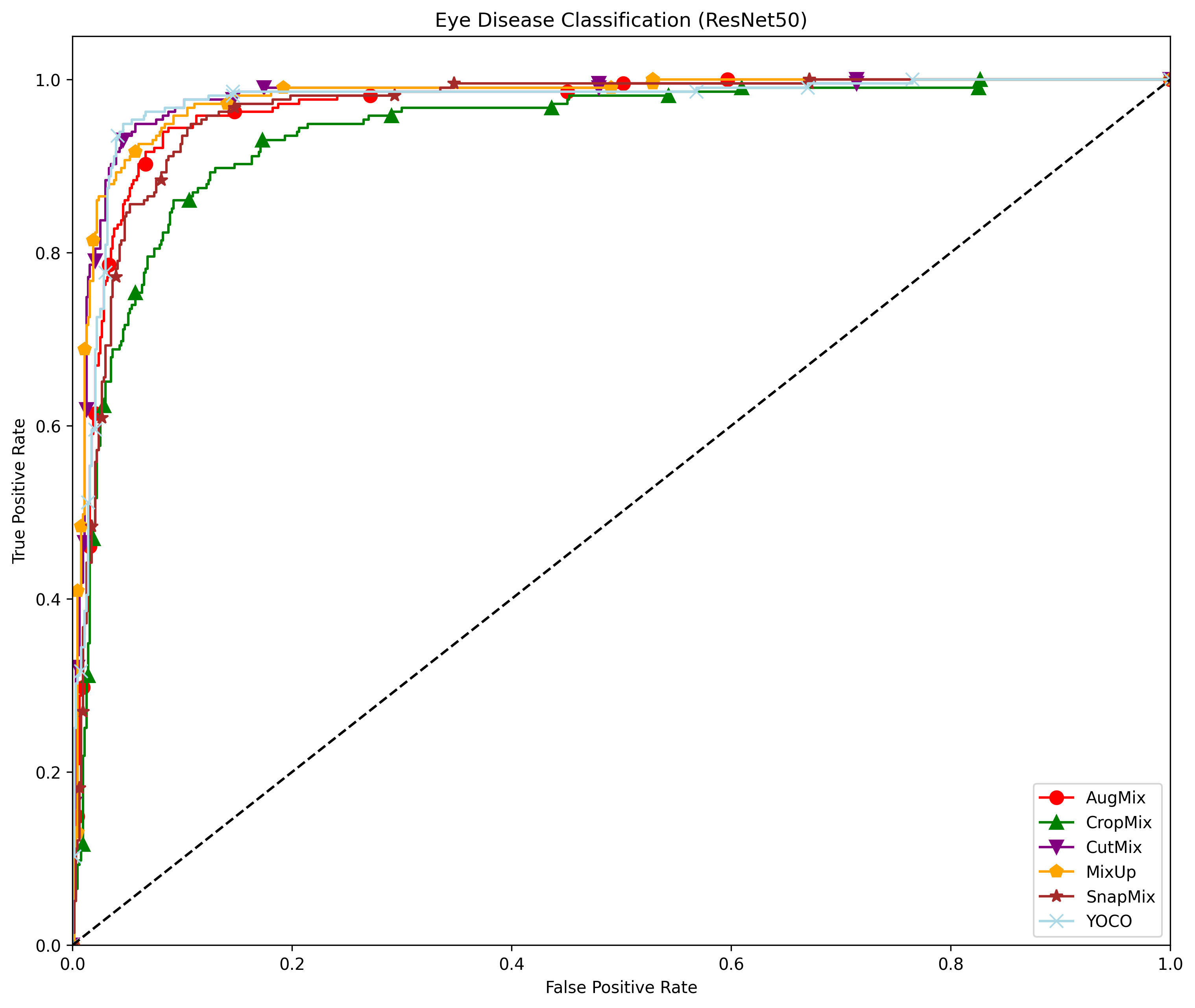}
    \caption{\small ROC curves with ResNet-50 on \textbf{eye diseases dataset} \cite{eye_diseases_classification_dataset}.}
    \label{Eye-resnet_roc}
\end{minipage}

\end{figure}

\section{Experiments}

\subsection{Datasets and Evaluation Matrices}

\paragraph{Brain Tumor Classification Dataset}

 \cite{sartaj_bhuvaji_ankita_kadam_prajakta_bhumkar_sameer_dedge_swati_kanchan_2020} includes four categories: glioma tumor, meningioma tumor, no tumor, and pituitary tumor. A pie chart \cref{fig:brain-tumor-pie} reveals an imbalanced class distribution. t-SNE visualization \cref{fig:brain-tumor-tsne} shows pituitary tumor forms distinct clusters, while glioma tumor and meningioma tumor partially overlap. No tumor is dispersed and overlaps significantly with other categories, increasing classification difficulty.

\paragraph{Eye Diseases Classification Dataset}

 \cite{eye_diseases_classification_dataset} includes four categories: cataract, diabetic retinopathy, glaucoma, and normal. A balanced class distribution \cref{fig:eye-disease-pie} makes it suitable for multi-class classification. t-SNE visualization \cref{fig:eye-disease-tsne} shows clear clustering for cataract and diabetic retinopathy, while glaucoma and normal partially overlap, increasing classification difficulty.

\paragraph{Evaluation Matrices:}

\textit{Accuracy} measures overall correctness. \textit{Precision} assesses the accuracy of positive predictions, while \textit{Recall} measures the ability to identify actual positives. \textit{Sensitivity} (similar to Recall) focuses on detecting positives, and \textit{Specificity} evaluates the identification of negatives. \textit{F1-Score} balances Precision and Recall, and \textit{ROC AUC} reflects the model's ability to distinguish between classes, with higher values indicating better performance.

\begin{table}[t]
    \centering
    \caption{Ablation study on the CutMix interpolation parameter $\alpha$ using the ViT-B backbone on the eye diseases dataset, demonstrating CutMix performance across different $\alpha$ values.}
    \resizebox{\linewidth}{!}{%
    \begin{tabular}{c|c|c|c|c|c|c|c}
        \hline
        \textbf{$\alpha$ value} & \textbf{Accuracy} & \textbf{Precision} & \textbf{Recall} & \textbf{Sensitivity} & \textbf{Specificity} & \textbf{F1 Score} & \textbf{ROC AUC} \\
        \hline
        \textbf{$\boldsymbol{\alpha = 1.0}$} & \textbf{97.94} & \textbf{81.08} & \textbf{81.04} & \textbf{80.87} & \textbf{93.69} & \textbf{80.96} & \textbf{94.74} \\
        $\alpha = 0.8$ & 96.78 & 80.42 & 80.33 & 80.15 & 93.21 & 80.37 & 94.32 \\
        $\alpha = 0.6$ & 95.35 & 79.76 & 79.82 & 79.64 & 92.87 & 79.79 & 93.86 \\
        $\alpha = 0.4$ & 93.21 & 78.45 & 78.61 & 78.40 & 92.35 & 78.53 & 93.12 \\
        $\alpha = 0.2$ & 90.64 & 76.92 & 77.18 & 76.95 & 91.83 & 77.05 & 92.47 \\
        \hline
    \end{tabular}}
    \label{tab:ablation_alpha}
\end{table}

\subsection{Implementation Details}
The datasets were randomly split into Training and Testing sets with an 8:2 ratio, using a fixed random seed to ensure reproducibility. Six data augmentation methods were applied exclusively to the training sets, while the testing sets remained untouched to evaluate model generalization. All models were trained for 50 epochs using an initial learning rate of 0.001 and the Adam optimizer. Experiments ran on an NVIDIA A100-SXM4-40GB GPU, CUDA 12.4, Intel Xeon CPU @ 2.20GHz, and 80G memory.

\subsection{Comparative Study}

Tables~\ref{tab:compare_brain} and~\ref{tab:compare_eye} present our comparative experiments on augmentation techniques for medical image classification. For brain tumour classification (Table~\ref{tab:compare_brain}), \textbf{MixUp} with ResNet-50 achieves the highest performance gains, increasing accuracy to \textbf{79.19\%}, while \textbf{SnapMix} with ViT-B achieves the highest gains, increasing accuracy to \textbf{99.44\%}, as evidenced by the ROC curves in Figs.~\ref{Brain-VIT-B_roc} and~\ref{Brain-resnet_roc}.

For eye diseases classification (Table~\ref{tab:compare_eye}), \textbf{YOCO} with ResNet-50 achieves the highest performance gains, increasing accuracy by 0.83\% to \textbf{91.60\%} and ROC AUC by 1.77\% to \textbf{97.89\%}. Meanwhile, \textbf{CutMix} with ViT-B achieves the highest gains, increasing accuracy by 17.58\% to \textbf{97.94\%}, as shown in Figs.~\ref{Eye-VIT-B_roc} and~\ref{Eye-resnet_roc}.

Our study indicates that specific combinations excel in particular applications, with MixUp on ResNet-50 and SnapMix on ViT-B optimal for brain tumour classification and YOCO on ResNet-50 and CutMix on ViT-B optimal for eye disease classification.

\subsection{Ablation Study}

Tables~\ref{tab:ablation_alpha} and~\ref{tab:ablation_alpha_resnet50} present ablation studies on the CutMix interpolation parameter $\alpha$. Our experiments show that performance consistently improves as $\alpha$ increases across both ViT-B and ResNet-50 architectures, with the optimal value $\alpha=1.0$ yielding the highest scores across all metrics, ViT-B achieving \textbf{97.94\%} accuracy and ResNet-50 reaching \textbf{91.83\%} accuracy. These results underscore the importance of systematic hyperparameter sweeps for mix-based augmentations but as demonstrated in Section~4.3 they only reflect CutMix’s internal parameter tuning and do not alter the overall optimal strategies, where MixUp on ResNet-50 and SnapMix on ViT-B remain best for brain tumour classification and YOCO on ResNet-50 and CutMix on ViT-B remain best for eye disease classification. Moreover, this concrete example of CutMix tuning illustrates a general approach: the same protocol can and should be applied to fine-tune MixUp’s interpolation weight, SnapMix’s semantic mixing ratios and YOCO’s region split configurations, demonstrating that hyperparameter optimization holds practical significance across all mix-based augmentation methods.

\begin{table}[t]
    \centering
    \caption{Ablation study on the CutMix interpolation parameter $\alpha$ using ResNet-50 backbone on the eye diseases dataset, demonstrating CutMix performance across different $\alpha$ values.}
    \resizebox{\linewidth}{!}{%
    \begin{tabular}{c|c|c|c|c|c|c|c}
        \hline
        \textbf{$\alpha$ value} & \textbf{Accuracy} & \textbf{Precision} & \textbf{Recall} & \textbf{Sensitivity} & \textbf{Specificity} & \textbf{F1 Score} & \textbf{ROC AUC} \\
        \hline
        \textbf{$\boldsymbol{\alpha = 1.0}$} & \textbf{91.83} & \textbf{91.97} & \textbf{91.83} & \textbf{91.67} & \textbf{97.29} & \textbf{91.86} & \textbf{98.22} \\
        $\alpha = 0.8$ & 91.25 & 91.39 & 91.25 & 91.08 & 97.05 & 91.32 & 97.86 \\
        $\alpha = 0.6$ & 90.86 & 90.98 & 90.86 & 90.72 & 96.84 & 90.92 & 97.43 \\
        $\alpha = 0.4$ & 90.42 & 90.58 & 90.42 & 90.25 & 96.65 & 90.50 & 97.12 \\
        $\alpha = 0.2$ & 90.15 & 90.28 & 90.15 & 90.04 & 96.42 & 90.21 & 96.85 \\
        \hline
    \end{tabular}}
    \label{tab:ablation_alpha_resnet50}
\end{table}

\section{Conclusion}

In this work MediAug provides a reproducible benchmark for six mix based augmentation methods on brain tumour MRI and eye disease fundus classification. Experiments using ResNet 50 and ViT-B identify MixUp with ResNet 50 and SnapMix with ViT-B as optimal for brain tumour classification and YOCO with ResNet 50 and CutMix with ViT-B as optimal for eye disease classification. A CutMix ablation highlights the value of systematic hyperparameter tuning and can be applied to other mix based methods. By offering clear, reproducible guidance on optimal augmentation architecture combinations and tuning protocols MediAug empowers the medical imaging community to develop more robust, generalizable AI models and accelerate the safe and effective integration of deep learning into clinical decision making.

\clearpage
\appendix
\setcounter{page}{1}
\setcounter{section}{0}
\setcounter{table}{0}
\setcounter{figure}{0}
{\LARGE \textbf{Appendix}}
\section{Data Augmentation Methods}

\begin{figure}[H]
\centering
\includegraphics[width=\textwidth]{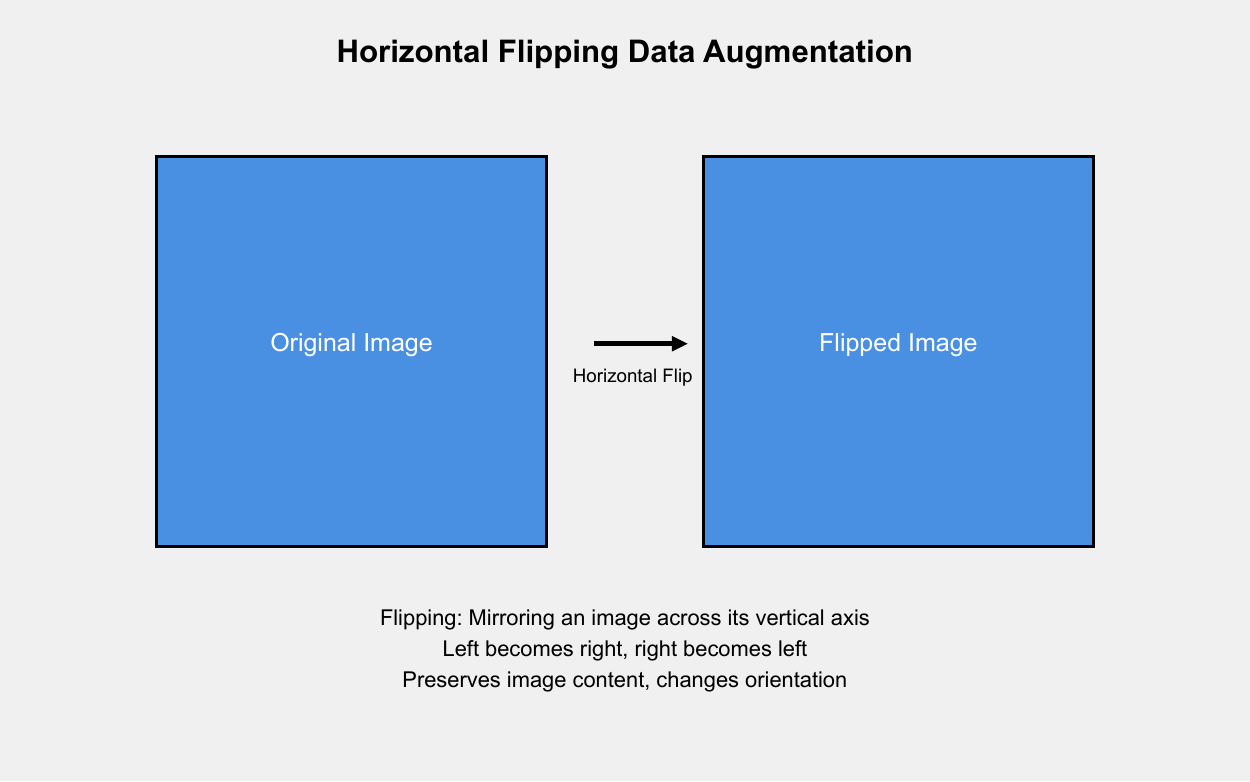}
\caption{\textit{Flipping}is a data augmentation technique that generates a mirror image by inverting pixel positions across a specified axis, typically horizontally, to expand training datasets and improve model generalization.}
\label{Flipping}
\end{figure}

1-1-a.\textit{Flipping} \cite{nalepa2019data,goceri2023medical} is a data augmentation technique that generates a mirrored version of an image by inverting pixel positions along a specified axis, typically horizontal or vertical, as shown in \cref{Flipping}. While horizontal flipping is widely used to enhance model robustness to spatial variations, vertical flipping is less common due to the potential semantic inconsistencies between the top and bottom regions of an image.

\begin{figure}[H]
\centering
\includegraphics[width=\textwidth]{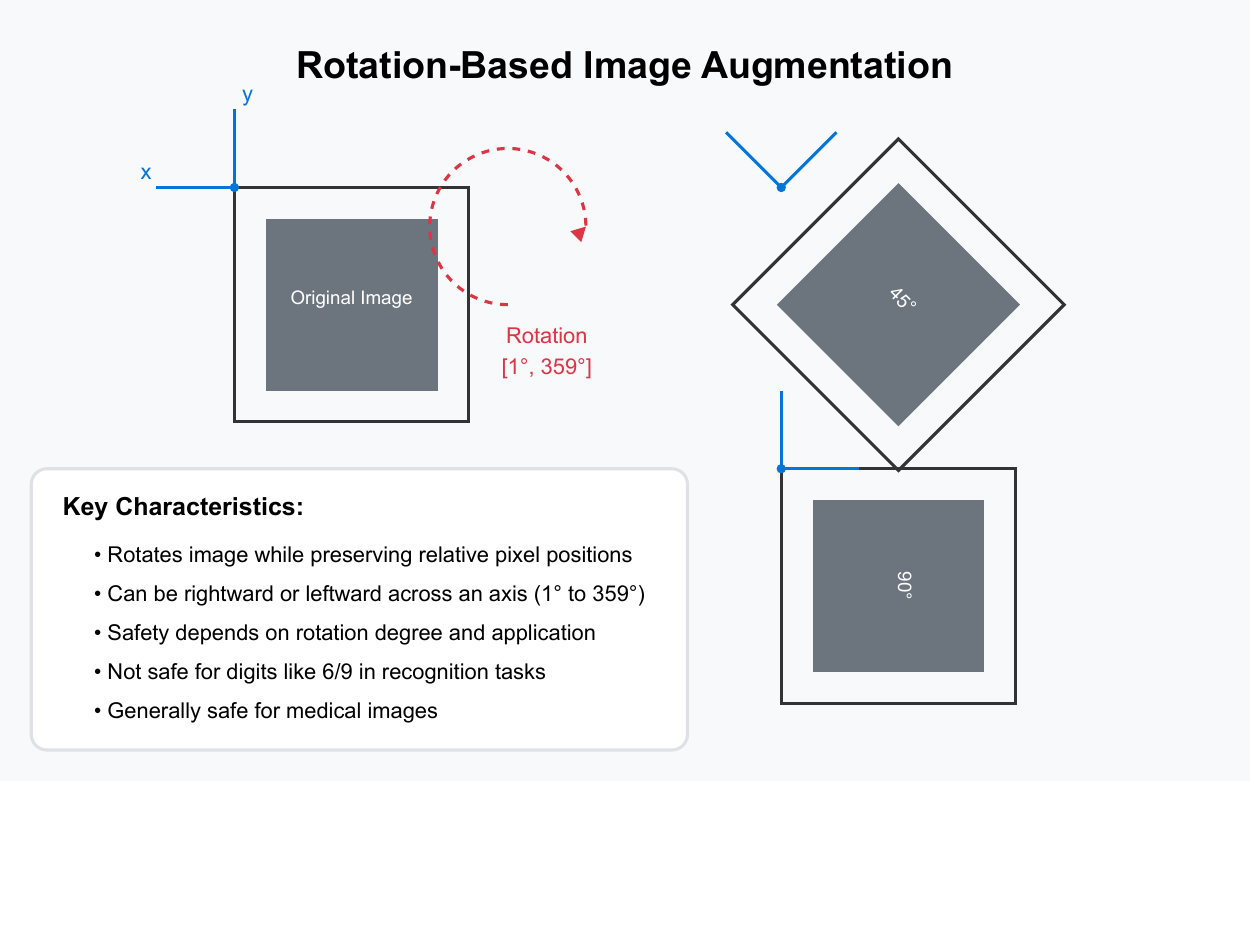}
\caption{\textit{Rotation} based image augmentation transforms the original image by pivoting it around its centre across an axis within 1-359 degrees while preserving pixel relationships.}
\label{Rotation}
\end{figure}

\newpage
1-1-b.\textit{Rotation} \cite{goceri2023medical} is a data augmentation technique that transforms an image by rotating it around a fixed axis within a specified range of angles, thereby enhancing model robustness to orientation variations, as shown in \cref{Rotation}. While this method is generally effective for tasks like medical imaging, its applicability depends on the rotation degree, as excessive rotation may alter label semantics in certain contexts, such as digit recognition.

\begin{figure}[H]
\centering
\includegraphics[width=\textwidth]{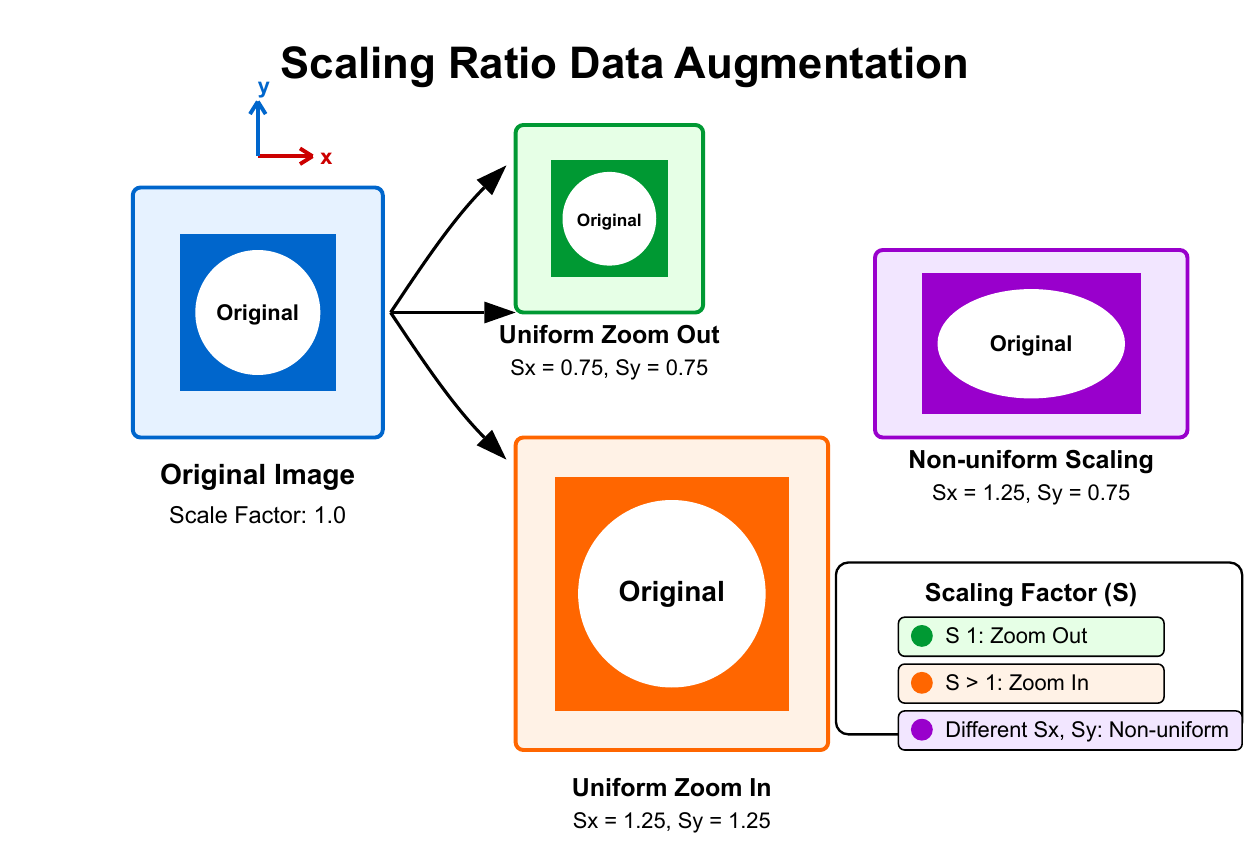}
\caption{\textit{Scaling Ratio} is a data augmentation technique that transforms images by applying different scaling factors to the x and y axes, creating zoom-in, zoom-out, or non-uniform distortion effects.}
\label{Scaling Ratio}
\end{figure}

\newpage
1-1-c.\textit{ Scaling Ratio} \cite{goceri2023medical} is a data augmentation technique that resizes images along one or more axes using a specified scaling factor, which can either be uniform or non-uniform across axes, as shown in \cref{Scaling Ratio}. This method effectively simulates zoom-in effects when the scaling factor is greater than 1 and zoom-out effects when it is less than 1, enabling models to learn features across varying spatial resolutions.

\begin{figure}[H]
\centering
\includegraphics[width=\textwidth]{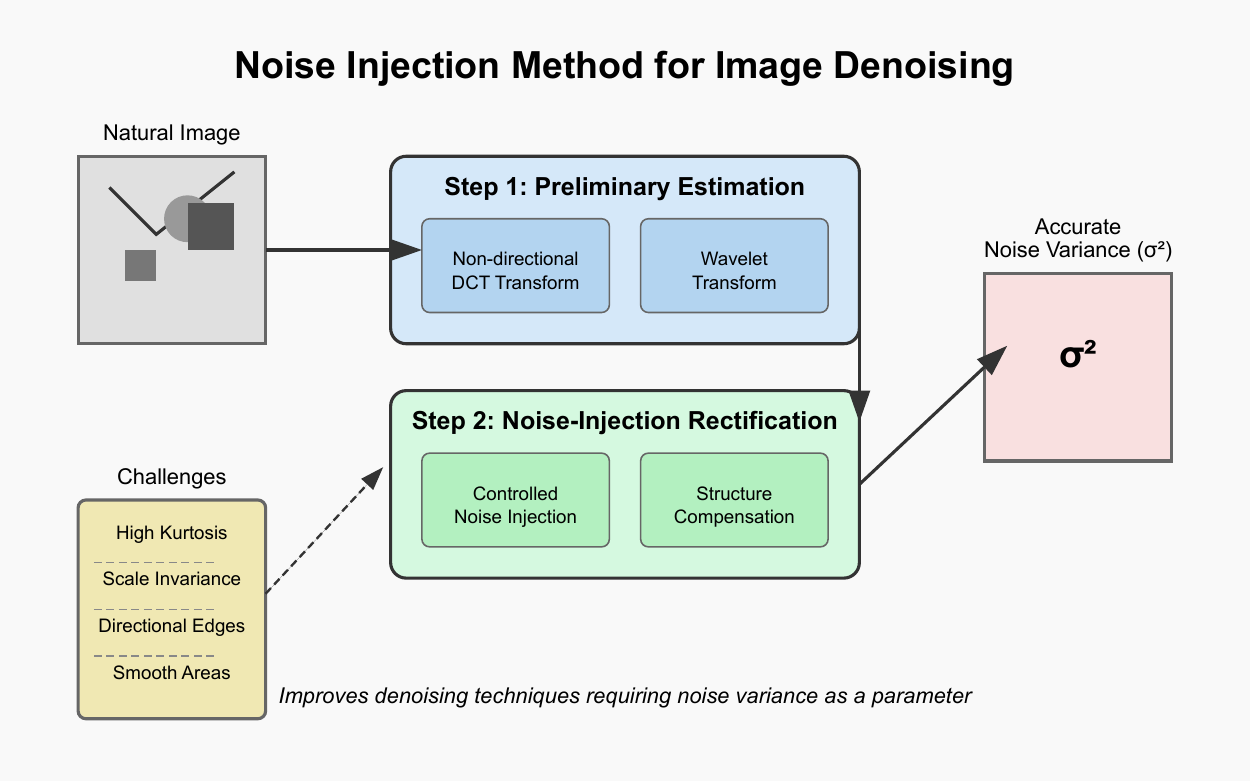}
\caption{\textit{Noise injection} method combines transform integration and controlled noise addition to accurately estimate variance in natural images, enabling improved denoising performance.}
\label{Noise injection}
\end{figure}

\newpage
1-1-d.\textit{ Noise injection} \cite{tang2014noise} is a data augmentation method that introduces random noise into images to enhance model robustness and generalization by simulating real-world imperfections, as shown in \cref{Noise injection}. This study proposes a novel two-step noise estimation framework, leveraging wavelet and DCT transforms, to accurately infer noise variance and improve the performance of denoising techniques across diverse visual content and noise levels. 

\begin{figure}[H]
\centering
\includegraphics[width=\textwidth]{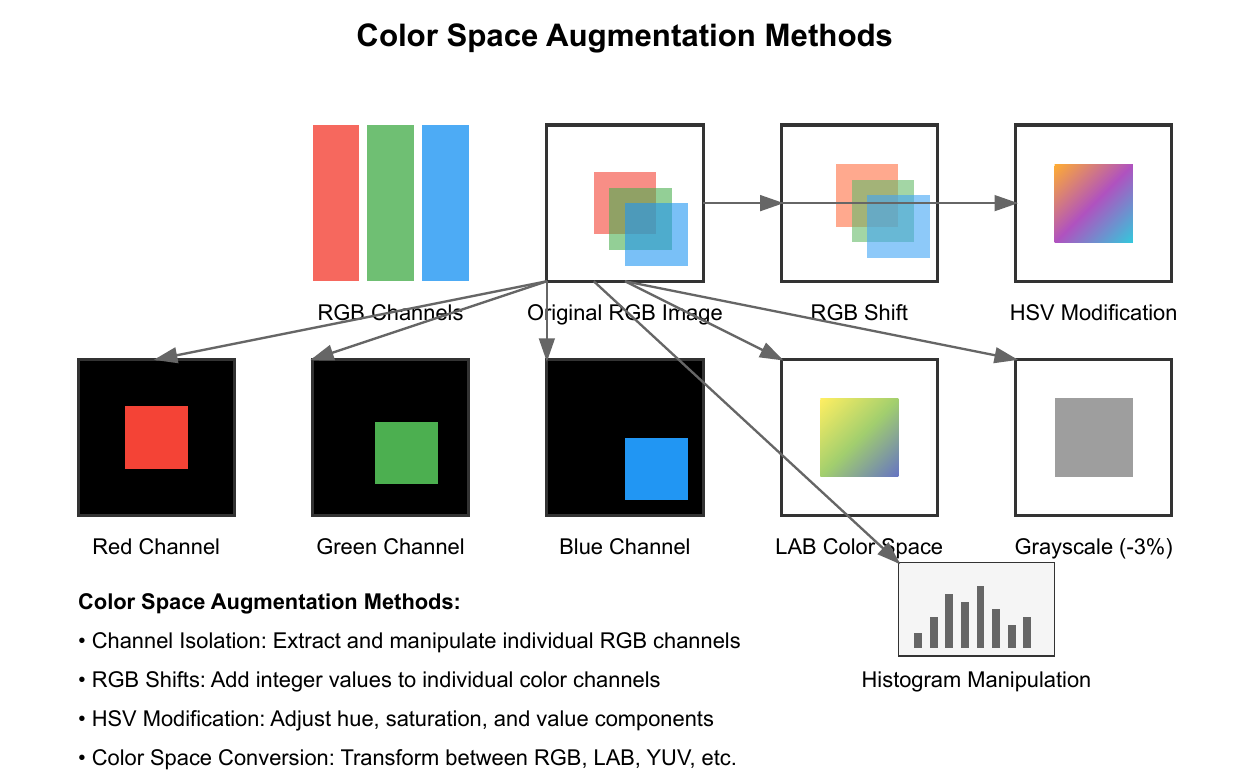}
\caption{\textit{Colour space} augmentation transforms images by manipulating RGB channels, applying colour shifts, modifying HSV components, isolating individual channels, and converting between different colour spaces while preserving spatial properties.}
\label{Colour space}
\end{figure}

\newpage
1-1-e.\textit{ Colour space} \cite{chatfield2014return} augmentation involves modifying the colour properties of an image by manipulating its colour channels, such as red, green, and blue (RGB), or by converting between different colour spaces like HSV or grayscale, as shown in \cref{Colour space}. This technique enables the generation of diverse image variations while preserving spatial information, though care must be taken with certain transformations, such as grayscale conversion, to avoid potential performance degradation.

\begin{figure}[H]
\centering
\includegraphics[width=\textwidth]{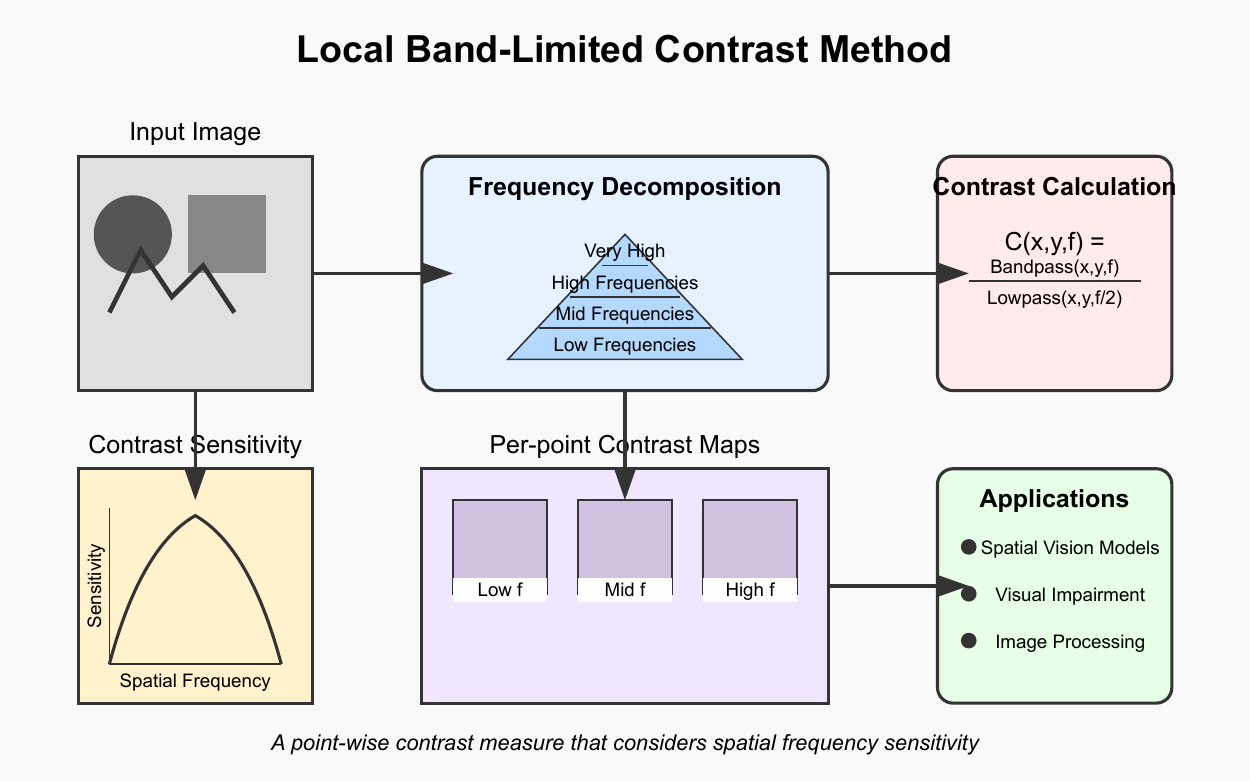}
\caption{\textit{Contrast} is calculated as a point-wise measure using frequency decomposition to analyze spatial vision and image processing applications.}
\label{Contrast}
\end{figure}

\newpage
1-1-f.\textit{ Contrast} \cite{peli1990contrast} augmentation is a data augmentation technique that adjusts the intensity differences within an image to enhance its visual features and improve model performance under varying lighting conditions, as shown in \cref{Contrast}. This study introduces a local band-limited contrast definition, considering spatial frequency content, to better understand contrast perception in complex images and its implications for image-processing algorithms.

\begin{figure}[H]
\centering
\includegraphics[width=\textwidth]{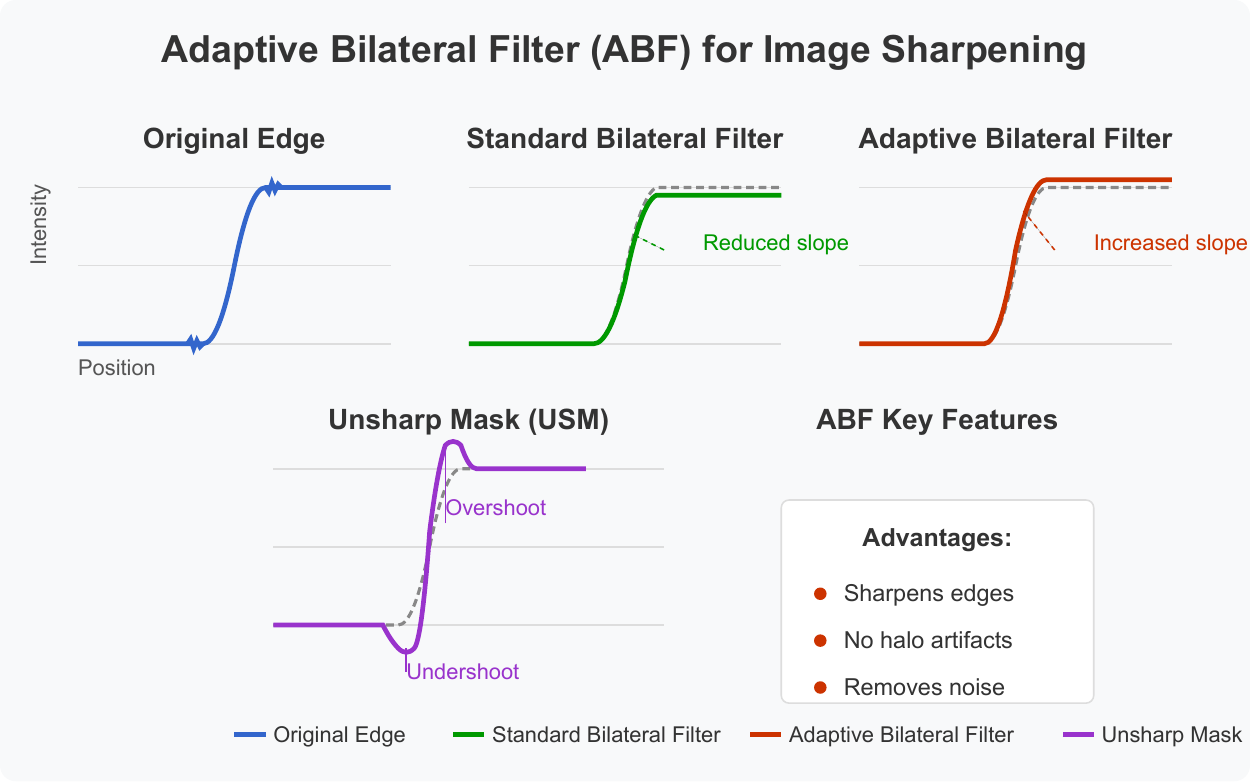}
\caption{\textit{Sharpening} with Adaptive Bilateral Filter increases edge slopes without halos while removing noise, unlike traditional Unsharp Mask techniques.}
\label{Sharpening}
\end{figure}

\newpage
1-1-g.\textit{ Sharpening} \cite{zhang2008adaptive} is a data augmentation method that enhances image sharpness by increasing edge slopes while minimizing noise and avoiding artifacts like overshoot or halo effects, as shown in \cref{Sharpening}. The adaptive bilateral filter (ABF) achieves this by transforming histograms with adaptive parameters, outperforming traditional methods like unsharp masking in both sharpness and noise reduction.

\begin{figure}[H]
\centering
\includegraphics[width=\textwidth]{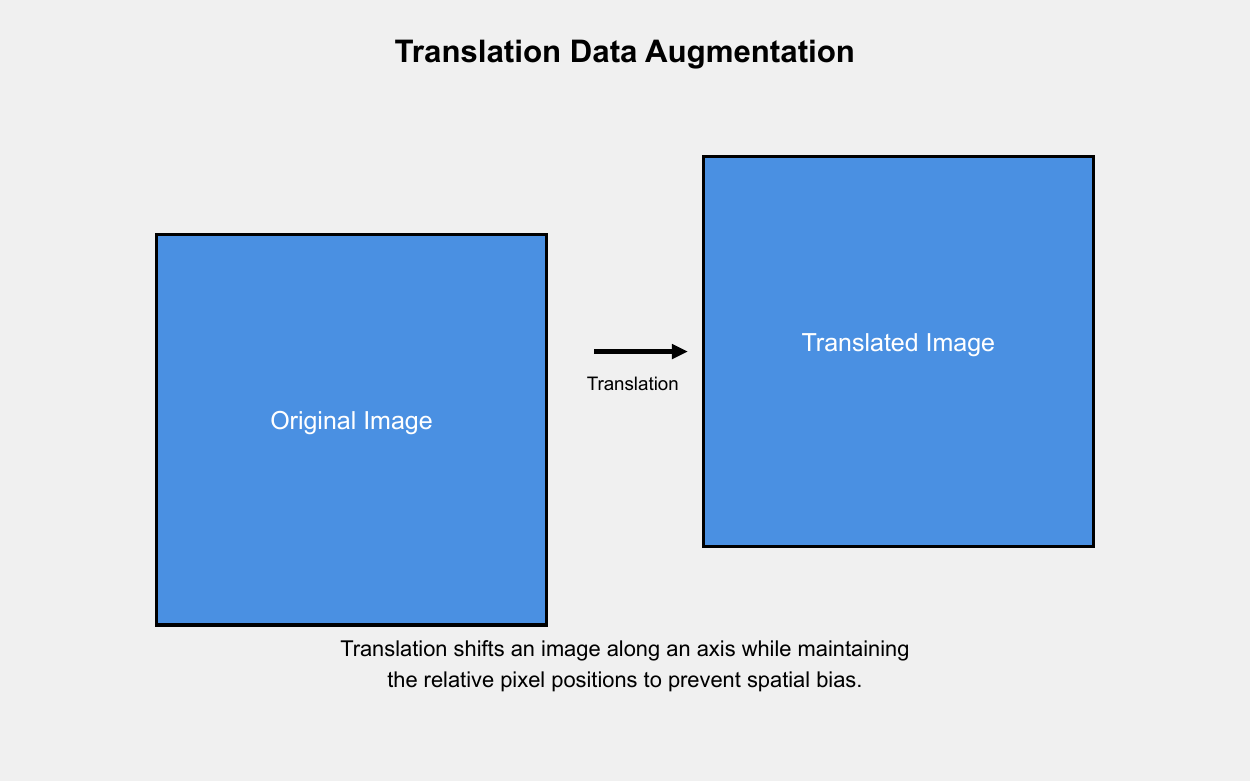}
\caption{\textit{Translation} is a data augmentation method that shifts an image along an axis while preserving pixel relationships, helping machine learning models develop spatial invariance and prevent location-specific biases.}
\label{Translation}
\end{figure}

\newpage
1-1-h.\textit{ Translation} \cite{shorten2019survey,nalepa2019data,goceri2023medical} is a data augmentation technique that shifts an image along a specified axis using a translation vector while preserving the relative positions of its pixels, as shown in \cref{Translation}. This method reduces positional bias by ensuring that models learn spatially invariant features rather than relying on properties tied to a specific location within the image.

\begin{figure}[H]
\centering
\includegraphics[width=\textwidth]{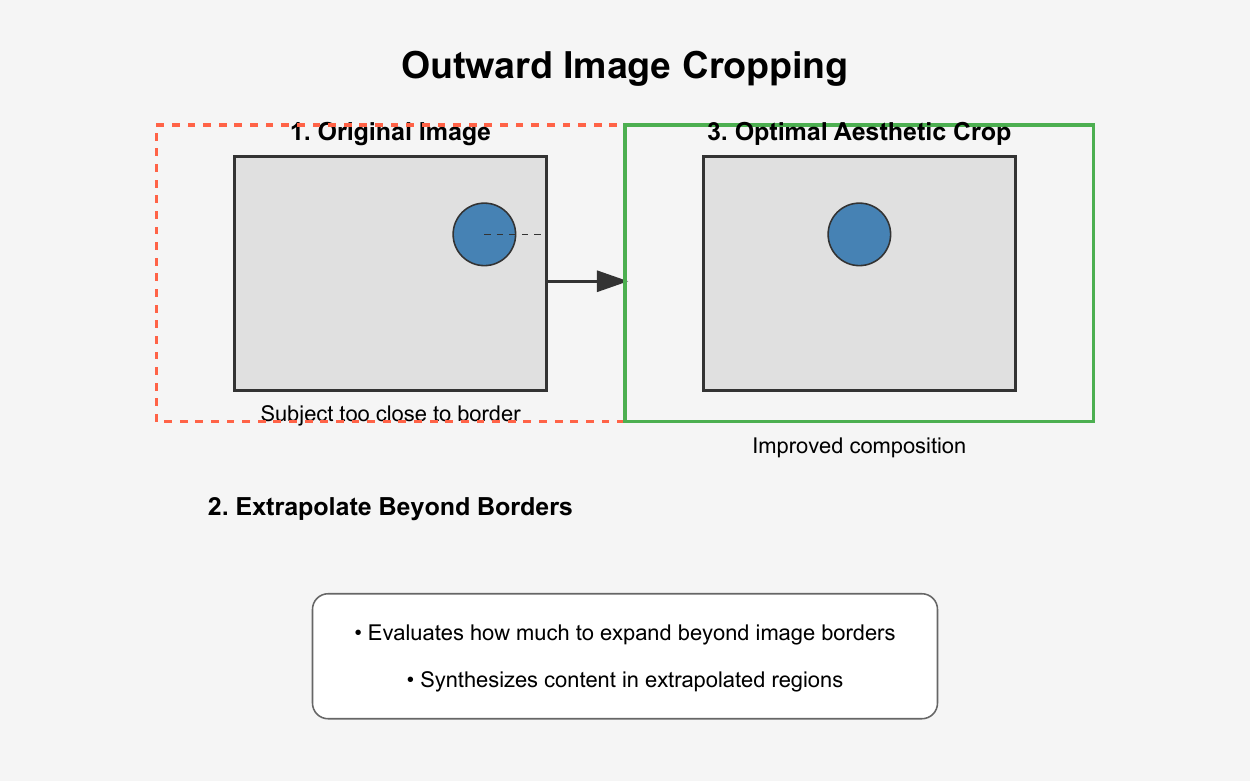}
\caption{\textit{Cropping} a novel method that extrapolates image content outward to achieve optimal aesthetic composition.}
\label{Cropping}
\end{figure}

\newpage
1-1-i.\textit{ Cropping} \cite{zhong2021aesthetic} of the image is a widely utilized data augmentation technique in computer vision, aimed at refining image composition, enhancing visual aesthetics, and improving the robustness of machine learning models, as shown in \cref{Cropping}.

\begin{figure}[H]
\centering
\includegraphics[width=\textwidth]{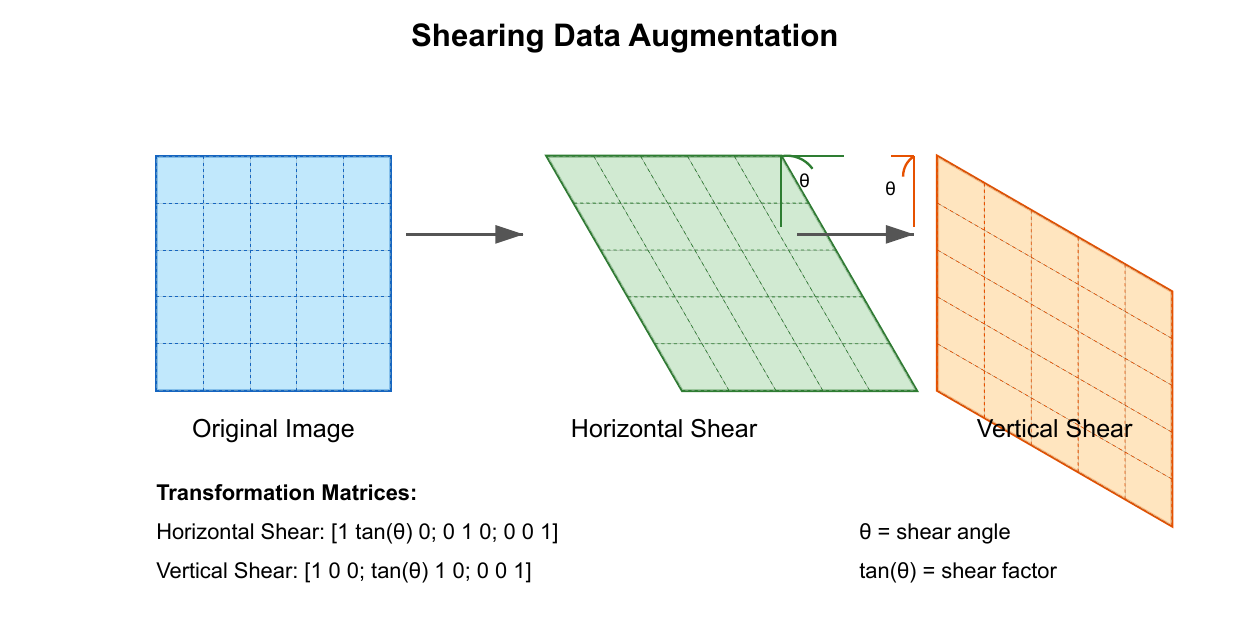}
\caption{\textit{Shearing} transforms an image by sliding one edge along a vertical or horizontal axis to create a parallelogram effect based on a specified angle.}
\label{Shearing}
\end{figure}

\newpage
1-1-j.\textit{ Shearing} \cite{goceri2023medical} is a data augmentation technique that distorts an image by shifting one edge along the horizontal or vertical axis, effectively transforming it into a parallelogram, as shown in \cref{Shearing}. Controlled by a shear angle, this method introduces perspective distortion, allowing models to learn features invariant to geometric transformations.

\begin{figure}[H]
\centering
\includegraphics[width=\textwidth]{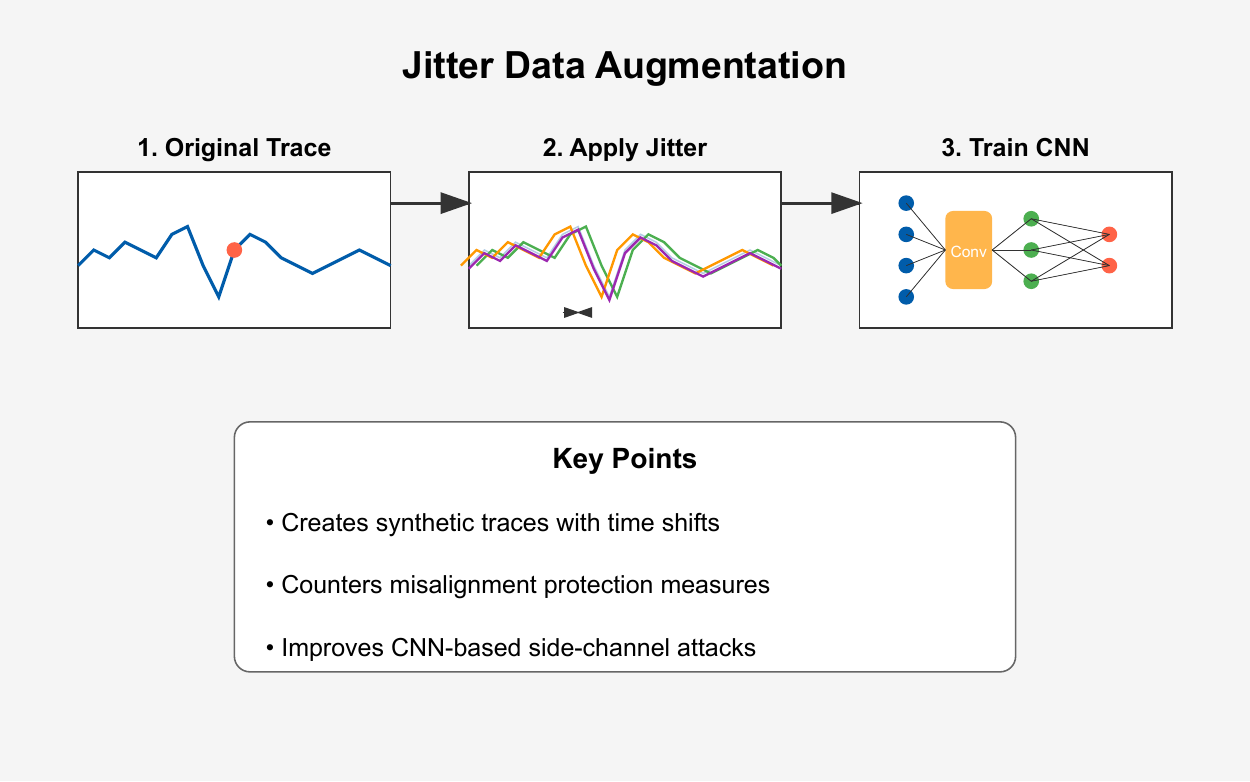}
\caption{\textit{Jitter} data augmentation: A technique that generates synthetic power trace variations with time shifts to train CNN models for robust side-channel attacks against protected cryptographic implementations.}
\label{Jitter}
\end{figure}

\newpage
1-1-k.\textit{ Jitter} \cite{cagli2017convolutional} is a data augmentation technique commonly used in machine learning to simulate temporal or spatial misalignments in input data, thereby improving model robustness against variations caused by noise or countermeasures, as shown in \cref{Jitter}. In the context of cryptographic implementation analysis, jitter augmentation is particularly effective for addressing misalignment issues in profiling attacks, enabling end-to-end strategies such as Convolutional Neural Networks to achieve high performance without requiring extensive preprocessing or precise feature selection. 

\begin{figure}[H]
\centering
\includegraphics[width=\textwidth]{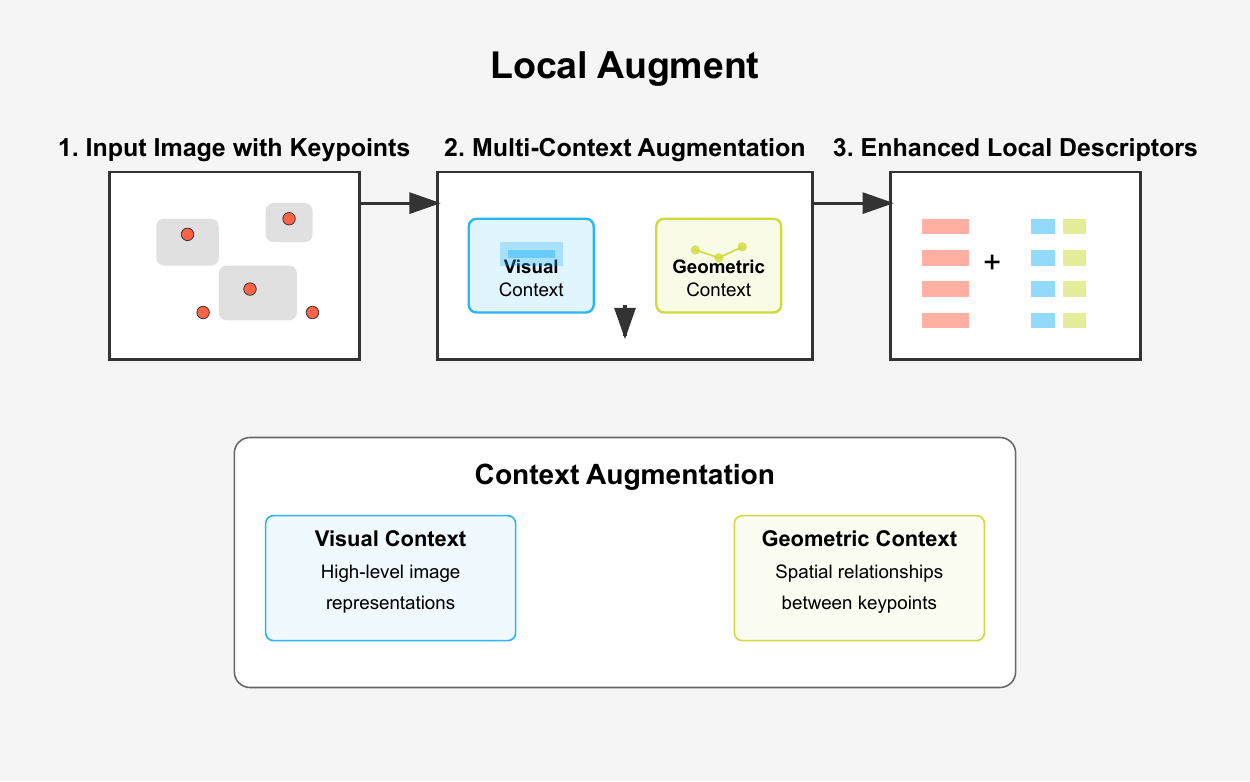}
\caption{\textit{Local Augment}: a feature enhancement method that incorporates both visual and geometric contextual information to improve keypoint matching across diverse scenes.}
\label{Local Augment}
\end{figure}

\newpage
1-1-l.\textit{ Local Augment} \cite{luo2019contextdesc} is a data augmentation method designed to enhance local feature descriptors by incorporating contextual information, such as high-level visual representations and geometric relationships derived from keypoint distributions, as shown in \cref{Local Augment}. By integrating these cross-modality contexts, this approach improves the robustness and generalization of local features, enabling more effective performance in tasks like geometric matching across diverse scenes.

\begin{figure}[H]
\centering
\includegraphics[width=\textwidth]{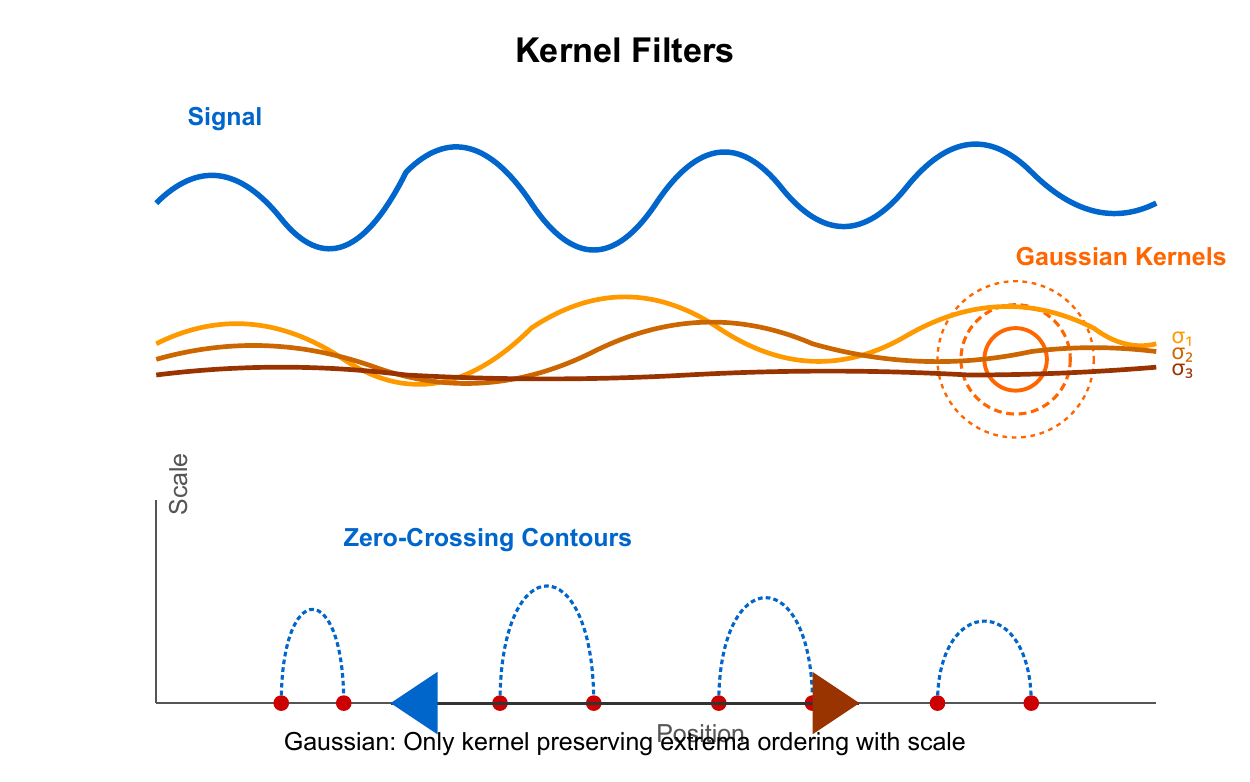}
\caption{\textit{Kernel Filters} a scale-space transformation method that convolves signals with Gaussian kernels of varying bandwidths to create hierarchical representations through zero-crossing analysis.}
\label{Kernel Filters}
\end{figure}

\newpage
1-1-m.\textit{ Kernel Filters} \cite{babaud1986uniqueness} such as Gaussian filters play a critical role in data augmentation by introducing variations such as blurring and sharpening, thereby enhancing the diversity of training data, as shown in \cref{Kernel Filters}. In scale-space filtering, Gaussian kernels are uniquely suited for hierarchical signal analysis due to their ability to maintain consistent extrema behaviour and well-defined zero-crossing contours.

\begin{figure}[H]
\centering
\includegraphics[width=\textwidth]{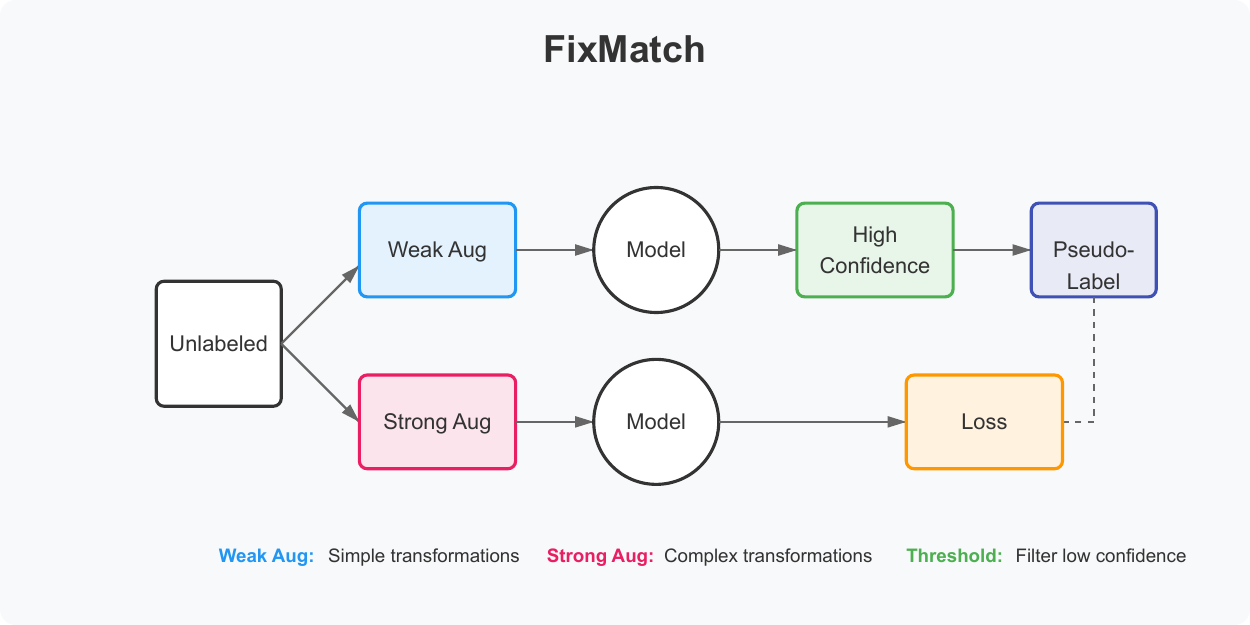}
\caption{\textit{FixMatch} a semi-supervised learning method that generates pseudo-labels from weakly-augmented unlabeled images and trains the model to predict these labels from strongly-augmented versions of the same images.}
\label{FixMatch}
\end{figure}

\newpage
1-1-n.\textit{ FixMatch} \cite{sohn2020fixmatch} a streamlined yet highly effective semi-supervised learning (SSL) algorithm that utilizes high-confidence pseudo-labels derived from weakly-augmented images to guide training on their strongly-augmented counterparts, as shown in \cref{FixMatch}. Despite its simplicity, FixMatch attains state-of-the-art performance across various SSL benchmarks, underscoring its efficacy even with extremely limited labelled data.

\begin{figure}[H]
\centering
\includegraphics[width=\textwidth]{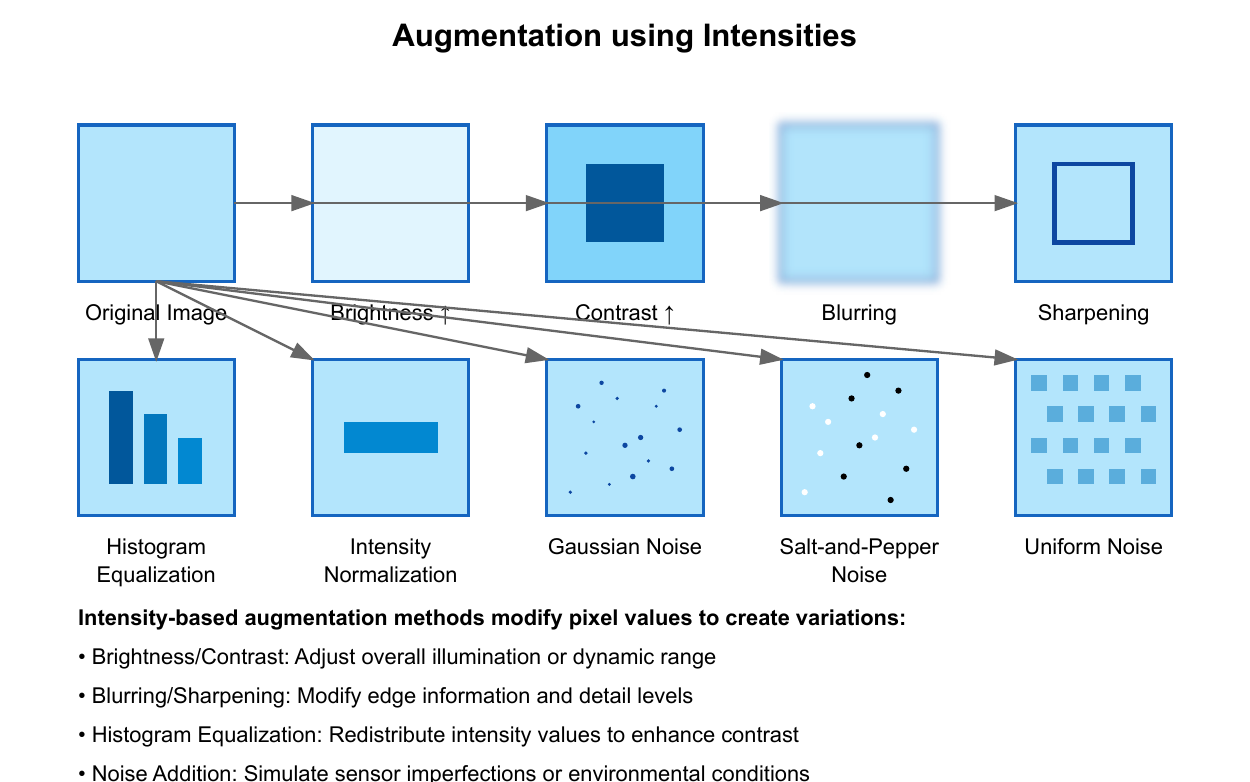}
\caption{\textit{Augmentation using intensities} transforms images by modifying pixel values through brightness/contrast adjustments, blurring, normalization, histogram equalization, sharpening, and the addition of various noise patterns.}
\label{Intensities}
\end{figure}

\newpage
1-1-o.\textit{Augmentation using intensities} \cite{goceri2023medical} involves modifying image properties such as contrast, brightness, blurring, sharpening, or adding noise to enhance model robustness against variations in image quality, as shown in \cref{Intensities}. Different types of noise, such as Gaussian, salt-and-pepper, or uniform, are applied by altering pixel values through specific distribution-based sampling methods, simulating diverse imaging conditions.

\begin{figure}[H]
\centering
\includegraphics[width=\textwidth]{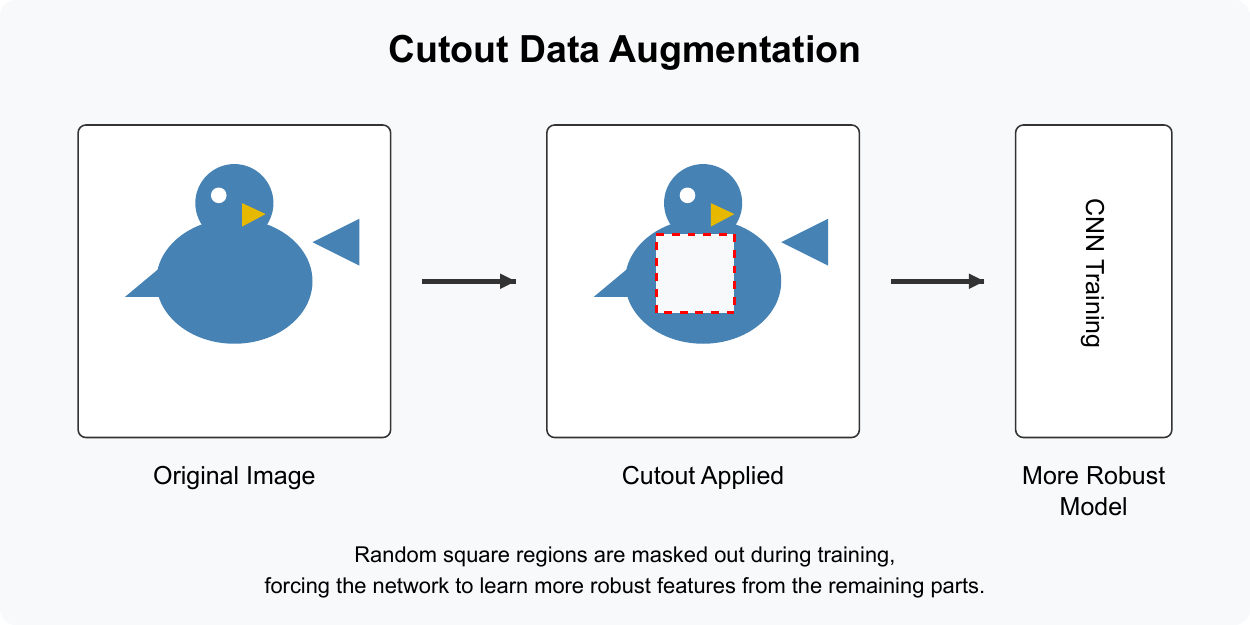}
\caption{\textit{cutout} data augmentation method randomly masks square regions of input images during training to improve CNN robustness and prevent overfitting.}
\label{cutout}
\end{figure}

\newpage
1-2-a.\textit{ cutout} \cite{devries2017improved} is a simple yet effective regularization technique that involves randomly masking out square regions of input during training, helping to improve the robustness and generalization of convolutional neural networks (CNNs), as shown in \cref{cutout}. By addressing the overfitting issues often faced by CNNs due to their large model capacity, cutout enhances performance and can be seamlessly combined with existing data augmentation methods and regularizers.

\begin{figure}[H]
\centering
\includegraphics[width=\textwidth]{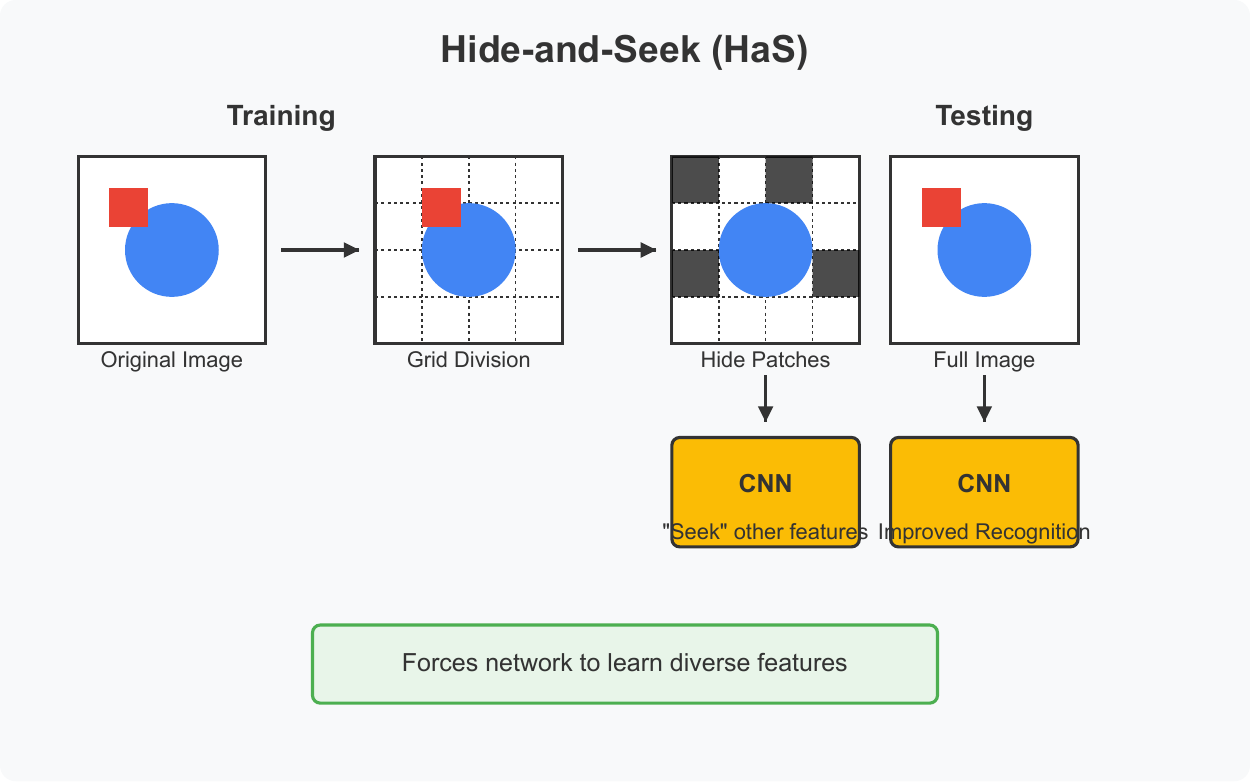}
\caption{\textit{Hide-and-Seek (HaS)} randomly masks image patches during training to force neural networks to learn diverse features beyond the most discriminative regions.}
\label{Hide-and-Seek (HaS)}
\end{figure}

\newpage
1-2-b.\textit{ Hide-and-Seek (HaS)} \cite{singh2018hide} is a versatile data augmentation technique that improves visual recognition tasks by randomly hiding patches in training images, encouraging the network to focus on other relevant features when key discriminative regions are obscured, as shown in \cref{Hide-and-Seek (HaS)}. This method is compatible with any network, requires no modifications during testing, and demonstrates significant improvements in tasks such as object localization, image classification, and semantic segmentation.

\begin{figure}[H]
\centering
\includegraphics[width=\textwidth]{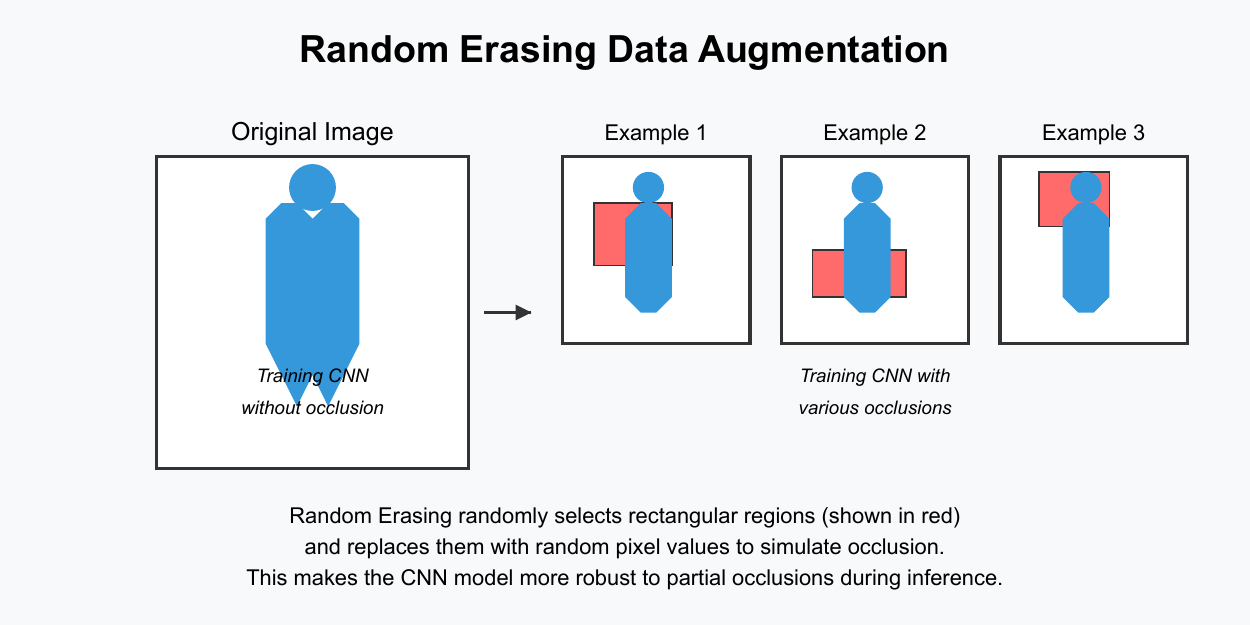}
\caption{\textit{Random Erasing} augments training data by randomly erasing rectangular regions of images with random pixel values to improve CNN robustness to occlusion.}
\label{Random Erasing}
\end{figure}

\newpage
1-2-c.\textit{ Random Erasing} \cite{zhong2020random} is a simple yet effective data augmentation technique that improves the robustness of convolutional neural networks (CNNs) by randomly selecting and erasing rectangular regions in training images with random pixel values, as shown in \cref{Random Erasing}. This method, which is parameter-free and compatible with most CNN-based recognition models, enhances performance across tasks such as image classification, object detection, and person re-identification by reducing overfitting and increasing occlusion robustness.

\begin{figure}[H]
\centering
\includegraphics[width=\textwidth]{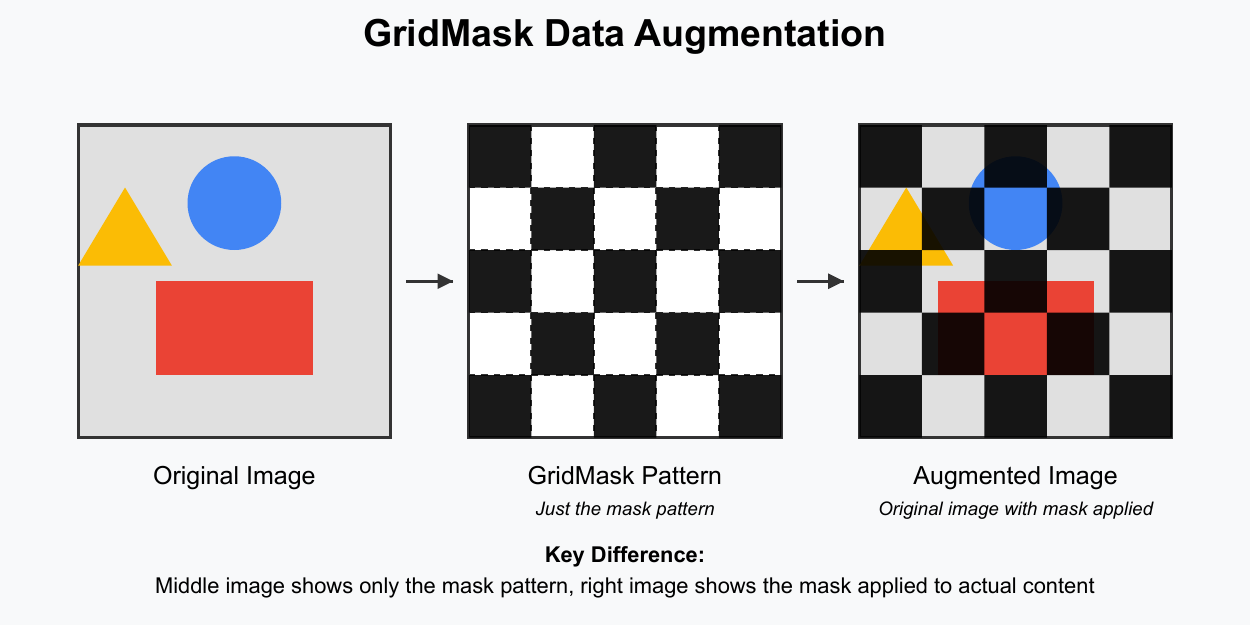}
\caption{\textit{GridMask} systematically removes information in a grid pattern to enhance model robustness by forcing neural networks to learn from partially obscured images.}
\label{GridMask}
\end{figure}

\newpage
1-2-d.\textit{ GridMask} \cite{chen2020gridmask} is a novel data augmentation method that improves performance in various computer vision tasks by systematically removing regions of input images to enhance model robustness, as shown in \cref{GridMask}. Compared to existing methods like AutoAugment, GridMask is computationally efficient and achieves superior results on benchmarks such as ImageNet, COCO2017, and Cityscapes.

\begin{figure}[H]
\centering
\includegraphics[width=\textwidth]{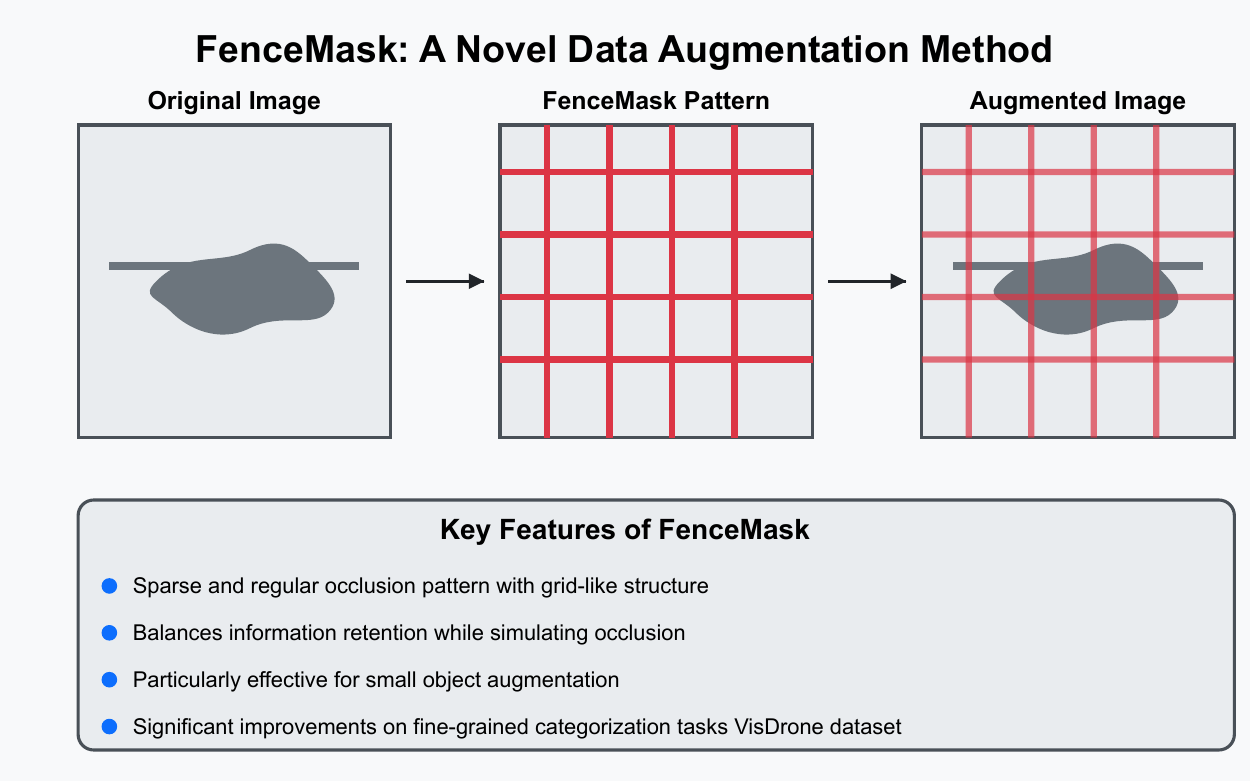}
\caption{\textit{FenceMask} a grid-based occlusion data augmentation method that balances information retention with object occlusion for improved performance in computer vision tasks.}
\label{FenceMask}
\end{figure}

\newpage
1-2-e.\textit{ FenceMask} \cite{li2020fencemask} is a novel data augmentation method that simulates object occlusion to balance occlusion and information retention, achieving superior performance in various computer vision tasks, as shown in \cref{FenceMask}. Extensive experiments on datasets like CIFAR, ImageNet, COCO2017, and VisDrone demonstrate its effectiveness, particularly in fine-grained visual categorization and small object detection.

\begin{figure}[H]
\centering
\includegraphics[width=\textwidth]{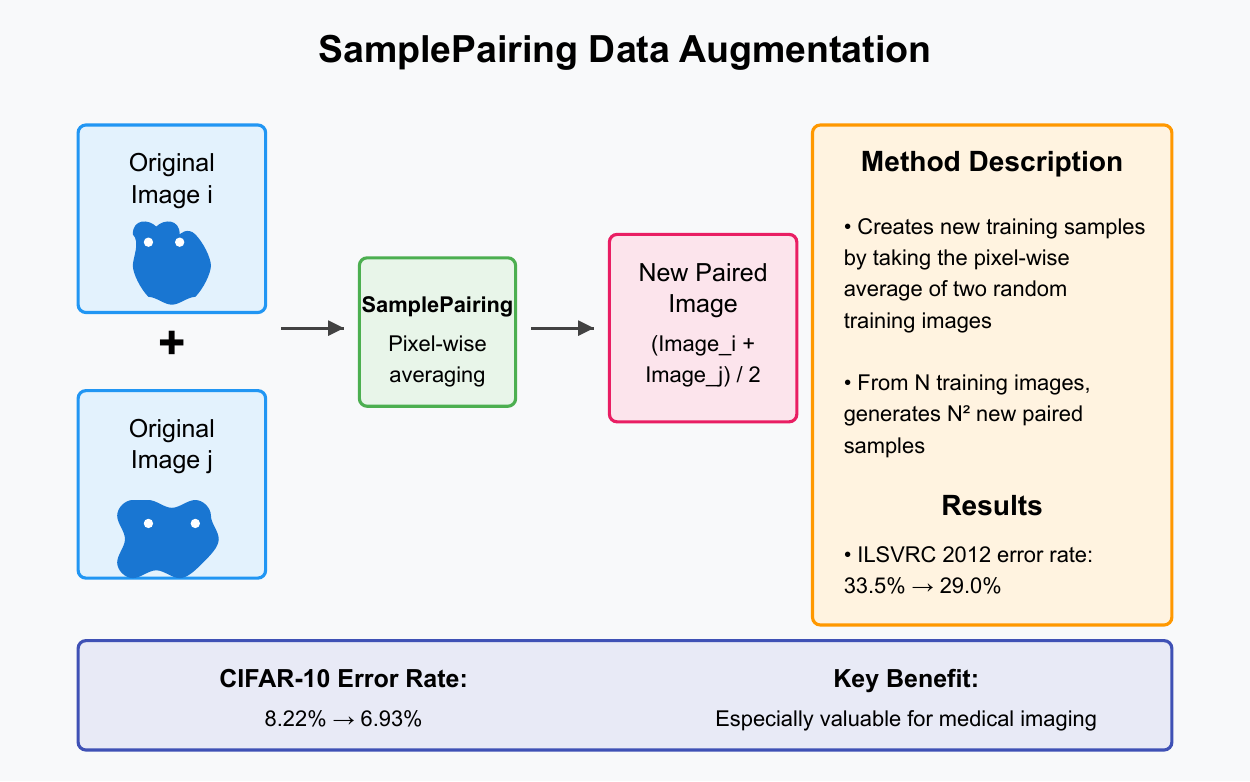}
\caption{\textit{Pairing Samples} data augmentation method creates new training samples by pixel-wise averaging of randomly selected image pairs, significantly improving classification accuracy across datasets.}
\label{Pairing Samples}
\end{figure}

\newpage
1-3-a.\textit{ Pairing Samples} \cite{inoue2018data} is a novel data augmentation technique for image classification that creates new training samples by overlaying two randomly selected images from the dataset, resulting in a significant increase in the diversity of training data, as shown in \cref{Pairing Samples}. This method has shown substantial improvements in classification accuracy across various datasets and proves particularly effective in tasks with limited training data, such as medical imaging.

\begin{figure}[H]
\centering
\includegraphics[width=\textwidth]{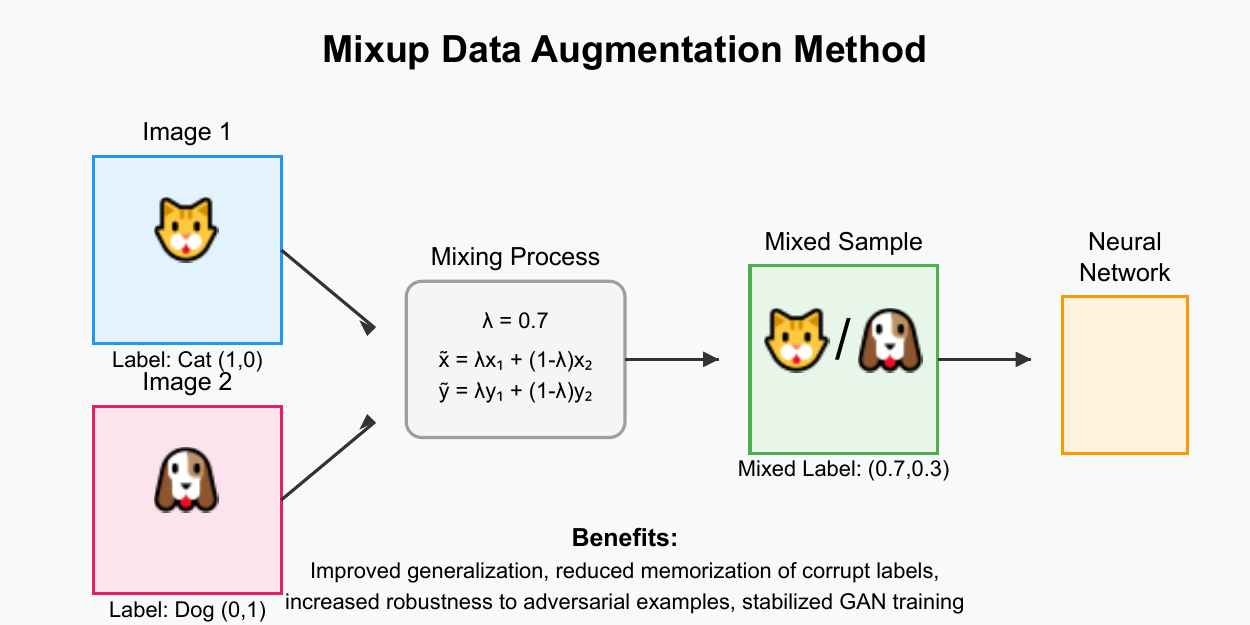}
\caption{\textit{Mixup} augments training data by linearly interpolating pairs of images and their labels with a mixing parameter \ensuremath{\lambda} to improve model generalization and robustness.}
\label{mixup}
\end{figure}

\newpage
1-3-b.\textit{ Mixup} \cite{zhang2017mixup} is a data augmentation technique that trains neural networks on convex combinations of pairs of examples and their corresponding labels, encouraging linear behavior between training samples, as shown in \cref{mixup}. This method improves generalization, enhances robustness to adversarial examples, reduces label memorization, and stabilizes the training of generative adversarial networks across various datasets and architectures.

\begin{figure}[H]
\centering
\includegraphics[width=\textwidth]{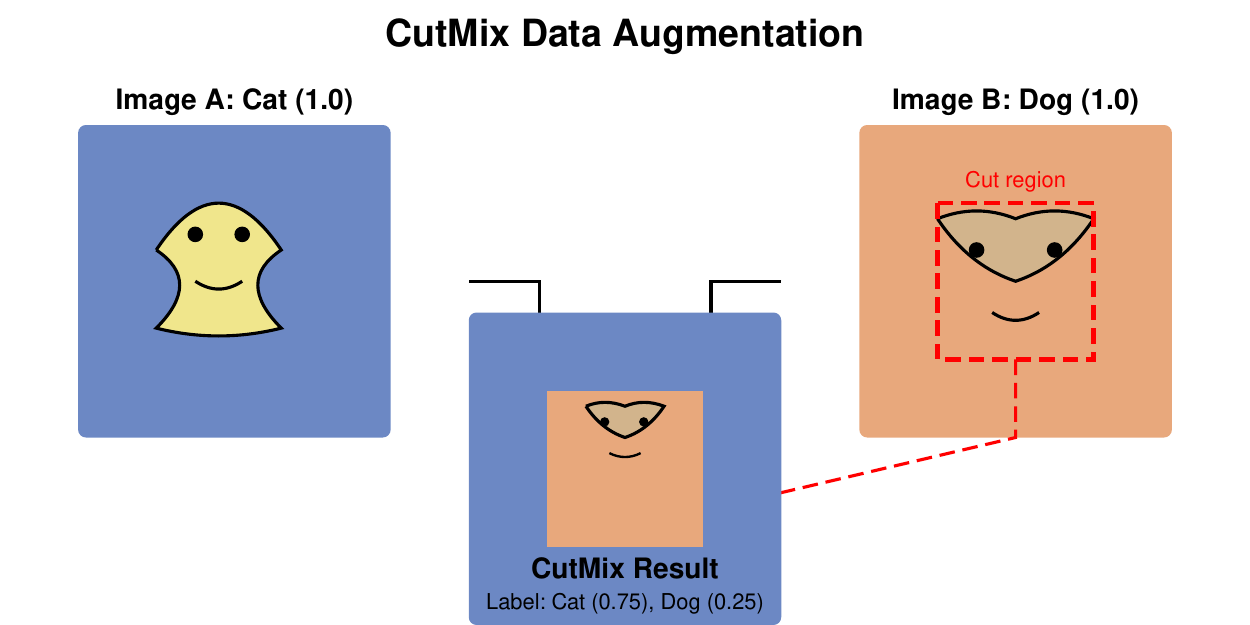}
\caption{\textit{CutMix} augmentation creates new training examples by cutting regions from one image and pasting them onto another while proportionally mixing their class labels.}
\label{cutmix}
\end{figure}

\newpage
1-3-c.\textit{ CutMix} \cite{yun2019cutmix} is a data augmentation technique that enhances model generalization by replacing regions of an image with patches from other images while proportionally mixing their labels, as shown in \cref{cutmix}. Unlike traditional regional dropout methods that discard informative pixels, CutMix retains useful information, leading to improved classification, object localization, and robustness across multiple benchmarks.

\begin{figure}[H]
\centering
\includegraphics[width=\textwidth]{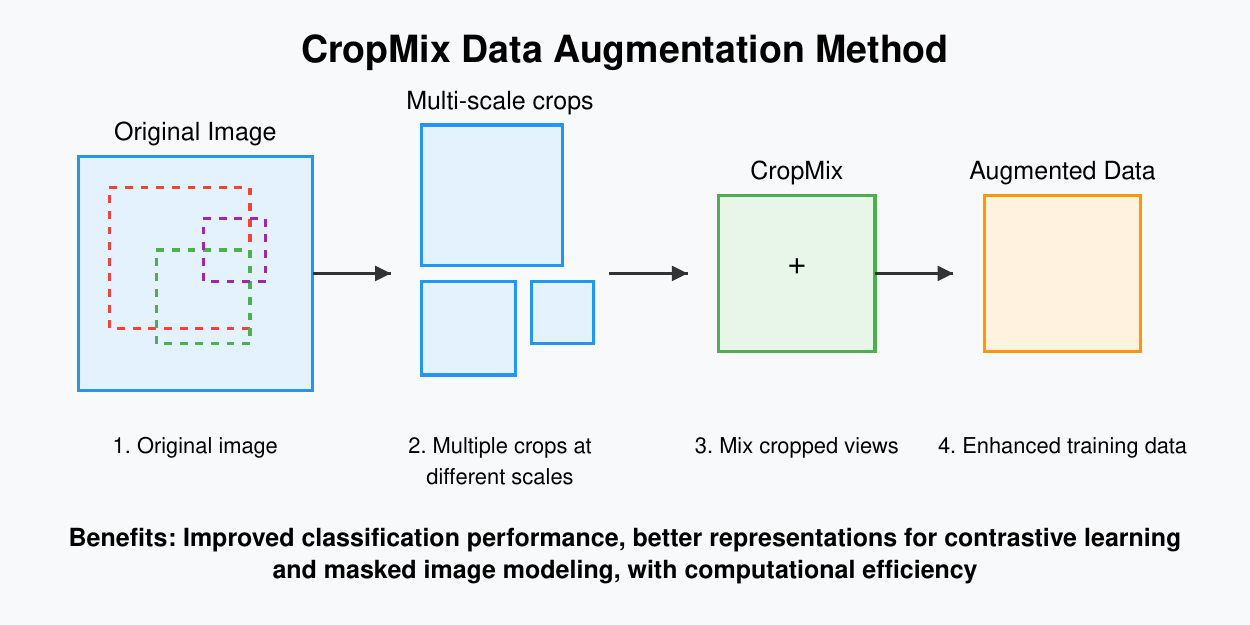}
\caption{\textit{CropMix} augments training data by mixing multiple crops at different scales from the same image to capture richer features and improve model performance.}
\label{cropmix}
\end{figure}

\newpage
1-3-d.\textit{ CropMix} \cite{han2022cropmix} is a novel data augmentation method designed to generate a rich input distribution by cropping an image multiple times at distinct scales, ensuring the capture of multi-scale information, as shown in \cref{cropmix}. By mixing these cropped views to form new training data, CropMix enhances performance across diverse vision tasks, including classification, contrastive learning, and masked image modeling, while maintaining computational efficiency.

\begin{figure}[H]
\centering
\includegraphics[width=\textwidth]{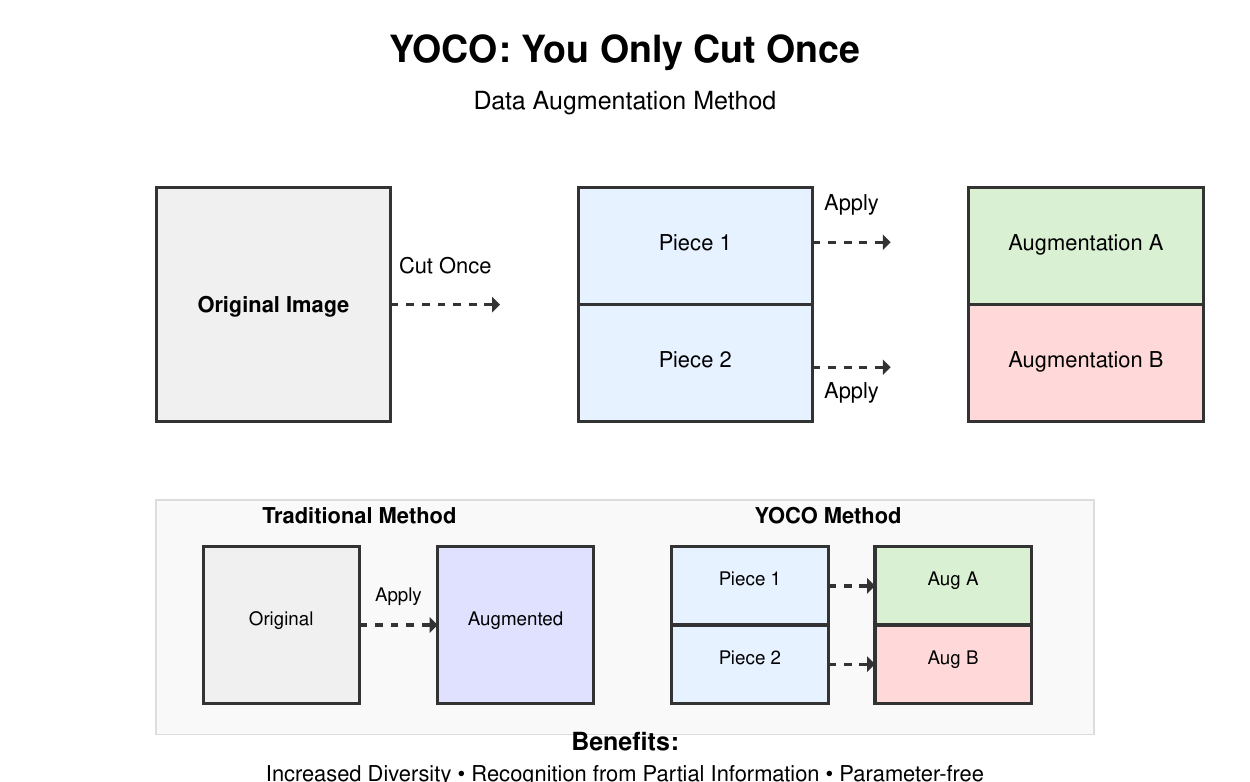}
\caption{\textit{YOCO} This schematic representation illustrates the You Only Cut Once (YOCO) data augmentation technique, which segments input images for independent transformations, thereby enhancing sample diversity and enabling neural networks to recognize objects from partial information with demonstrated performance improvements across multiple architectures.}
\label{yoco}
\end{figure}

\newpage
1-3-e.\textit{ YOCO} \cite{han2022you} (You Only Cut Once) is a novel data augmentation technique that partitions an image into two segments and applies augmentations independently to each segment, as shown in \cref{yoco}. YOCO enhances augmentation diversity, facilitates object recognition from partial information, and achieves significant performance improvements across various neural network architectures and tasks, including CIFAR and ImageNet classification as well as contrastive pre-training.

\begin{figure}[H]
\centering
\includegraphics[width=\textwidth]{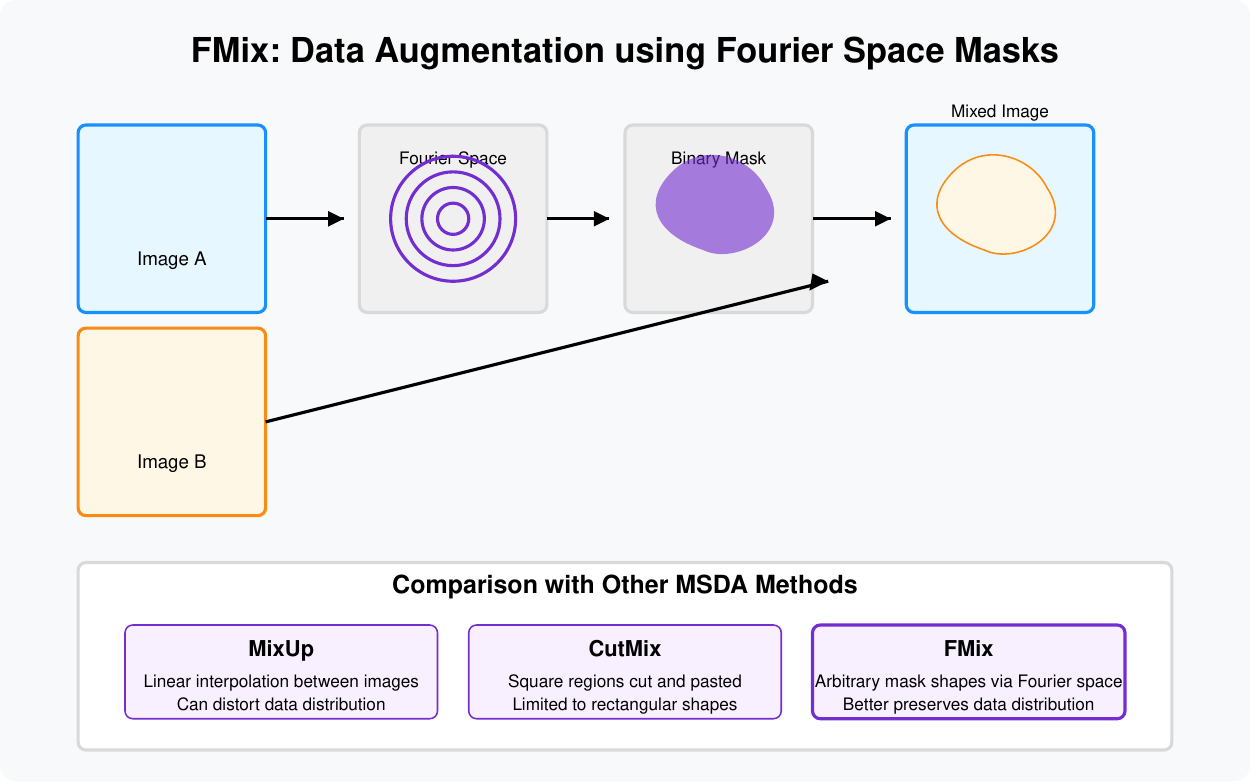}
\caption{\textit{FMix} generates augmented training samples by combining image pairs using organically-shaped binary masks derived from low-frequency Fourier space sampling, offering advantages over conventional MixUp and CutMix approaches.}
\label{FMix}
\end{figure}

\newpage
1-3-f.\textit{ FMix} \cite{harris2020fmix} is a novel Mixed Sample Data Augmentation (MSDA) technique that employs random binary masks, generated from low-frequency images in Fourier space, to create augmented samples with diverse shapes, as shown in \cref{FMix}. Building on the strengths of CutMix, FMix improves generalization and prevents memorization without distorting the data distribution, achieving superior performance across various datasets and tasks, including state-of-the-art results on CIFAR-10.

\begin{figure}[H]
\centering
\includegraphics[width=\textwidth]{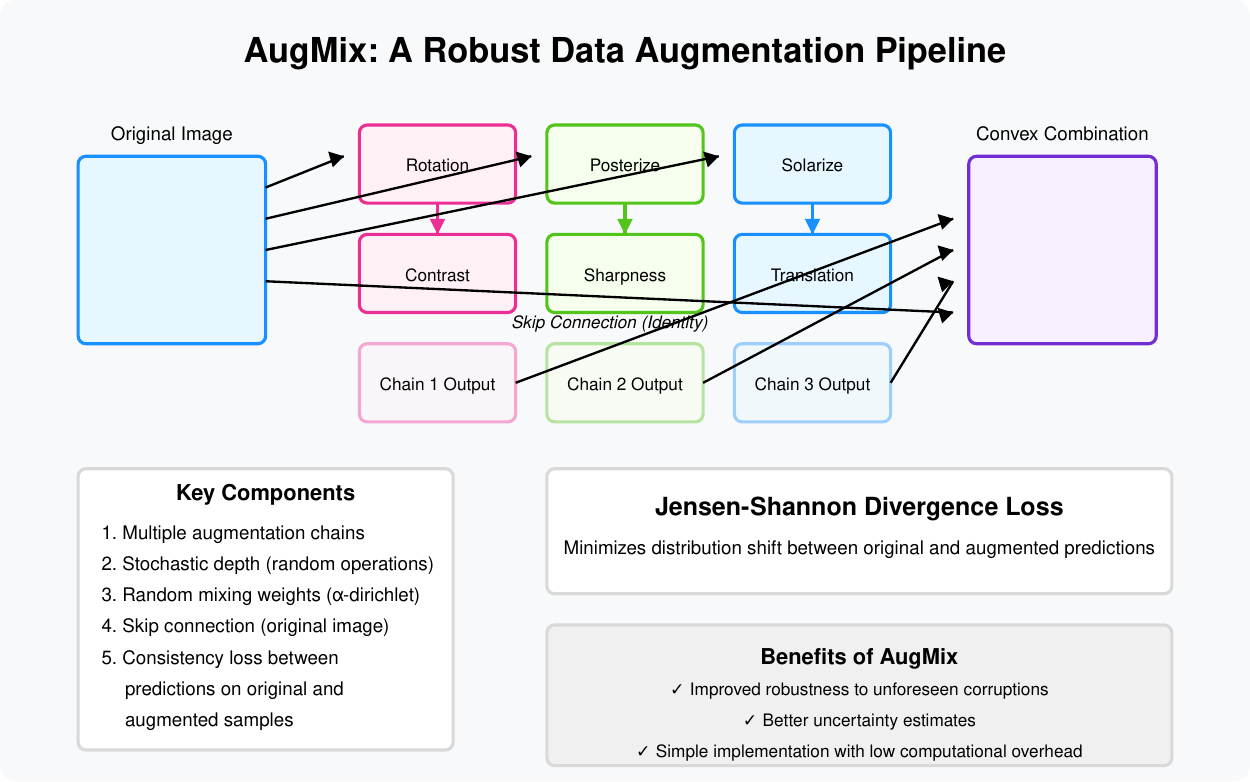}
\caption{\textit{AugMix} enhances model robustness through a combination of diverse augmentation chains, stochastic operations, and consistency regularization via Jensen-Shannon divergence loss between original and augmented predictions.}
\label{AugMix}
\end{figure}

\newpage
1-3-g.\textit{ AugMix} \cite{hendrycks2019augmix} is a data augmentation technique developed to improve the robustness and uncertainty estimates of image classifiers by addressing the challenges posed by unforeseen data distribution changes, as shown in \cref{AugMix}. This method is computationally efficient, straightforward to implement and has demonstrated substantial improvements in robustness and uncertainty measures on challenging image classification benchmarks, significantly narrowing the gap to optimal performance in certain scenarios.

\begin{figure}[H]
\centering
\includegraphics[width=\textwidth]{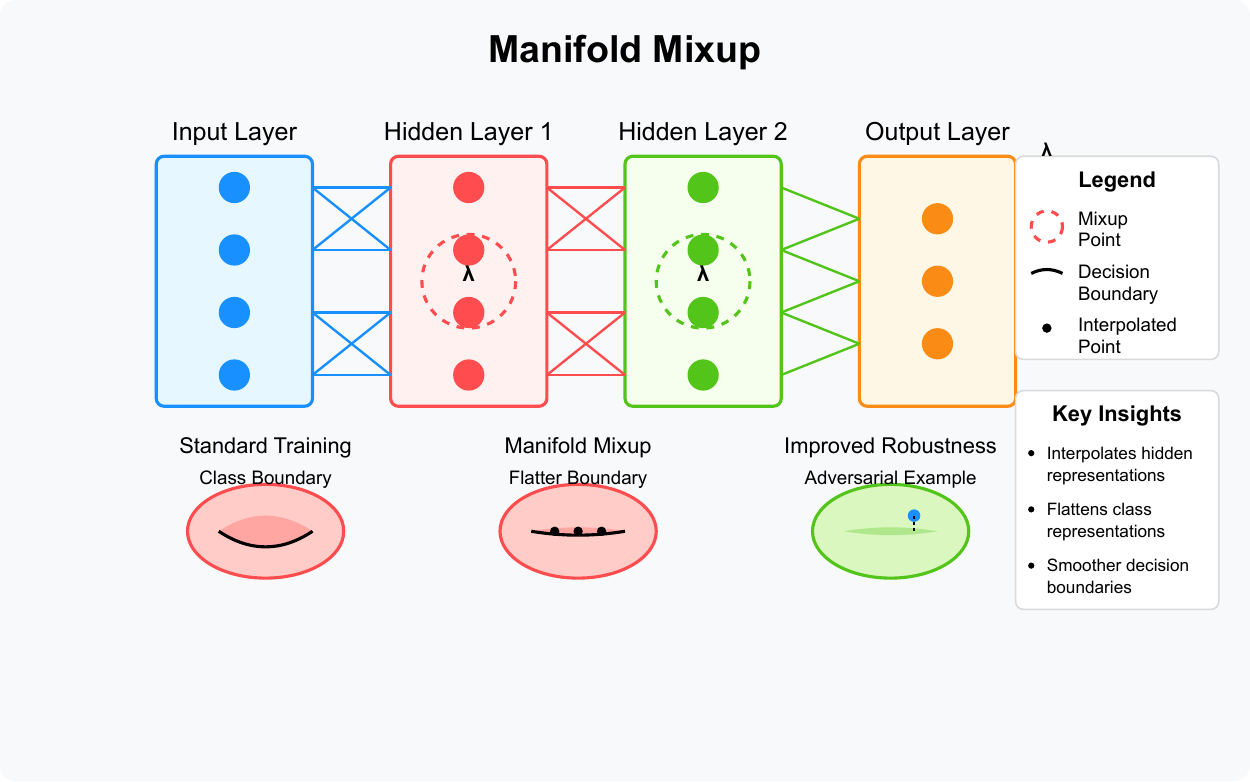}
\caption{\textit{ManifoldMix} interpolates hidden layer representations to create smoother decision boundaries and improve model robustness against adversarial examples.}
\label{ManifoldMix}
\end{figure}

\newpage
1-3-h.\textit{ ManifoldMix} \cite{verma2019manifold} is a data augmentation method that improves the generalization and robustness of deep neural networks by performing interpolations in hidden representations of the network, as shown in \cref{ManifoldMix}. This technique encourages smoother decision boundaries and flatter class representations, leading to improved performance on supervised learning tasks, robustness against adversarial attacks, and better test log-likelihood, all with minimal computational overhead.

\begin{figure}[H]
\centering
\includegraphics[width=\textwidth]{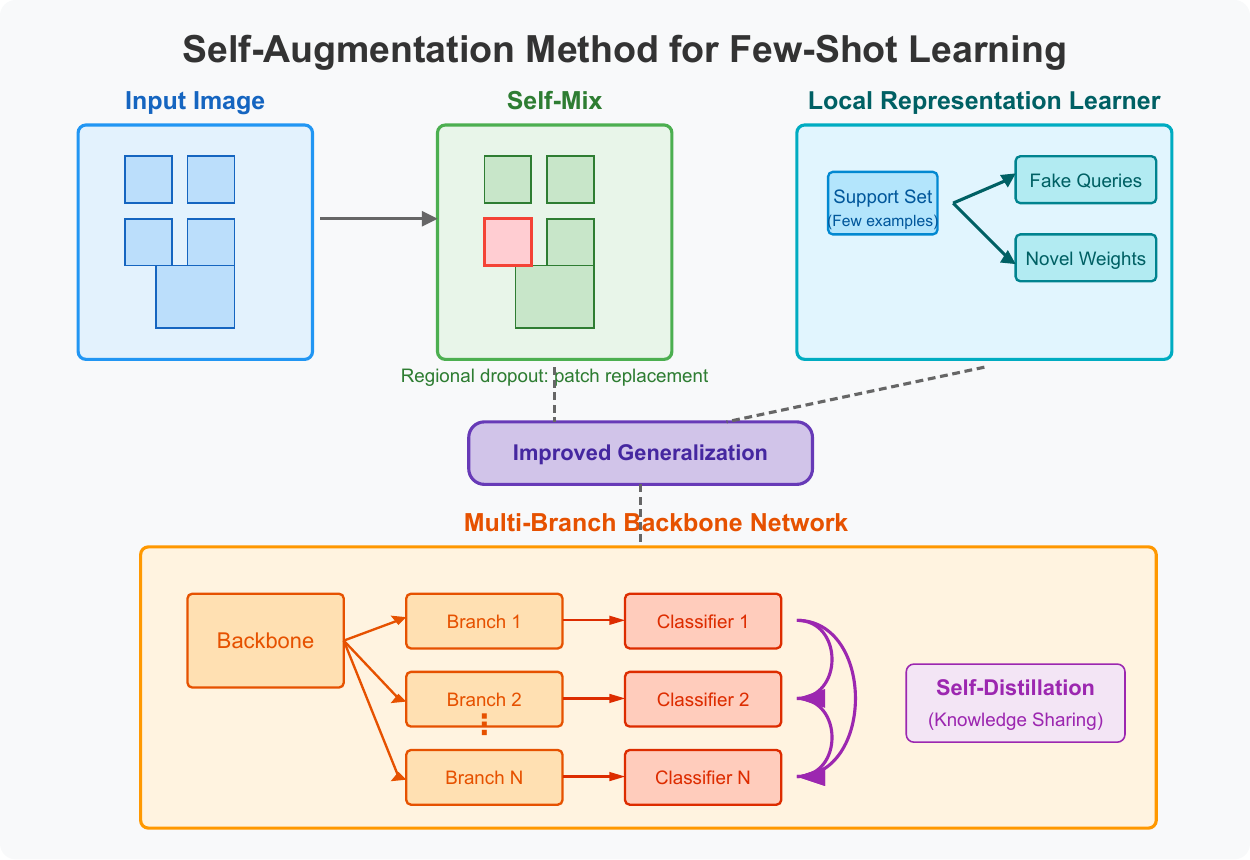}
\caption{\textit{Self-Augmentation} combines regional dropout (Self-Mix) with knowledge sharing across network branches (Self-Distillation) to improve few-shot learning generalization.}
\label{Self-Augmentation}
\end{figure}

\newpage
1-3-i.\textit{ Self-Augmentation} \cite{seo2021self} is a novel approach to learning in a few shots that combines self-mix, a regional dropout technique that replaces image patches with values of the same image, and self-distillation to enhance generalization by preventing overfitting to dataset-specific structures, as shown in \cref{Self-Augmentation}. By incorporating auxiliary branches for knowledge sharing and a local representation learner to generate synthetic queries and novel class weights, this method achieves state-of-the-art performance on standard few-shot learning benchmarks.

\begin{figure}[H]
\centering
\includegraphics[width=\textwidth]{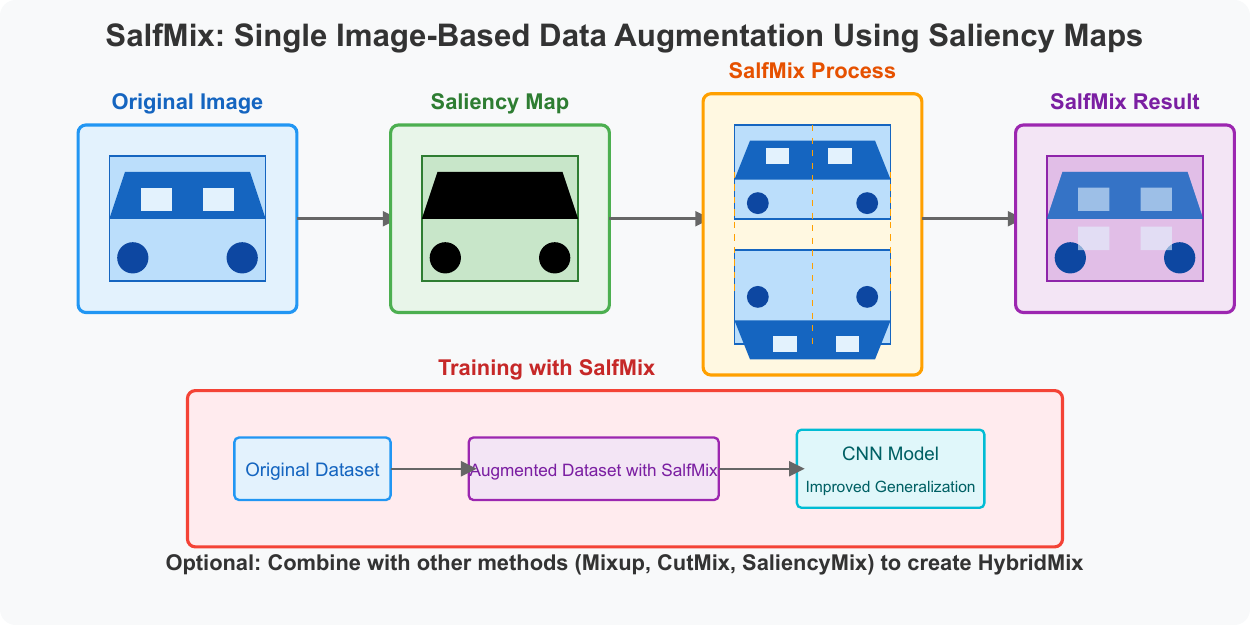}
\caption{\textit{SalfMix} The SalfMix technique uses saliency maps to create augmented training data by intelligently combining original and transformed image regions.}
\label{SalfMix}
\end{figure}

\newpage
1-3-j.\textit{ SalfMix} \cite{choi2021salfmix} is a novel data augmentation method that generates self-mixed images using saliency maps, focusing on improving generalizability by applying mixing strategies to a single image, as shown in \cref{SalfMix}. Additionally, the combination of SalfMix with state-of-the-art two-image augmentation methods, termed HybridMix, achieves superior performance on multiple classification and object detection benchmarks, demonstrating its effectiveness and versatility.

\begin{figure}[H]
\centering
\includegraphics[width=\textwidth]{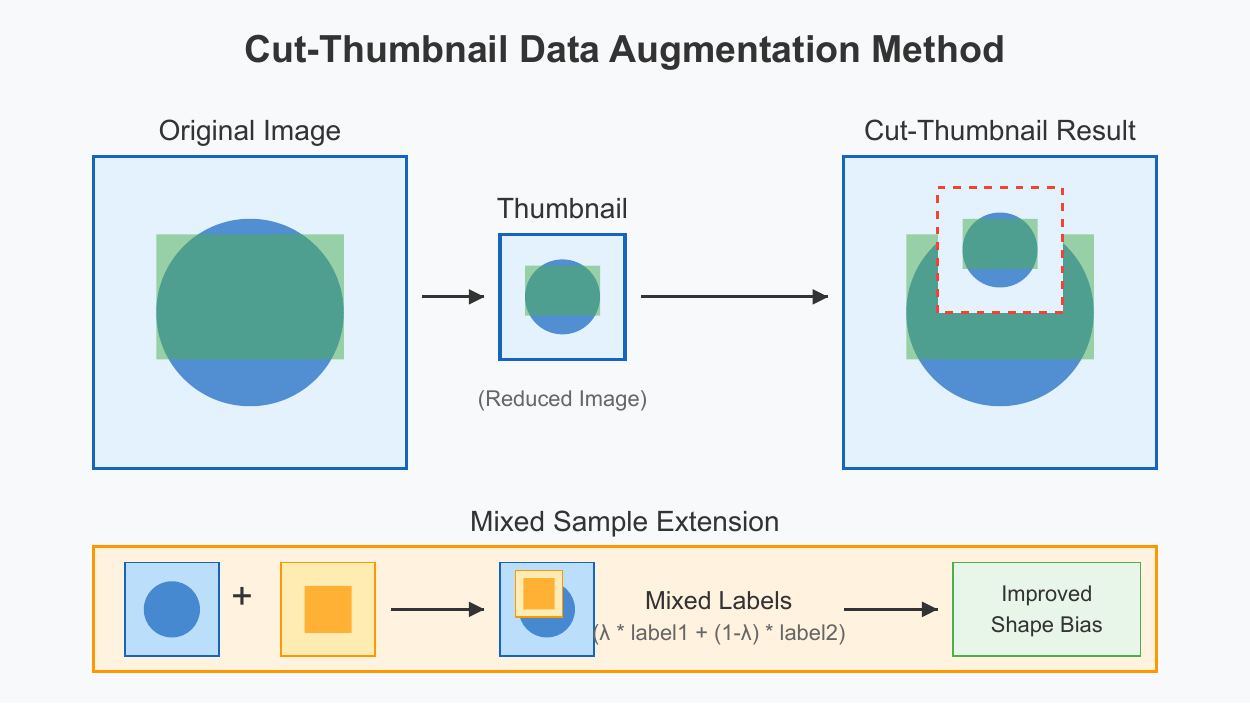}
\caption{\textit{Cut-Thumbnail} data augmentation Method diagram showing how images are processed into reduced thumbnails and reinserted to enhance shape bias in machine learning models. }
\label{Cut-Thumbnail}
\end{figure}

\newpage
1-3-k.\textit{ Cut-Thumbnail} \cite{xie2021cut} a novel data augmentation strategy that embeds a reduced version of an image (thumbnail) into a random region of the original image, improving both local and global information, as shown in \cref{Cut-Thumbnail}. This method achieves significant performance gains across various tasks, with ResNet-50 on ImageNet improving by over 2.8\%, reaching 79.21\% top-1 accuracy.

\begin{figure}[H]
\centering
\includegraphics[width=\textwidth]{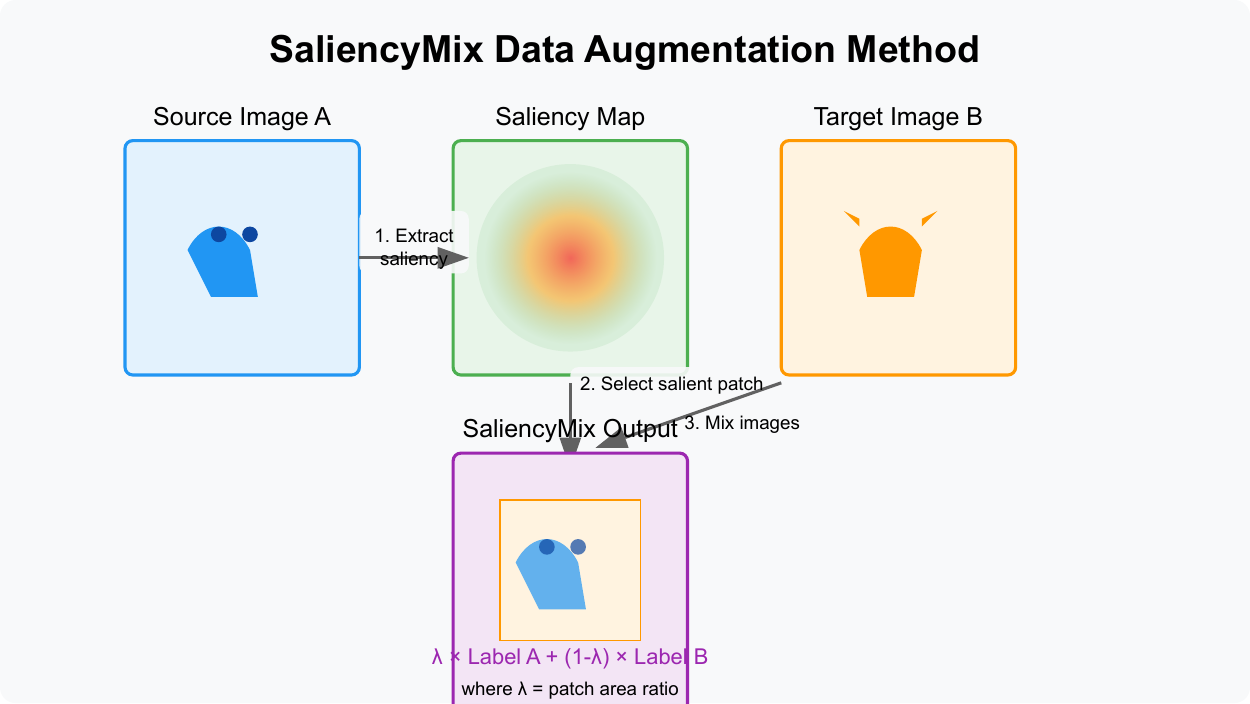}
\caption{\textit{SaliencyMix} augmentation identifies salient regions in a source image and transfers them to a target image. The resulting mixed image is paired with a weighted combination of both labels, proportional to the area of the transferred patch.}
\label{SALIENCYMIX}
\end{figure}

\newpage
1-3-l.\textit{ SaliencyMix} \cite{uddin2020saliencymix} a novel data augmentation strategy that leverages saliency maps to select representative image patches, which are then mixed with target images to guide the model toward learning more meaningful feature representations, as shown in \cref{SALIENCYMIX}. SaliencyMix achieves state-of-the-art performance on ImageNet classification, reducing top-1 error to 21.26\% and 20.09\% for ResNet-50 and ResNet-101, respectively, while also enhancing model robustness and object detection performance.

\begin{figure}[H]
\centering
\includegraphics[width=\textwidth]{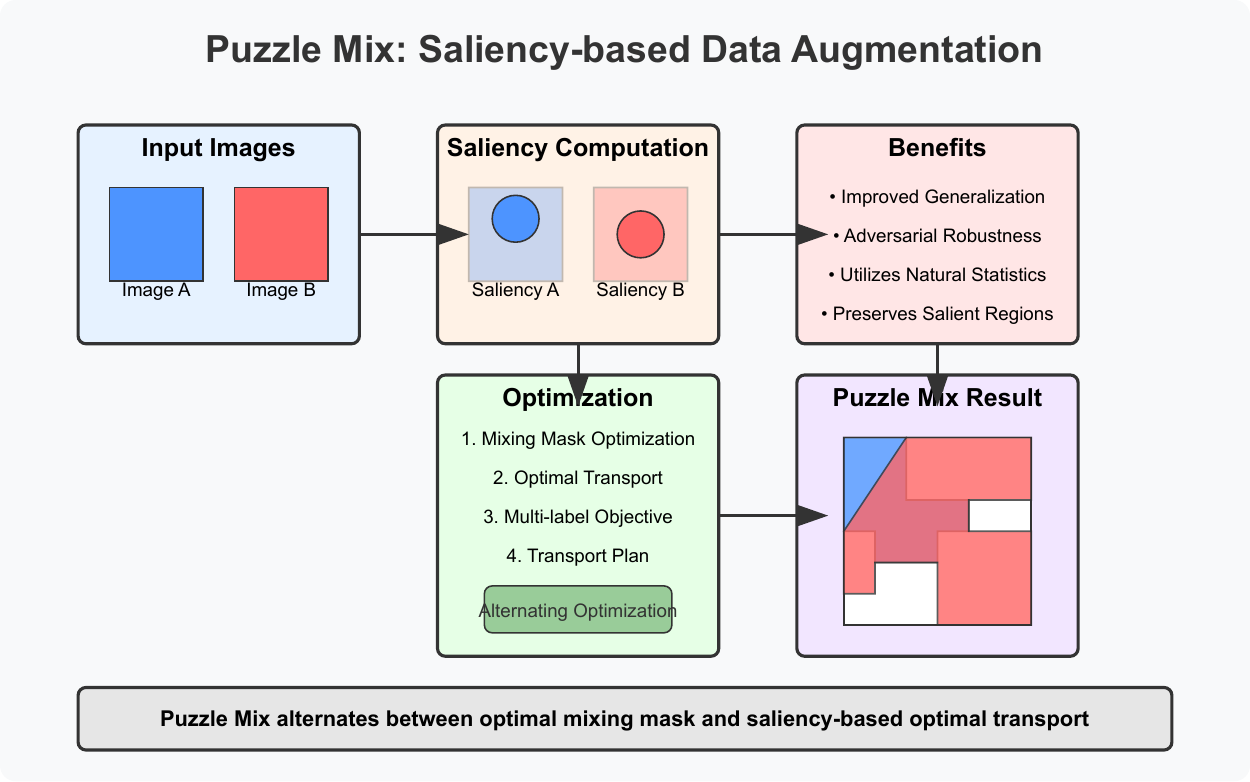}
\caption{\textit{Puzzle Mix} intelligently combines images by preserving salient regions through alternating optimization of mixing masks and optimal transport to enhance model generalization and adversarial robustness.}
\label{Puzzle Mix}
\end{figure}

\newpage
1-3-m.\textit{ Puzzle Mix} \cite{kim2020puzzle} a novel mixup-based data augmentation method that explicitly incorporates saliency information and underlying data statistics to generate more informative virtual examples, as shown in \cref{Puzzle Mix}. By optimizing a multi-label objective and a saliency-discounted optimal transport objective, Puzzle Mix achieves state-of-the-art generalization and adversarial robustness on CIFAR-100, Tiny-ImageNet, and ImageNet datasets.

\begin{figure}[H]
\centering
\includegraphics[width=\textwidth]{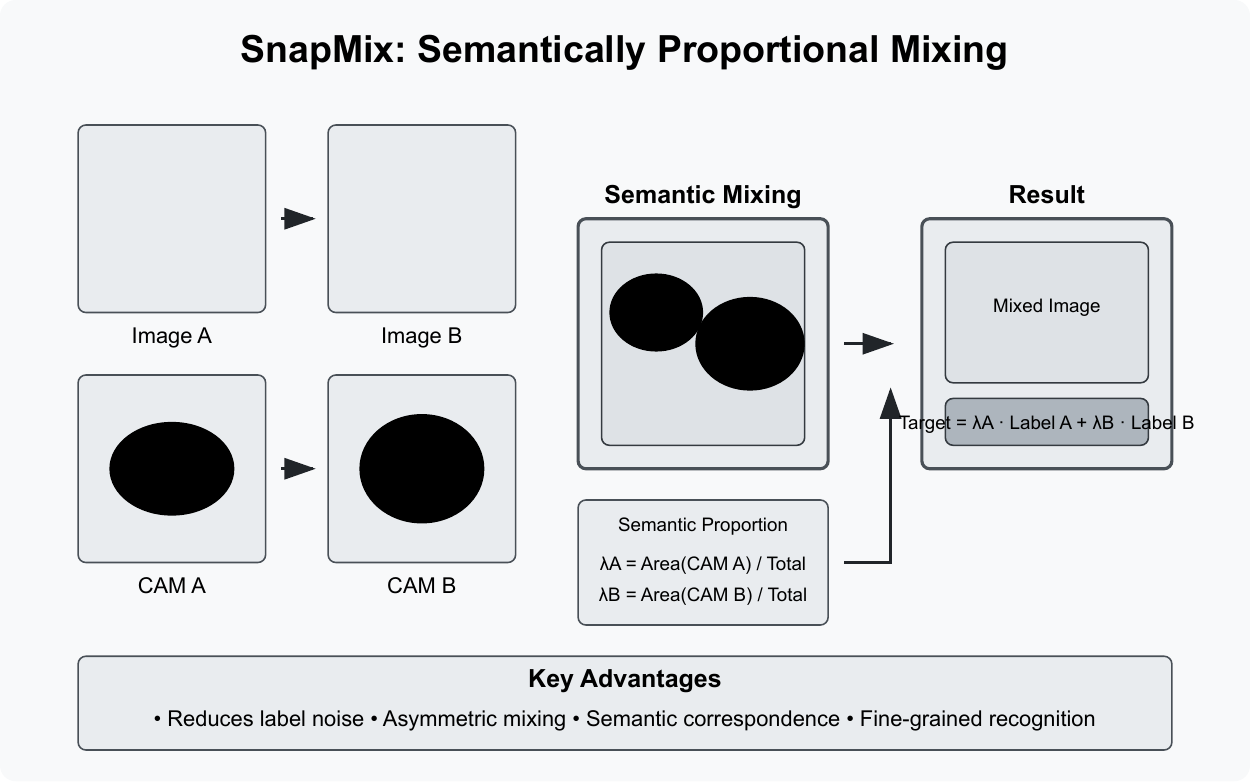}
\caption{\textit{SnapMix} a data augmentation technique that uses Class Activation Maps to create semantically proportional mixed images with corresponding mixed labels for improved fine-grained recognition.}
\label{SnapMix}
\end{figure}

\newpage
1-3-n.\textit{ SnapMix} \cite{huang2021snapmix} a semantically proportional data mixing augmentation method that leverages class activation maps (CAM) to reduce label noise in fine-grained recognition by ensuring semantic correspondence between mixed images and target labels, as shown in \cref{SnapMix}. SnapMix consistently outperforms existing mix-based approaches across various datasets and achieves state-of-the-art performance in fine-grained recognition tasks.

\begin{figure}[H]
\centering
\includegraphics[width=\textwidth]{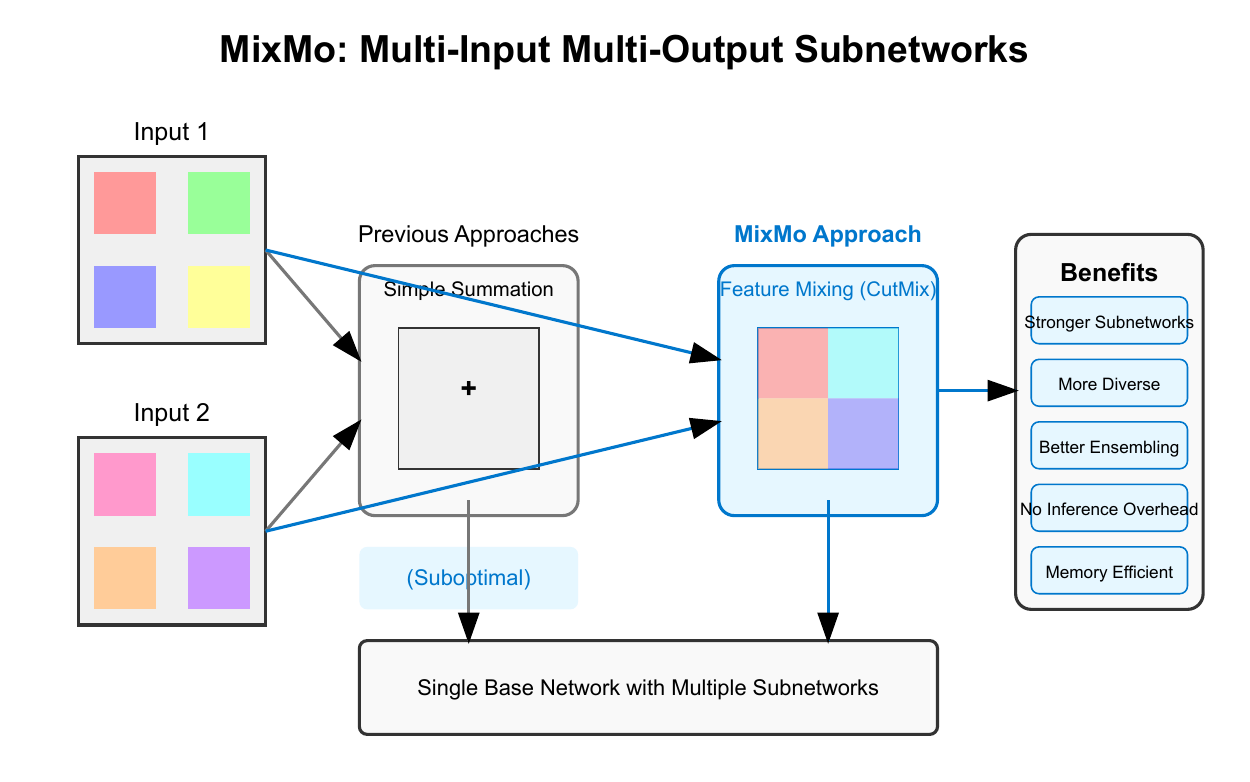}
\caption{\textit{MixMo} framework for neural networks, showing how it improves on previous approaches by using feature mixing (CutMix) between multiple inputs to create stronger, more diverse subnetworks with better performance.}
\label{MixMo}
\end{figure}

\newpage
1-3-o.\textit{ MixMo} \cite{rame2021mixmo} a generalized framework for training multi-input multi-output deep subnetworks by replacing the suboptimal summing operation in prior approaches with a more effective binary mixing mechanism inspired by mixed sample data augmentations, as shown in \cref{MixMo}. MixMo enhances subnetwork diversity and performance, achieving state-of-the-art results on CIFAR-100 and Tiny ImageNet, while outperforming data-augmented deep ensembles without additional inference or memory overhead.

\begin{figure}[H]
\centering
\includegraphics[width=\textwidth]{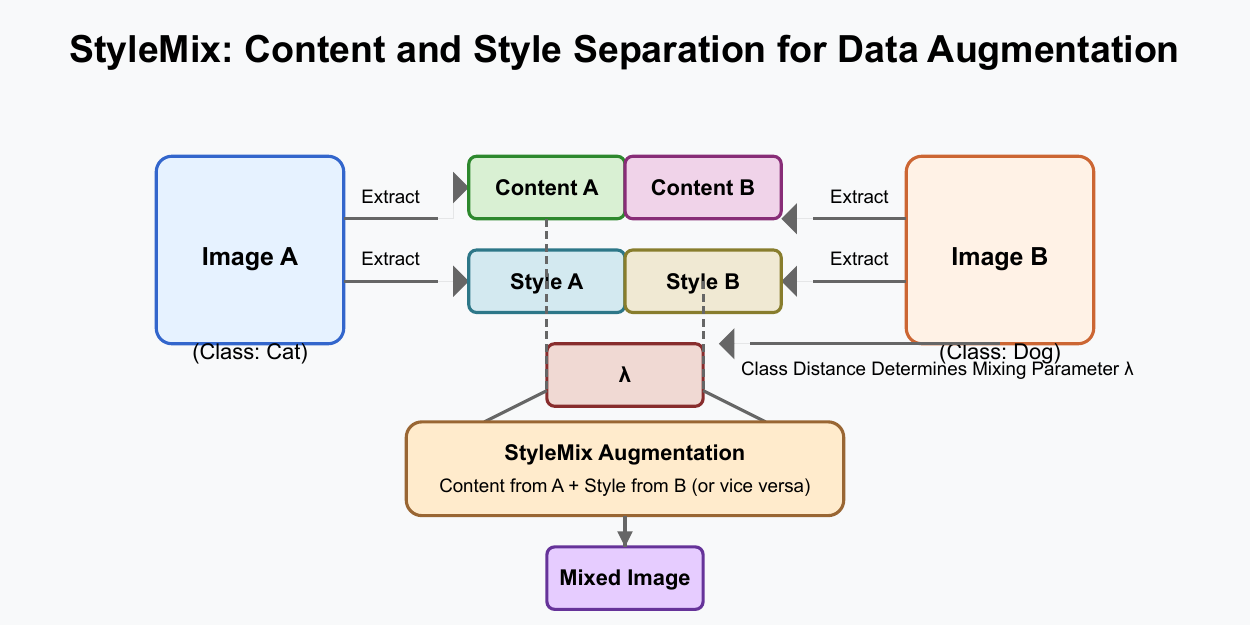}
\caption{\textit{StyleMix} data augmentation method separately extracts and recombines content and style features from image pairs based on class distance to create robust training samples.}
\label{StyleMix}
\end{figure}

\newpage
1-3-p.\textit{ StyleMix} \cite{hong2021stylemix} the first mixup-based augmentation method that separately manipulates the content and style features of input image pairs to create more diverse and robust training samples, as shown in \cref{StyleMix}. By dynamically adjusting the degree of style mixing based on class distance, StyleMix enhances model generalization and achieves competitive performance on CIFAR-100, CIFAR-10, and ImageNet, while improving robustness against adversarial attacks.

\begin{figure}[H]
\centering
\includegraphics[width=\textwidth]{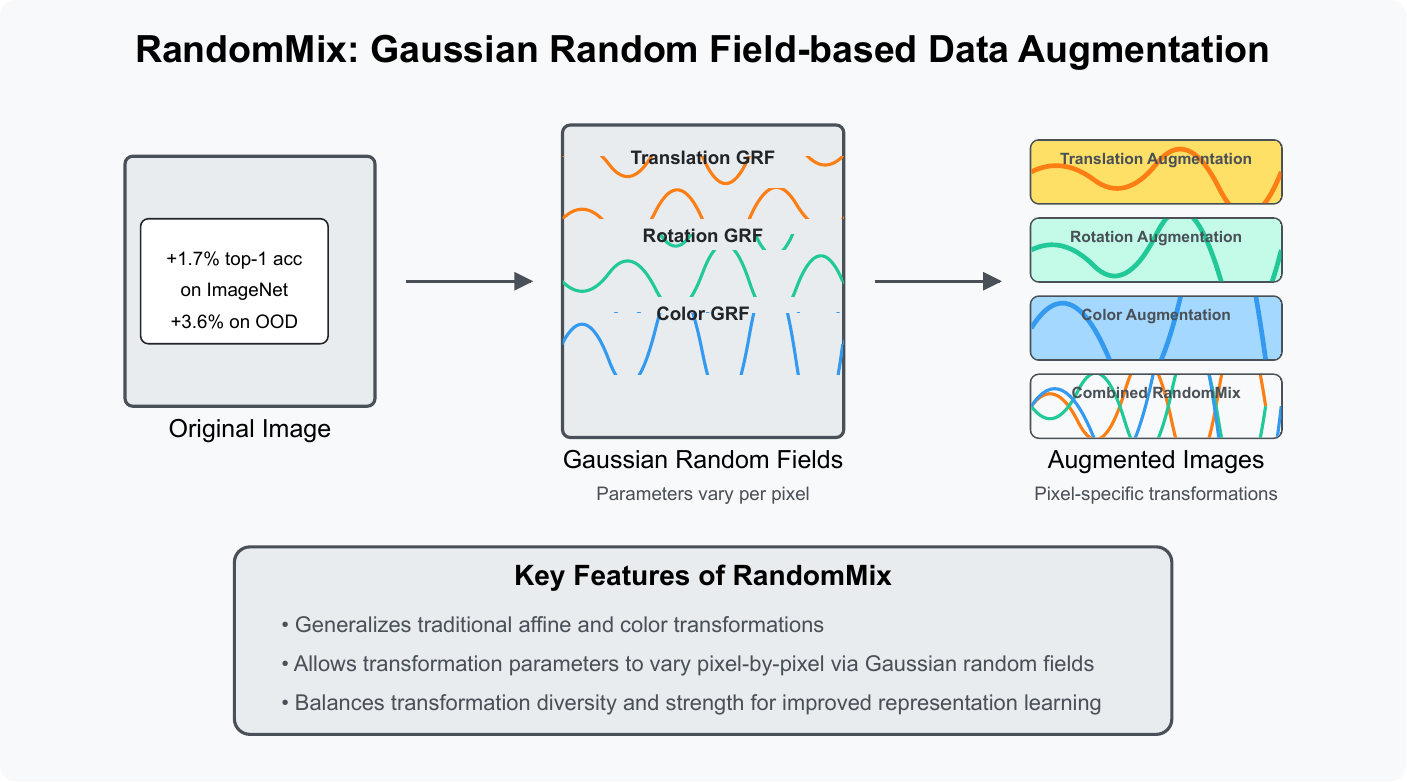}
\caption{\textit{RandomMix} generates diverse image augmentations by applying pixel-specific transformations via Gaussian random fields that generalize traditional affine and colour transformations.}
\label{RandomMix}
\end{figure}

\newpage
1-3-q.\textit{ RandomMix} \cite{mansfield2023random} a novel family of local transformations based on Gaussian random fields that significantly expand the augmentation space for self-supervised representation learning by allowing transformation parameters to vary across pixels, as shown in \cref{RandomMix}. RandomMix achieves notable improvements in downstream classification tasks, including a 1.7\% top-1 accuracy gain on ImageNet and a 3.6\% improvement on out-of-distribution iNaturalist, highlighting its effectiveness in enhancing representation learning.

\begin{figure}[H]
\centering
\includegraphics[width=\textwidth]{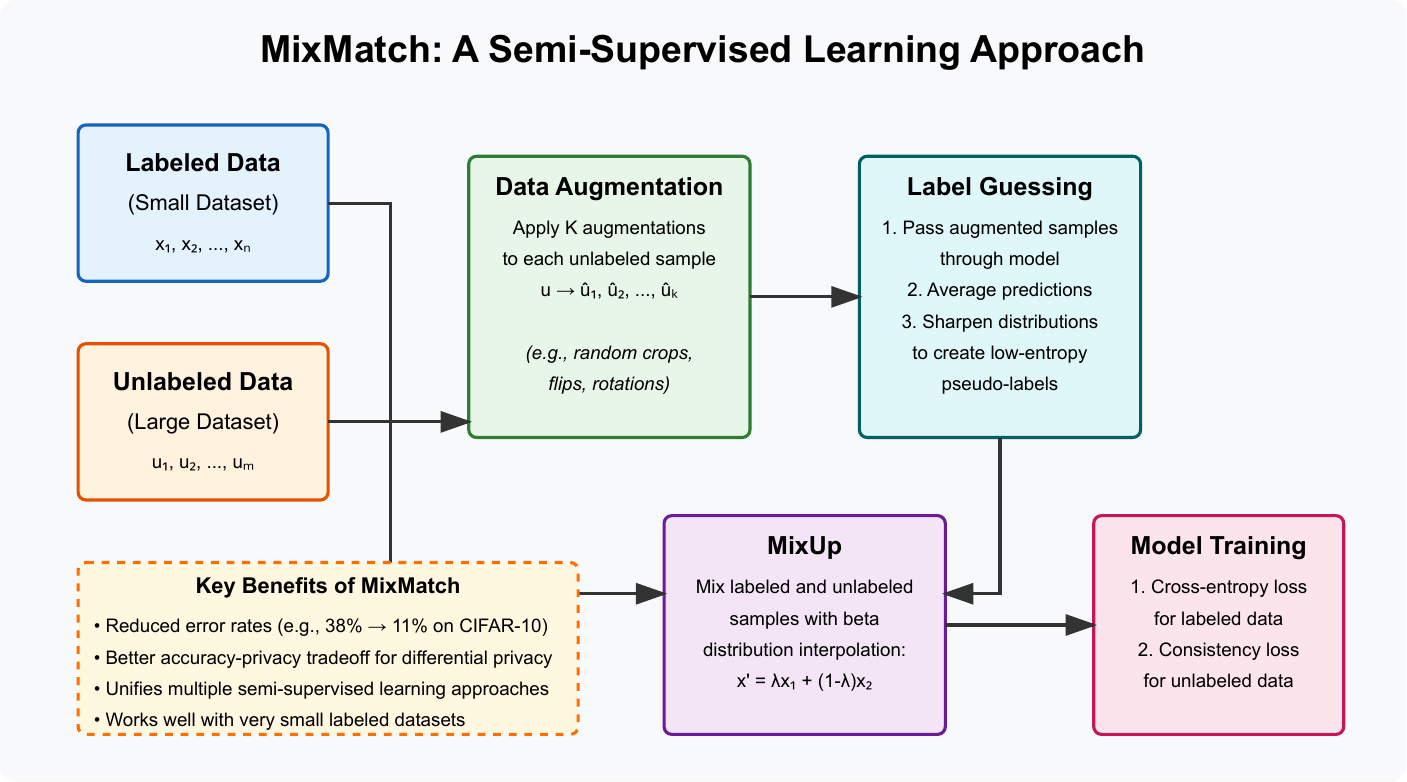}
\caption{\textit{MixMatch} a unified semi-supervised learning approach that combines data augmentation, low-entropy pseudo-labelling, and MixUp to achieve state-of-the-art results with limited labelled data.}
\label{MixMatch}
\end{figure}

\newpage
1-3-r.\textit{ MixMatch} \cite{berthelot2019mixmatch} is a novel semi-supervised learning algorithm that unifies dominant approaches by generating low-entropy pseudo-labels for augmented unlabeled data and blending labelled and unlabeled examples using MixUp, achieving state-of-the-art results across various datasets, as shown in \cref{MixMatch}. For instance, it reduces error rates on benchmarks like CIFAR-10 (from 38\% to 11\% with 250 labels) and STL-10, demonstrating its effectiveness in leveraging unlabeled data. Additionally, MixMatch provides a dramatically improved accuracy-privacy trade-off for differential privacy, and an ablation study highlights the key components contributing to its success.

\begin{figure}[H]
\centering
\includegraphics[width=\textwidth]{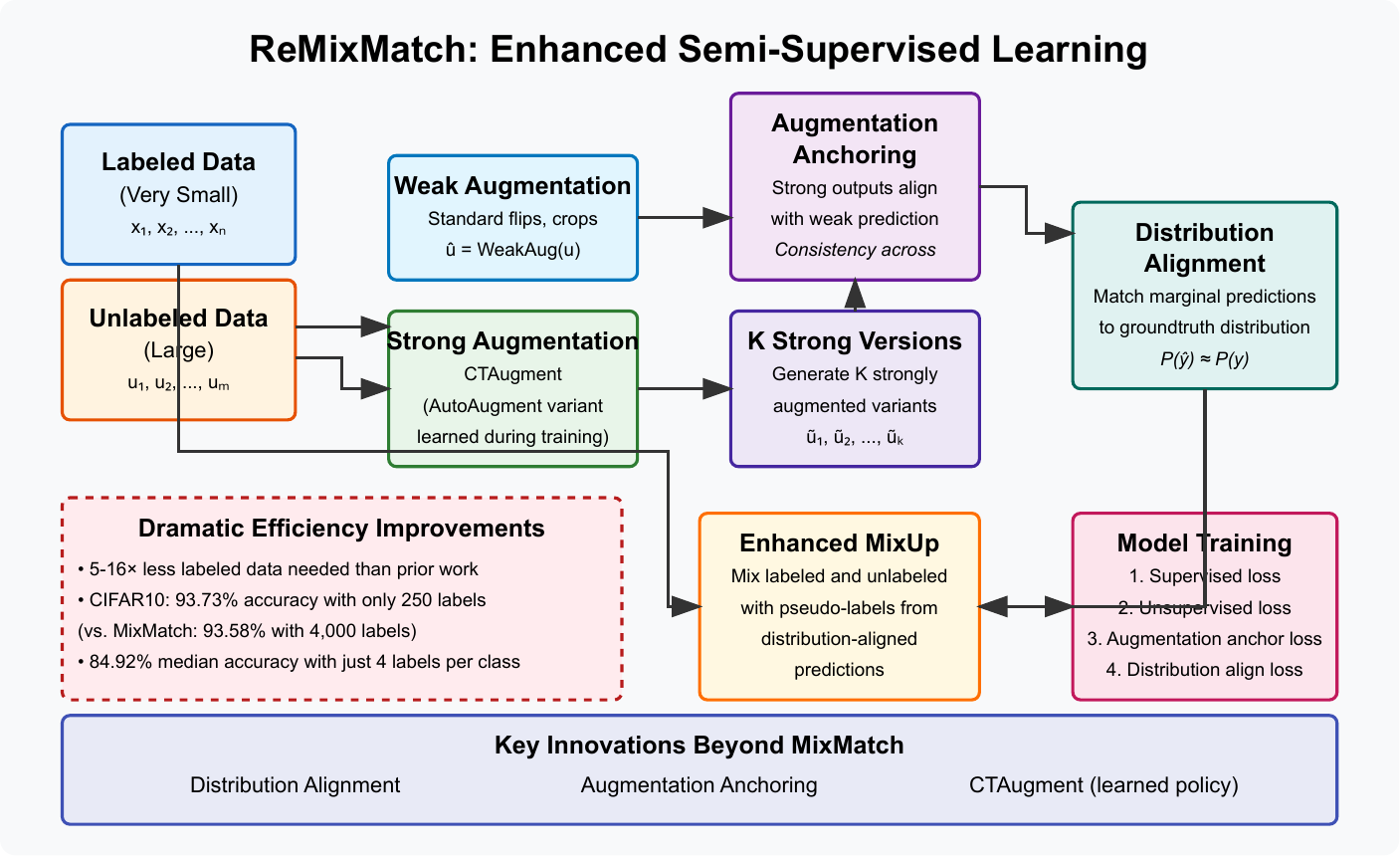}
\caption{\textit{ReMixMatch} an advanced semi-supervised learning approach that improves upon MixMatch by incorporating distribution alignment, augmentation anchoring, and learned strong augmentations to achieve superior performance with significantly less labelled data.}
\label{ReMixMatch}
\end{figure}

\newpage
1-3-s.\textit{ ReMixMatch} \cite{berthelot2019remixmatch} improves upon the MixMatch semi-supervised learning algorithm by introducing distribution alignment, which aligns the marginal distribution of predictions with ground truth labels, and augmentation anchoring, which enforces consistency across strongly and weakly augmented versions of the same input, as shown in \cref{ReMixMatch}. These innovations, combined with a learned augmentation policy, make ReMixMatch significantly more data-efficient, achieving 93.73\% accuracy on CIFAR-10 with only 250 labelled examples, compared to MixMatch's performance requiring 4,000 examples.

\begin{figure}[H]
\centering
\includegraphics[width=\textwidth]{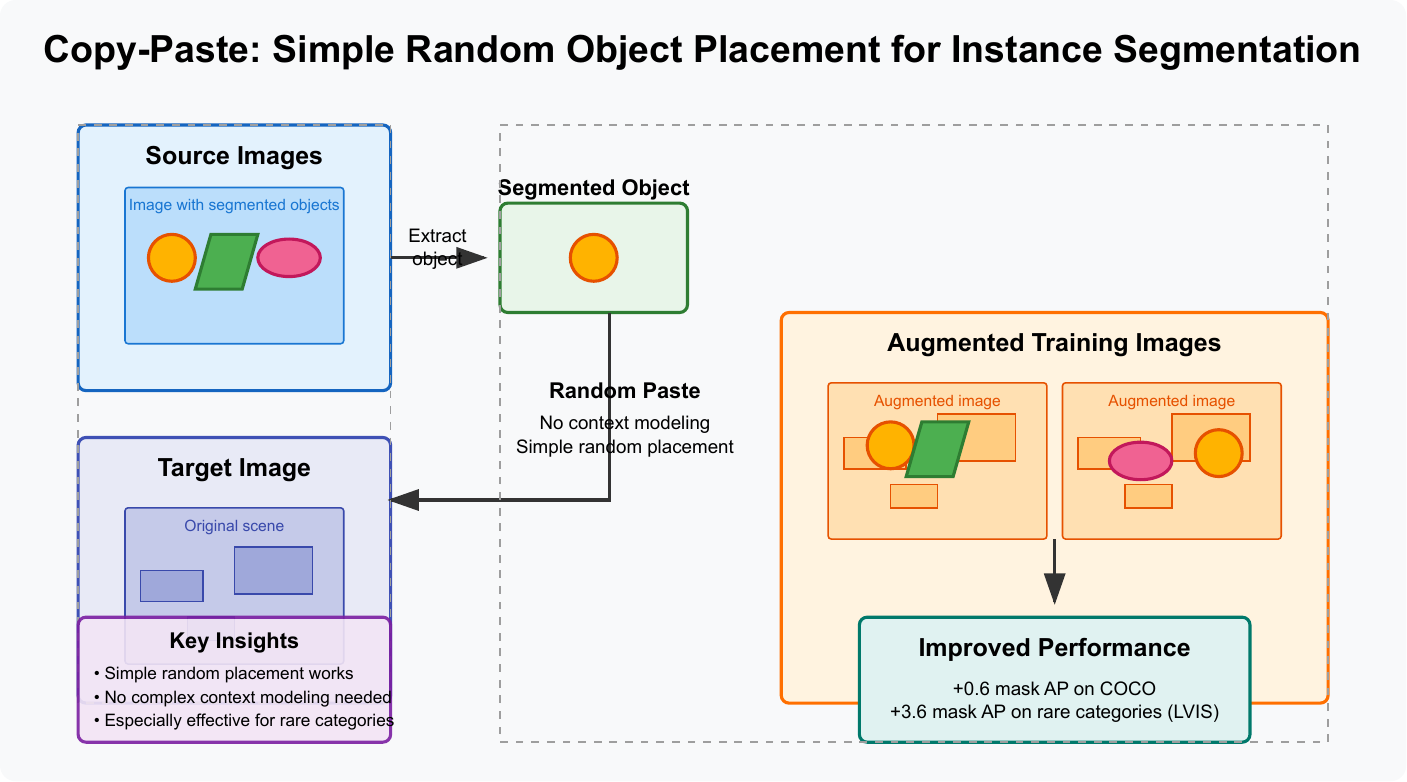}
\caption{\textit{Copy-Paste} a simple yet effective data augmentation technique that randomly places segmented objects onto images, significantly improving instance segmentation performance, especially for rare categories.}
\label{Copy-Paste}
\end{figure}

\newpage
1-3-t.\textit{ Copy-Paste} \cite{ghiasi2021simple} is a simple yet effective data augmentation technique for instance segmentation, where objects are randomly pasted onto images, providing substantial performance gains without relying on complex visual context modelling, as shown in \cref{Copy-Paste}. This approach not only improves the latest results in COCO (e.g. +0.6 mask AP and +1.5 box AP) but also demonstrates significant benefits in the LVIS benchmark, outperforming the LVIS 2020 Challenge winner by +3.6 mask AP on rare categories.

\begin{figure}[H]
\centering
\includegraphics[width=\textwidth]{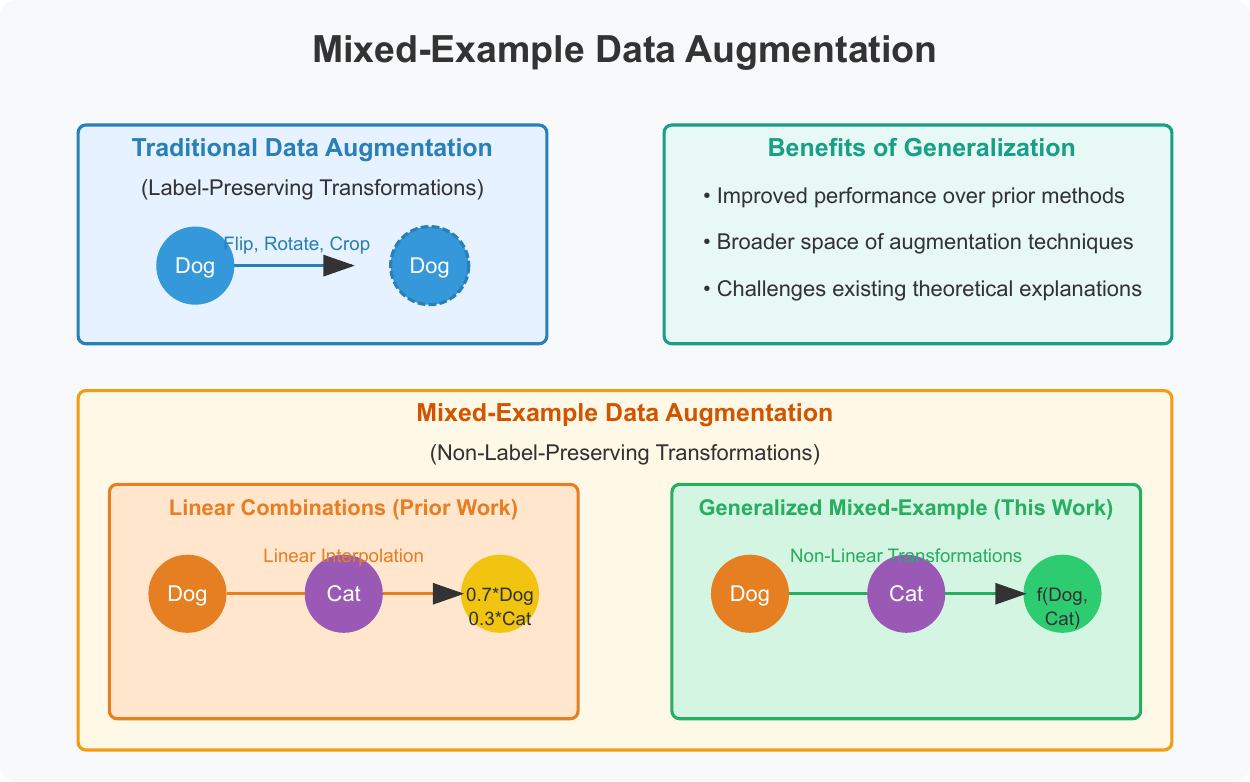}
\caption{\textit{Mixed-Example} data augmentation extends beyond traditional label-preserving transformations and linear combinations to explore a broader space of non-linear example mixing techniques for improved neural network training.}
\label{Mixed-Example}
\end{figure}

\newpage
1-3-u.\textit{ Mixed-Example} \cite{summers2019improved} data augmentation generalizes previous methods that combine pairs of examples through linear operations, exploring a broader space of augmentation techniques that challenge the necessity of linearity and improve upon prior state-of-the-art, as shown in \cref{Mixed-Example}. This generalized approach not only enhances performance but also uncovers novel augmentation strategies, highlighting the limitations of existing theories and suggesting the need for a more comprehensive understanding of why such methods are effective.

\begin{figure}[H]
\centering
\includegraphics[width=\textwidth]{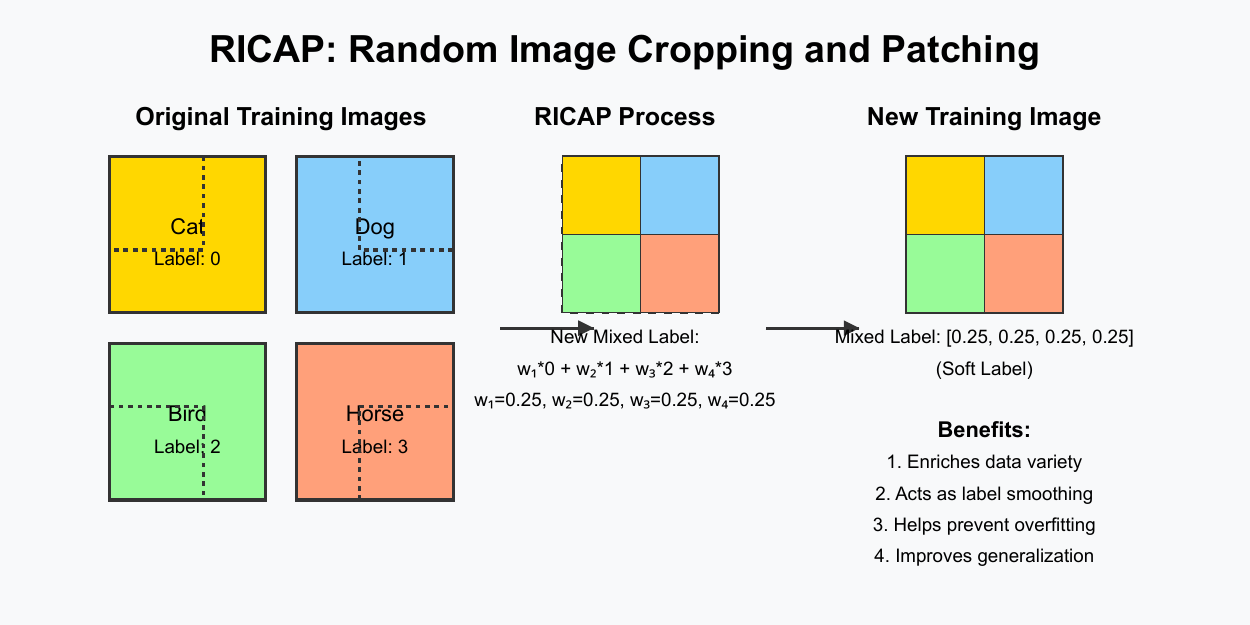}
\caption{\textit{RICAP} combines random crops from four different images into a single composite training sample while mixing their class labels to improve CNN generalization performance.}
\label{RICAP}
\end{figure}

\newpage
1-3-v.\textit{ RICAP} \cite{takahashi2018ricap} (Random Image Cropping and Patching) is a novel data augmentation technique that enhances training diversity by randomly cropping four images, patching them together, and mixing their class labels, thereby mitigating overfitting and improving model generalization, as shown in \cref{RICAP}. Evaluated with state-of-the-art CNNs, RICAP outperforms competitive methods like cutout and mixup, achieving a test error of 2.23\% on CIFAR-10 and demonstrating superior results on CIFAR-100 and ImageNet.

\begin{figure}[H]
\centering
\includegraphics[width=\textwidth]{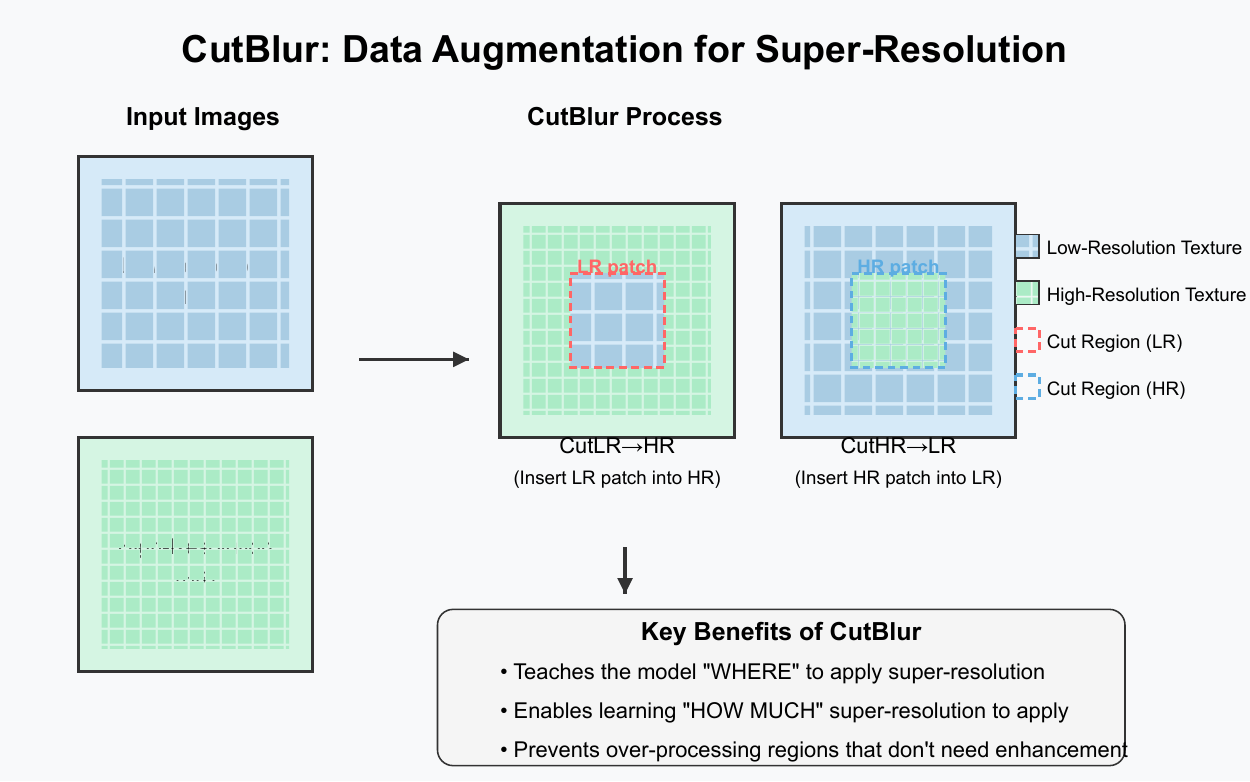}
\caption{\textit{CutBlur} augments super-resolution training by strategically exchanging patches between low and high-resolution image pairs to teach models where and how much enhancement to apply.}
\label{CutBlur}
\end{figure}

\newpage
1-3-w.\textit{ CutBlur} \cite{yoo2020rethinking} is a novel data augmentation technique designed for low-level vision tasks, such as super-resolution, which enhances model learning by cutting and pasting low-resolution patches into high-resolution images and vice versa, as shown in \cref{CutBlur}. This approach enables models to learn not only "how" but also "where" and "how much" to apply super-resolution, leading to consistent performance improvements across various tasks, including denoising and compression artifact removal.

\begin{figure}[H]
\centering
\includegraphics[width=\textwidth]{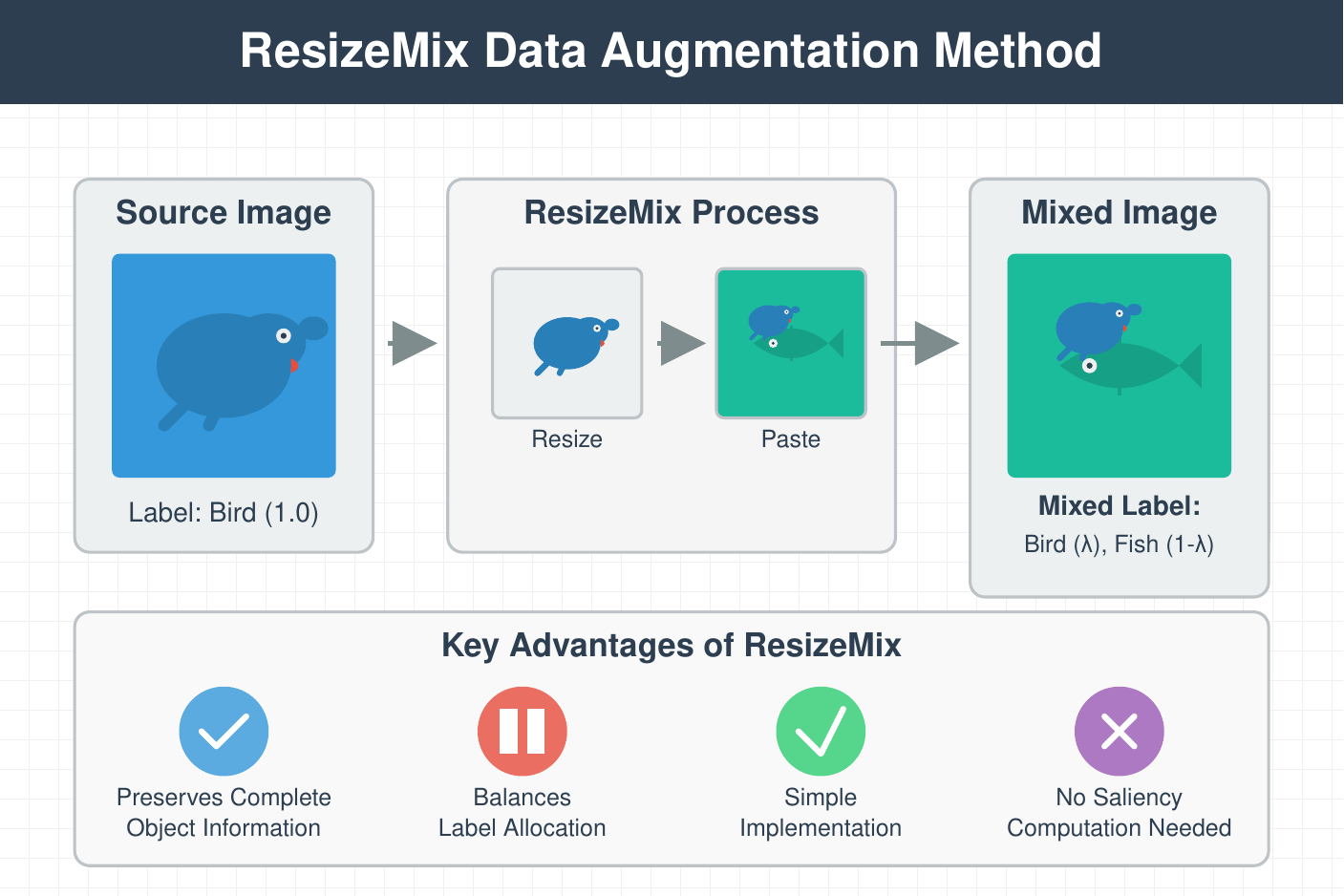}
\caption{\textit{ResizeMix} augments data by resizing a source image and pasting it onto a target image while preserving complete object information and proportionally mixing their labels.}
\label{ResizeMix}
\end{figure}

\newpage
1-3-x.\textit{ ResizeMix} \cite{qin2020resizemix} is a simple yet effective data augmentation technique that improves upon cut-based methods like CutMix by resizing a source image into a smaller patch and pasting it onto another image, preserving more substantial object information, as shown in \cref{ResizeMix}. This approach addresses issues like label misallocation and missing object information, demonstrating superior performance in image classification and object detection tasks while outperforming more complex augmentation methods without additional computational cost.

\begin{figure}[H]
\centering
\includegraphics[width=\textwidth]{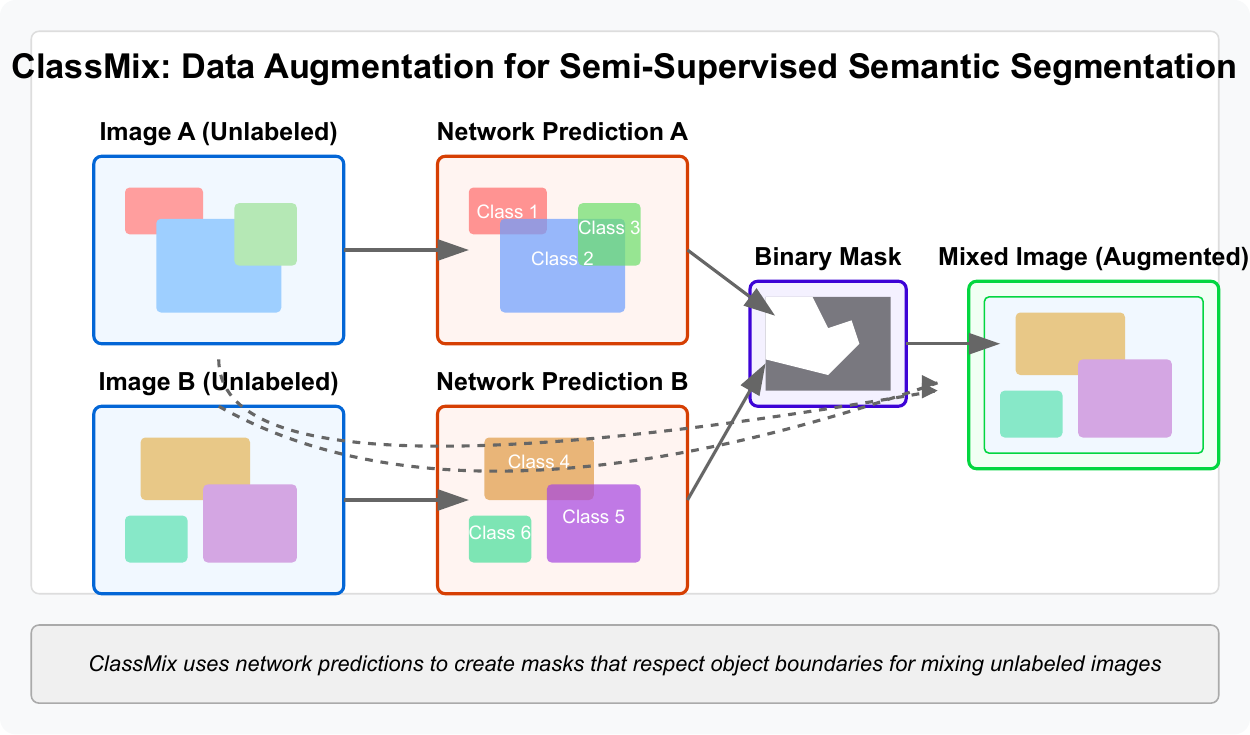}
\caption{\textit{Classmix} creates data-augmented images by using network predictions to generate masks that preserve object boundaries when mixing unlabeled images for semi-supervised semantic segmentation.}
\label{Classmix}
\end{figure}

\newpage
1-3-y.\textit{ Classmix} \cite{olsson2021classmix} is a novel data augmentation mechanism for semi-supervised semantic segmentation that generates augmentations by mixing unlabeled samples while leveraging network predictions to preserve object boundaries, as shown in \cref{Classmix}. Evaluations on standard benchmarks demonstrate its state-of-the-art performance, supported by extensive ablation studies analyzing various design choices and training regimes.

\begin{figure}[H]
\centering
\includegraphics[width=\textwidth]{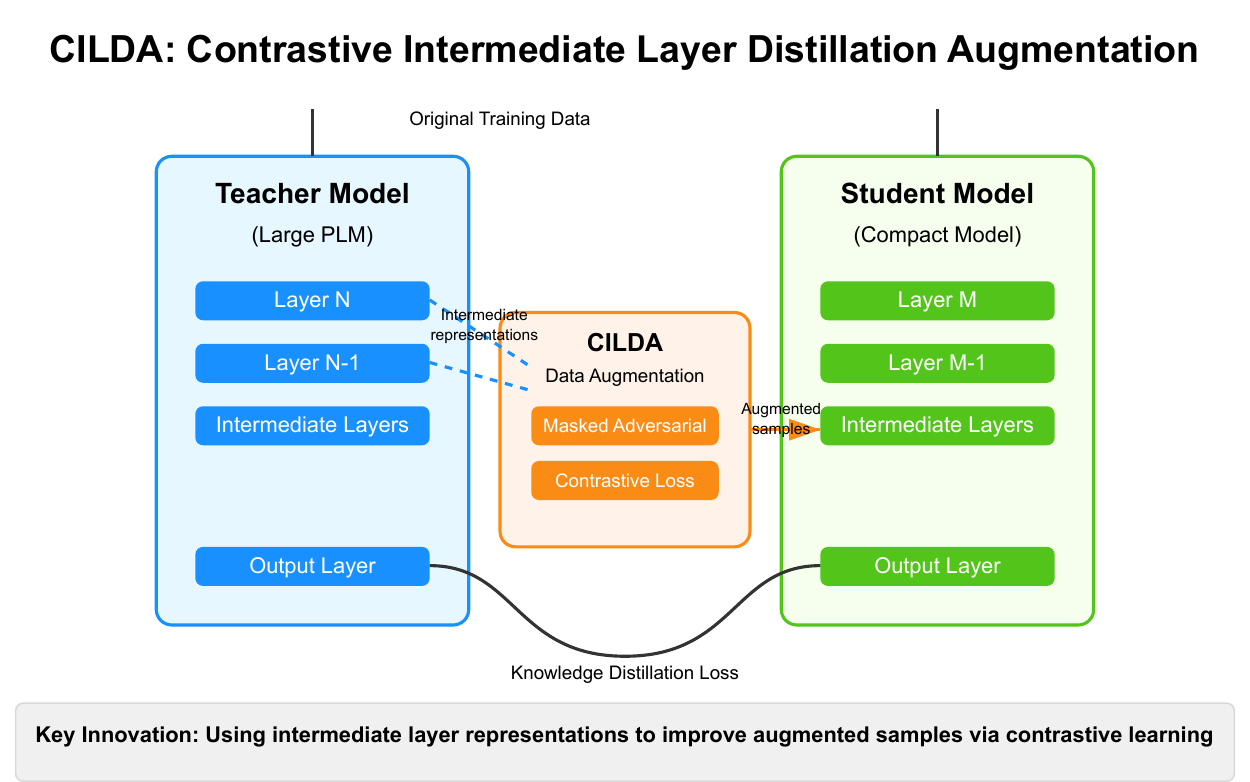}
\caption{\textit{CDA} enhances knowledge distillation by using intermediate layer representations and contrastive learning to generate high-quality augmented training samples.}
\label{CDA}
\end{figure}

\newpage
1-3-z.\textit{ CDA} \cite{haidar2022cilda} (Contrastive Data Augmentation) is a specialized data augmentation strategy designed to enhance contrastive learning by generating semantically consistent positive pairs and diverse negative pairs, as shown in \cref{CDA}. By leveraging techniques such as random cropping, color jittering, and strong augmentations, CDA aims to improve the quality of feature representations and boost performance in tasks like knowledge distillation and self-supervised learning.

\begin{figure}[H]
\centering
\includegraphics[width=\textwidth]{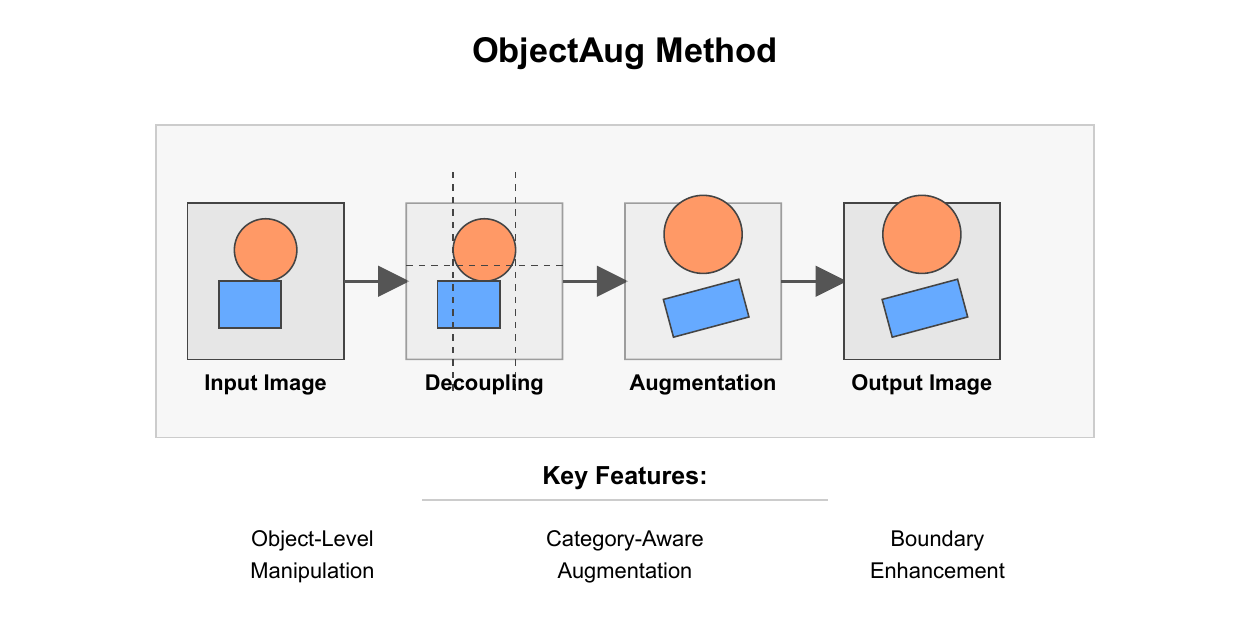}
\caption{\textit{ObjectAug} decouples objects from the background, individually augments them, and reassembles the image to improve semantic segmentation performance.}
\label{ObjectAug}
\end{figure}

\newpage
1-3-aa.\textit{ ObjectAug} \cite{zhang2021objectaug} an object-level data augmentation method for semantic segmentation that decouples images into objects and backgrounds, applies augmentations to each object, and restores artifacts via inpainting, as shown in \cref{ObjectAug}. This approach enhances boundary diversity and supports category-aware augmentation, improving segmentation performance.

\begin{figure}[H]
\centering
\includegraphics[width=\textwidth]{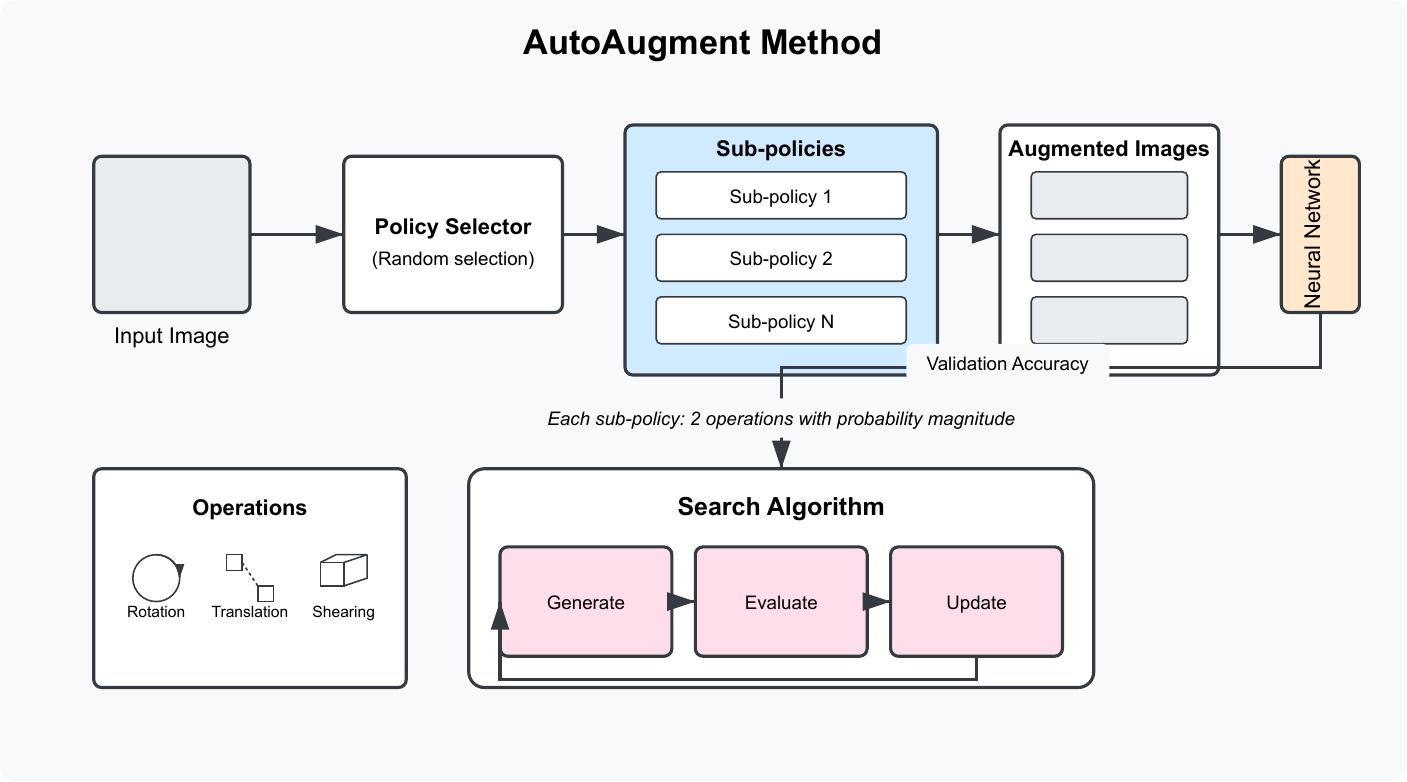}
\caption{\textit{AutoAugment} automatically searches for optimal data augmentation policies by selecting sub-policies with paired operations, each having learned probability and magnitude parameters.}
\label{AutoAugment}
\end{figure}

\newpage
2-1-a.\textit{ AutoAugment} \cite{cubuk2019autoaugment} that automatically search for improved data augmentation policies. It consists of two parts: search algorithm and search space. The search algorithm is designed to find the best policy with the highest validation accuracy, as shown in \cref{AutoAugment}. The search space contains many policies that detail various augmentation operations and magnitudes with which the operations are applied.

\begin{figure}[H]
\centering
\includegraphics[width=\textwidth]{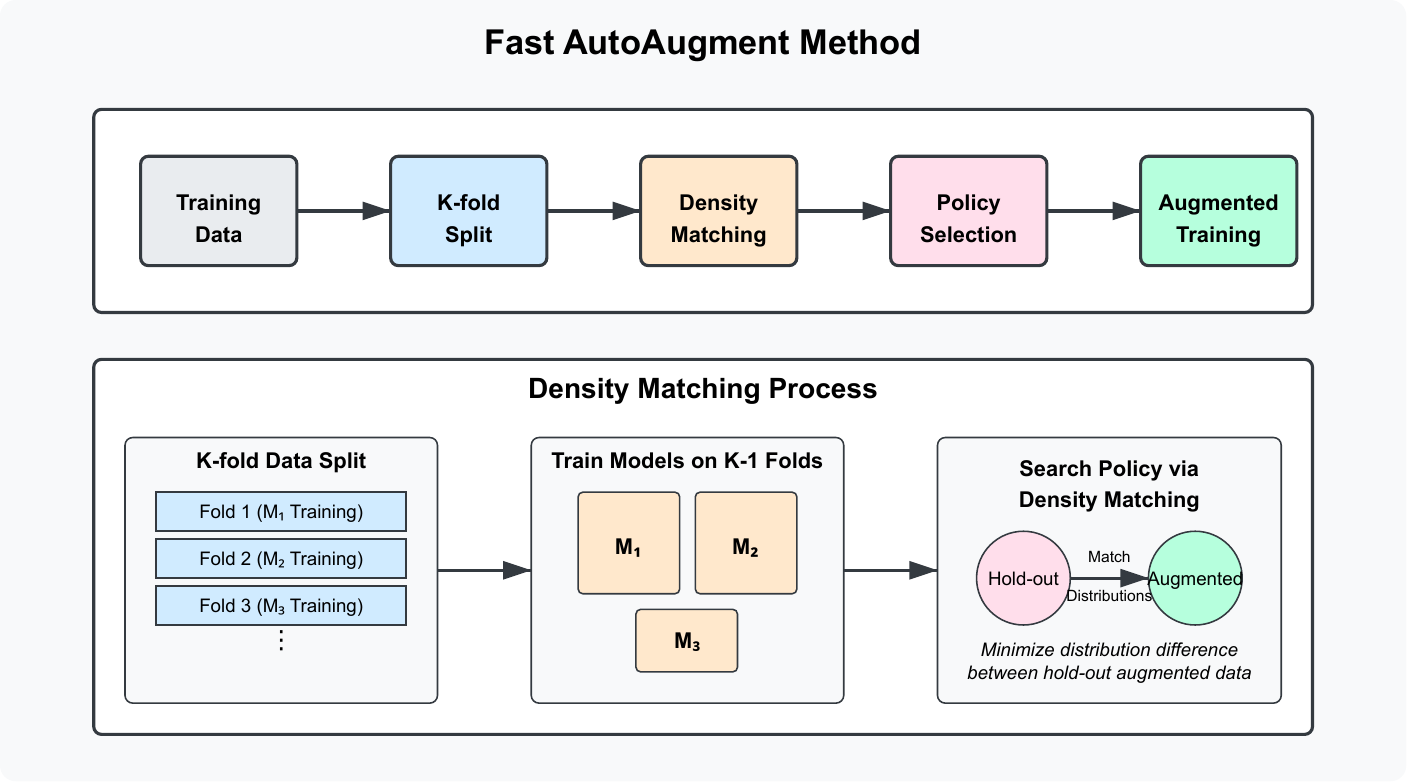}
\caption{\textit{Fast AutoAugment} accelerates data augmentation policy search through efficient density matching between hold-out and augmented data distributions, reducing computational requirements while maintaining performance.}
\label{Fast AutoAugment}
\end{figure}

\newpage
2-1-b.\textit{ Fast AutoAugment} \cite{lim2019fast} is a novel data augmentation algorithm that significantly reduces the search time for effective augmentation policies by employing a more efficient density-matching strategy, as shown in \cref{Fast AutoAugment}. Compared to AutoAugment, it achieves comparable performance on diverse image recognition tasks while reducing computational requirements by orders of magnitude. 

\begin{figure}[H]
\centering
\includegraphics[width=\textwidth]{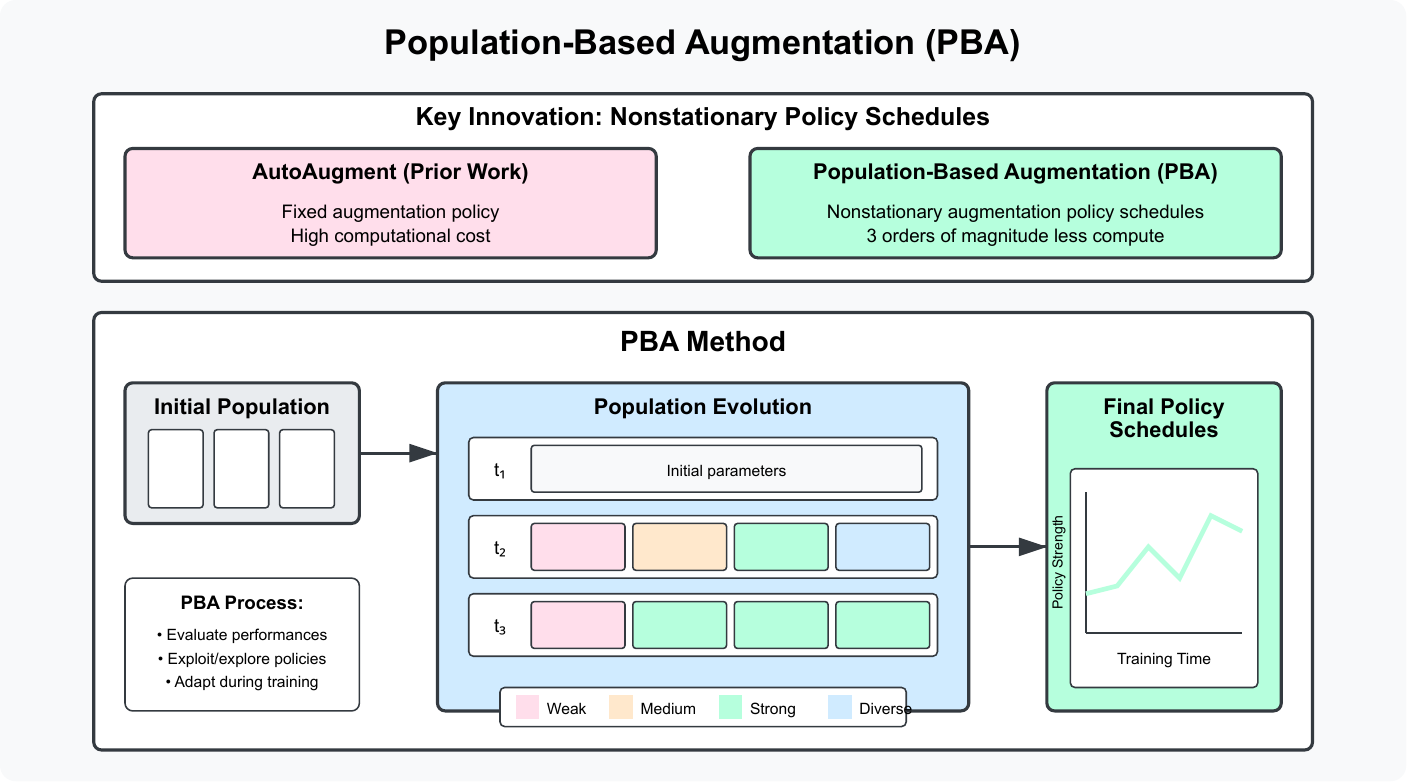}
\caption{\textit{Population-Based Augmentation(PBA)} evolves nonstationary policy schedules throughout training using evolutionary algorithms, achieving state-of-the-art performance with orders of magnitude less computation than previous methods.}
\label{PBA}
\end{figure}

\newpage
2-1-c.\textit{ Population-Based Augmentation(PBA)} \cite{ho2019population} that reduce the time cost of AutoAugment which generates nonstationary augmentation policy schedules instead of a fixed augmentation policy, as shown in \cref{PBA}. PBA can match the performance of AutoAugment on multiple datasets with less computation time.

\begin{figure}[H]
\centering
\includegraphics[width=\textwidth]{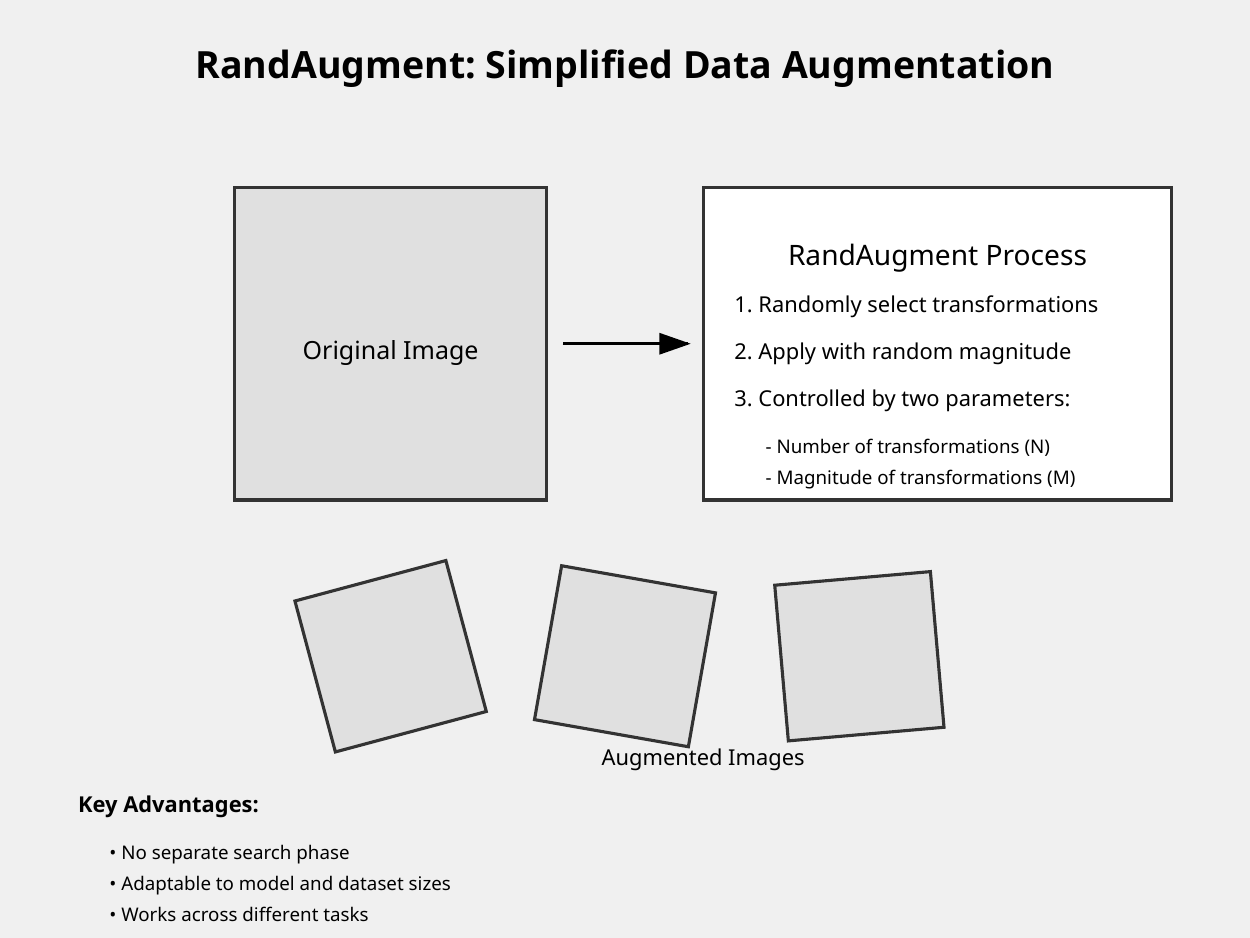}
\caption{\textit{RandAugment} simplifies data augmentation by randomly applying transformations with two tunable parameters, N (number of transformations) and M (magnitude), across different tasks and model sizes.}
\label{RandAugment}
\end{figure}

\newpage
2-1-d.\textit{ RandAugment} \cite{cubuk2020randaugment} is a streamlined data augmentation method that eliminates the need for a separate search phase by significantly reducing the search space, enabling direct application to target tasks, as shown in \cref{RandAugment}. Its parameterized design allows for adjustable regularization strength across different model and dataset sizes, achieving state-of-the-art performance on benchmarks like CIFAR-10/100, SVHN, ImageNet, and COCO while simplifying implementation and reducing the computational cost.

\begin{figure}[H]
\centering
\includegraphics[width=\textwidth]{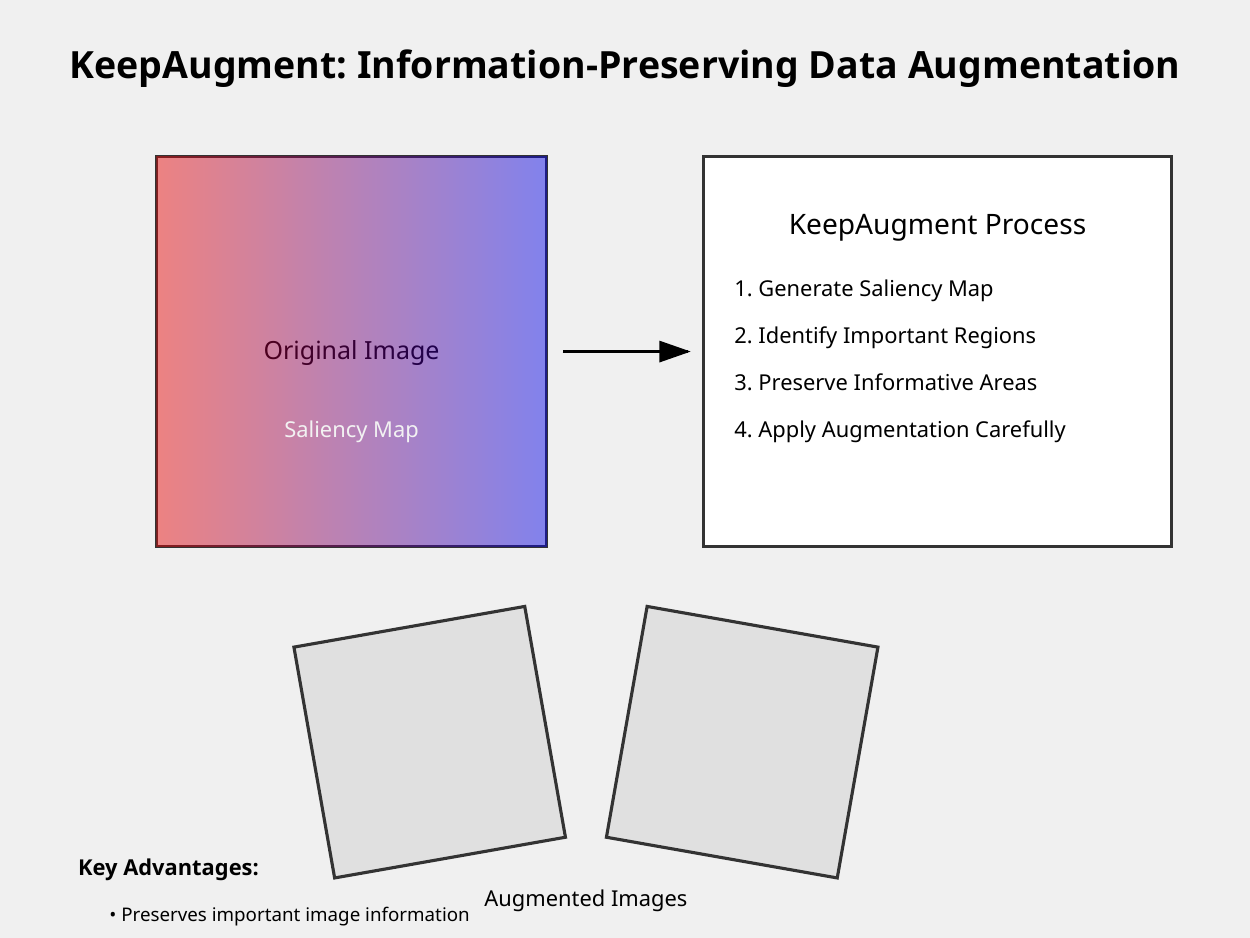}
\caption{\textit{KeepAugment} preserves important image regions during data augmentation by using saliency maps to generate more faithful and informative training examples.}
\label{KeepAugment}
\end{figure}

\newpage
2-1-e.\textit{ KeepAugment} \cite{gong2021keepaugment} is a novel data augmentation method designed to preserve the fidelity of augmented images by leveraging saliency maps to detect and retain important regions during augmentation, as shown in \cref{KeepAugment}. By generating more faithful training examples, KeepAugment improves upon existing augmentation techniques such as AutoAugment and Cutout, achieving superior performance across tasks like image classification, semi-supervised learning, object detection, and multi-camera tracking.

\begin{figure}[H]
\centering
\includegraphics[width=\textwidth]{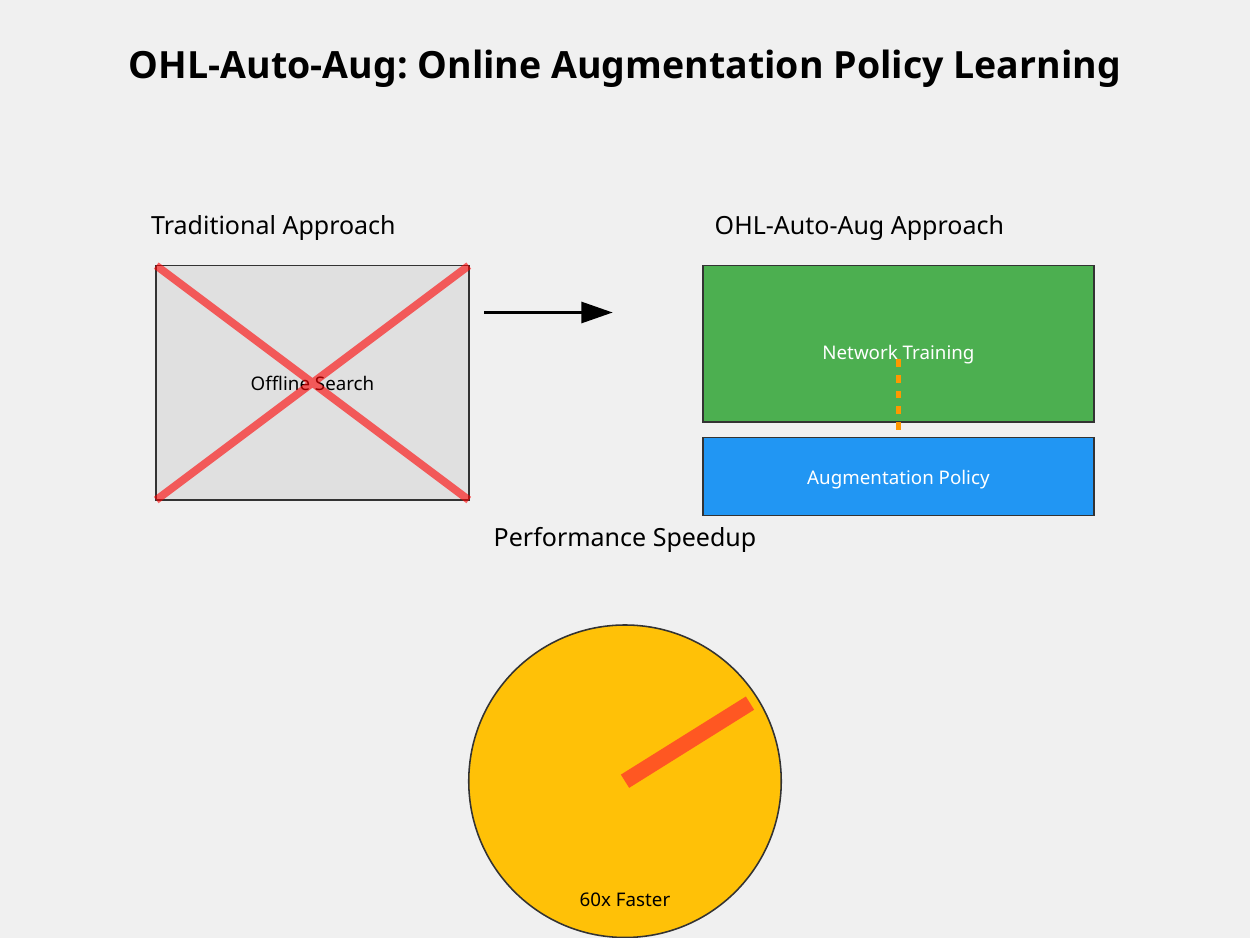}
\caption{\textit{OHL-Auto-Aug} enables online, joint optimization of augmentation policies with network training, dramatically reducing computational search costs while maintaining model performance.}
\label{OHL-Auto-Aug}
\end{figure}

\newpage
2-1-f.\textit{ OHL-Auto-Aug} \cite{lin2019online} (Online Hyper-parameter Learning for Auto-Augmentation) introduces an efficient approach to learning augmentation policy distributions jointly with network training, eliminating the need for a separate offline search phase, as shown in \cref{OHL-Auto-Aug}. This method significantly reduces computational costs, achieving up to 60x and 24x faster search on CIFAR-10 and ImageNet, respectively, while delivering competitive accuracy improvements over baseline models.

\begin{figure}[H]
\centering
\includegraphics[width=\textwidth]{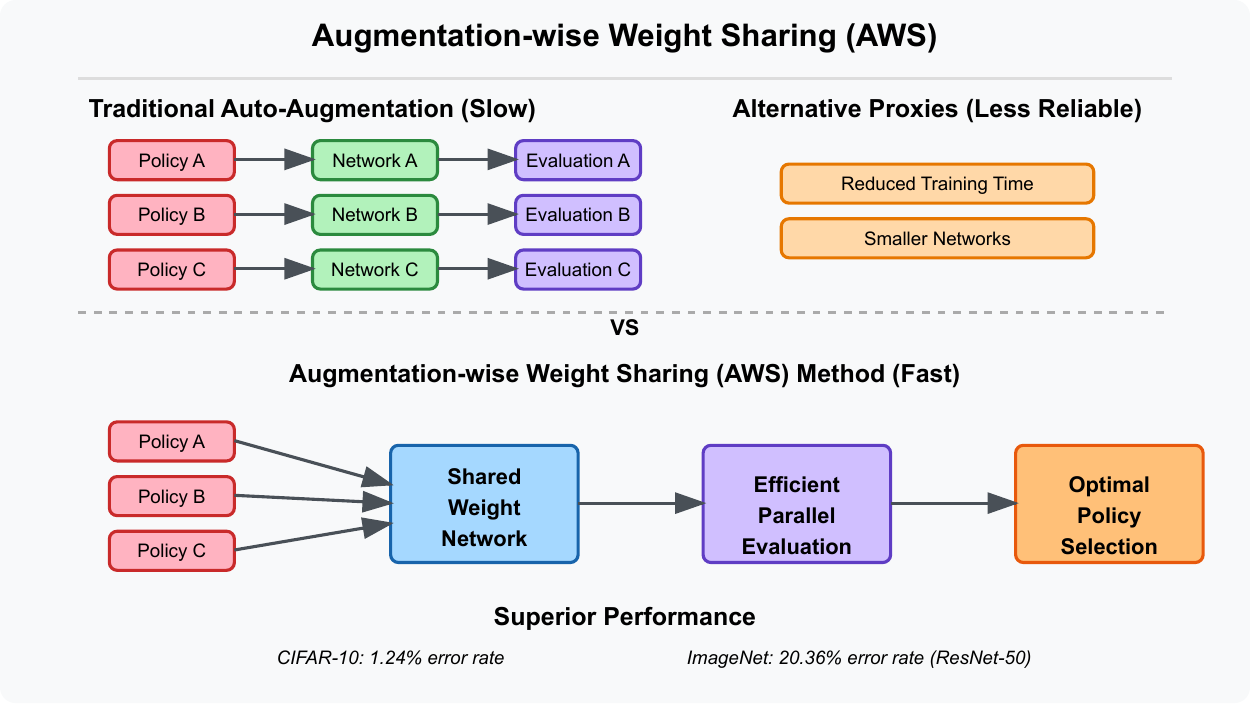}
\caption{\textit{Augmentation-wise Weight Sharing}  method enables efficient augmentation policy search by using a shared neural network to evaluate multiple data augmentation policies simultaneously, dramatically reducing search time while maintaining evaluation reliability.}
\label{Augmentation-wise Weight Sharing}
\end{figure}

\newpage
2-1-g.\textit{ Augmentation-wise Weight Sharing} \cite{tian2020improving} (AWS) introduces an efficient proxy task for evaluating augmentation policies by sharing model weights across different augmentations, significantly improving the speed and reliability of the search process, as shown in \cref{Augmentation-wise Weight Sharing}. This method achieves state-of-the-art results on benchmarks like CIFAR-10 and ImageNet, demonstrating both superior accuracy and efficiency compared to existing auto-augmentation approaches.

\begin{figure}[H]
\centering
\includegraphics[width=\textwidth]{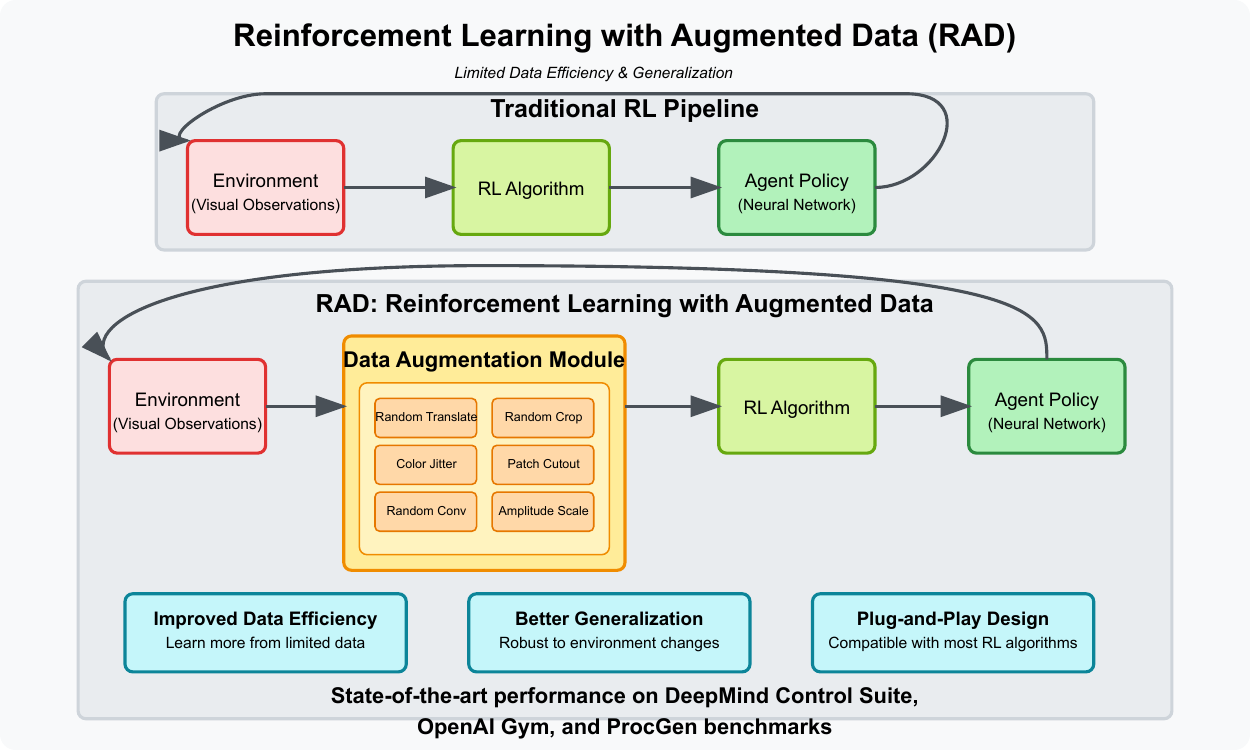}
\caption{\textit{RAD(Reinforcement Learning with Augmented Data)} enhances RL algorithms by applying diverse data augmentations to visual observations, improving data efficiency and generalization without modifying the core algorithm.}
\label{RAD}
\end{figure}

\newpage
2-1-h.\textit{ RAD(Reinforcement Learning with Augmented Data)} \cite{laskin2020reinforcement} is a versatile module that enhances RL algorithms by incorporating general data augmentations, improving both data efficiency and generalization to new environments, as shown in \cref{RAD}. RAD achieves state-of-the-art performance on benchmarks like the DeepMind Control Suite and OpenAI Gym, while also demonstrating superior test-time generalization on OpenAI ProcGen tasks.

\begin{figure}[H]
\centering
\includegraphics[width=\textwidth]{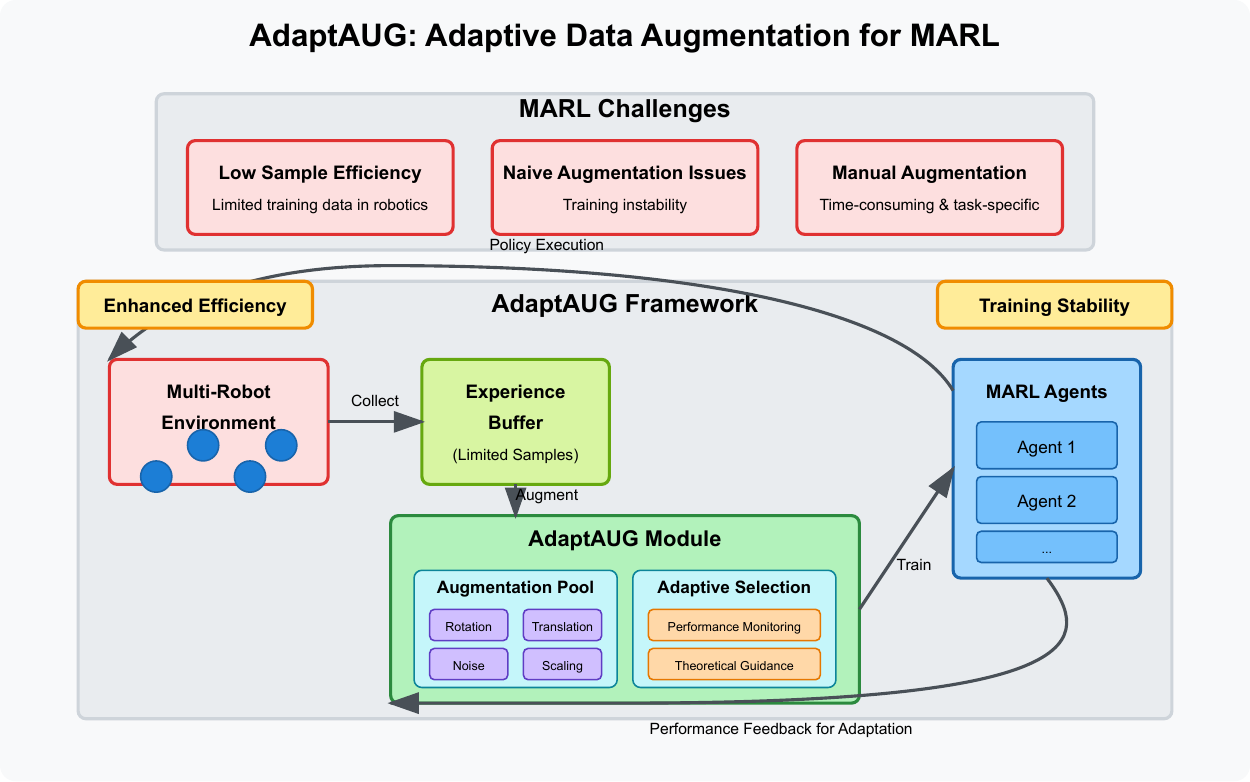}
\caption{\textit{MARL(Multi-Agent Reinforcement Learning)} selectively identifies beneficial data augmentations for multi-agent reinforcement learning, enhancing sample efficiency and training stability in multi-robot systems through adaptive augmentation selection guided by theoretical insights.}
\label{MARL}
\end{figure}

\newpage
2-1-i.\textit{ MARL(Multi-Agent Reinforcement Learning)} \cite{yu2024adaptaug} is a powerful framework for controlling multi-robot systems but suffers from low sample efficiency, limiting its practical applications, as shown in \cref{MARL}. To address this, adaptive methods like AdaptAUG have been proposed to optimize data augmentation strategies, significantly improving sample efficiency and performance in both simulated and real-world multi-robot tasks.

\begin{figure}[H]
\centering
\includegraphics[width=\textwidth]{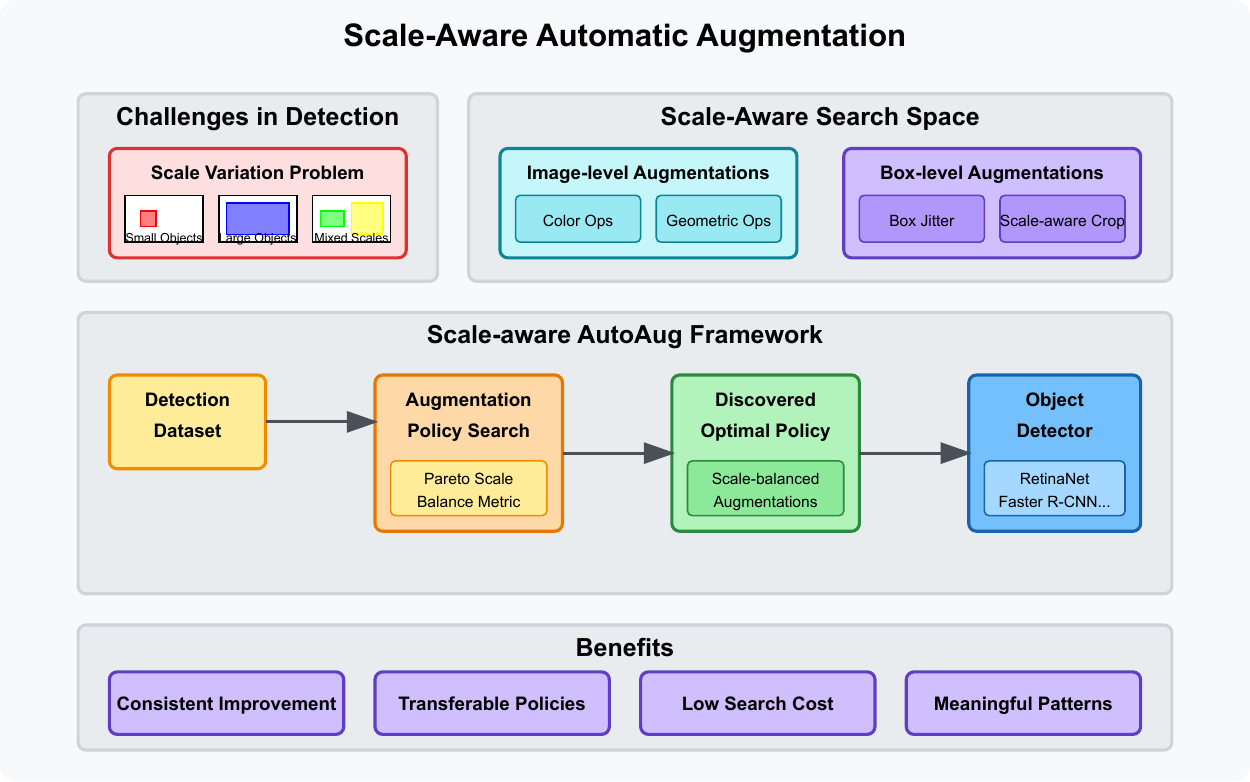}
\caption{\textit{Scale-Aware Automatic Augmentation} addresses object detection's scale variation challenge by automatically discovering optimal augmentation policies through a specialized search space and Pareto Scale Balance metric that maintains scale invariance across both image and box-level transformations.}
\label{Scale-Aware Automatic Augmentation}
\end{figure}

\newpage
2-1-j.\textit{ Scale-Aware Automatic Augmentation} \cite{chen2021scale} introduces a novel scale-aware search space and a Pareto Scale Balance metric to efficiently learn augmentation policies tailored for object detection, ensuring scale invariance at both image and box levels, as shown in \cref{Scale-Aware Automatic Augmentation}. This approach delivers significant performance improvements across various object detectors and tasks like instance segmentation and keypoint estimation while maintaining lower search costs compared to previous methods.

\begin{figure}[H]
\centering
\includegraphics[width=\textwidth]{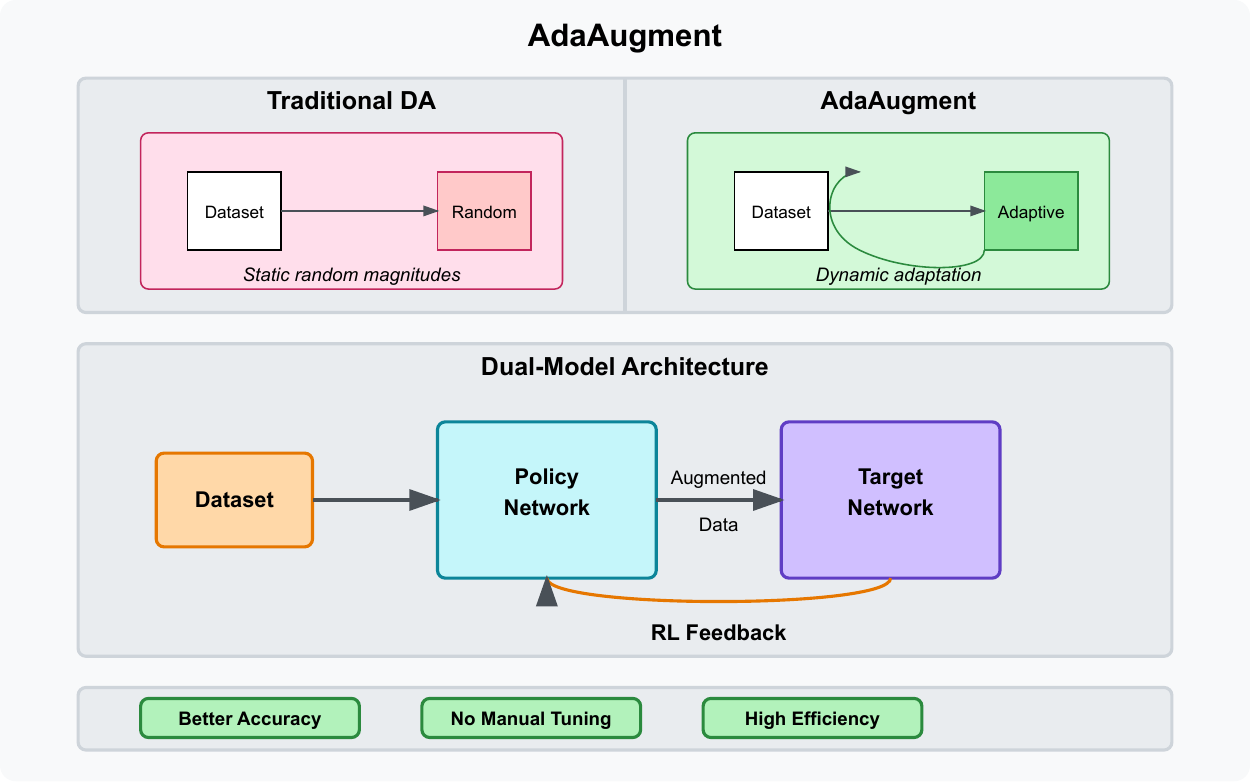}
\caption{\textit{ADA(Adaptive Data Augmentation)} employs a dual-model reinforcement learning architecture that dynamically adjusts data augmentation magnitudes based on real-time feedback from the target network, eliminating manual tuning and improving model generalization.}
\label{ADA(Adaptive Data Augmentation)}
\end{figure}

\newpage
2-1-k.\textit{ ADA(Adaptive Data Augmentation)} \cite{yang2024adaaugment} introduces a reinforcement learning-based framework to dynamically adjust augmentation magnitudes for individual training samples, addressing the misalignment between static augmentations and the evolving training status of models, as shown in \cref{ADA(Adaptive Data Augmentation)}. By leveraging a dual-model architecture, AdaAugment achieves superior generalization performance across various benchmarks, consistently outperforming existing data augmentation methods in both effectiveness and efficiency.

\begin{figure}[H]
\centering
\includegraphics[width=\textwidth]{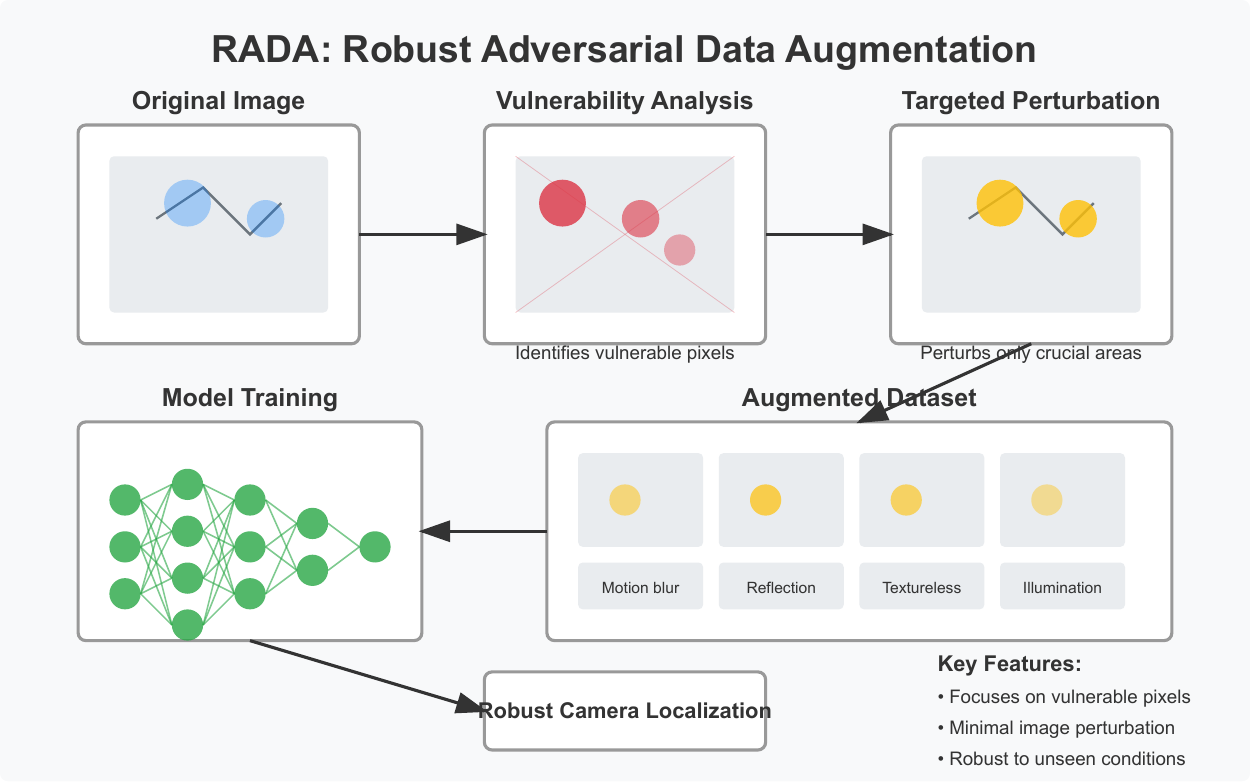}
\caption{\textit{RADA(Robust Adversarial Data Augmentation)} is a robust adversarial data augmentation method that improves camera localization by identifying and perturbing vulnerable pixels rather than applying general image transformations.}
\label{RADA(Robust Adversarial Data Augmentation)}
\end{figure}

\newpage
2-1-l.\textit{ RADA(Robust Adversarial Data Augmentation)} \cite{wang2023rada} introduces a targeted approach to data augmentation by perturbing the most vulnerable pixels, enhancing robustness in camera localization under challenging conditions, as shown in \cref{RADA(Robust Adversarial Data Augmentation)}. This method significantly outperforms existing techniques, achieving up to double the accuracy of state-of-the-art models, even in unseen adverse weather scenarios.

\begin{figure}[H]
\centering
\includegraphics[width=\textwidth]{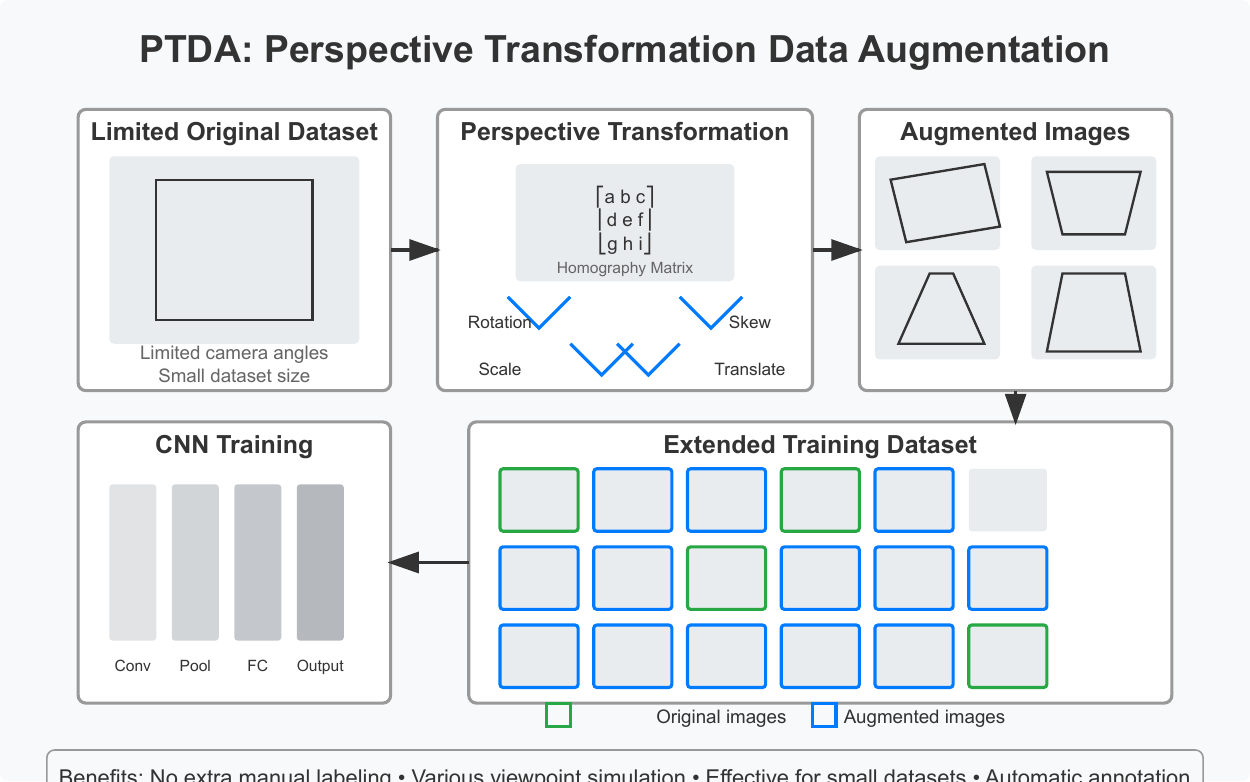}
\caption{\textit{PTDA(Perspective Transformation Data Augmentation)} is a data augmentation framework that generates new training images by applying perspective transformations to simulate different camera viewpoints without requiring additional manual annotations.}
\label{PTDA(Perspective Transformation Data Augmentation)}
\end{figure}

\newpage
2-1-m.\textit{ PTDA(Perspective Transformation Data Augmentation)} \cite{wang2019perspective} introduces a novel framework that generates annotated data by simulating images from various camera viewpoints, effectively expanding limited datasets without additional manual labelling, as shown in \cref{PTDA(Perspective Transformation Data Augmentation)}. This approach enhances the performance of deep CNNs, particularly on small or imbalanced datasets, as demonstrated by extensive experiments across multiple benchmarks.

\begin{figure}[H]
\centering
\includegraphics[width=\textwidth]{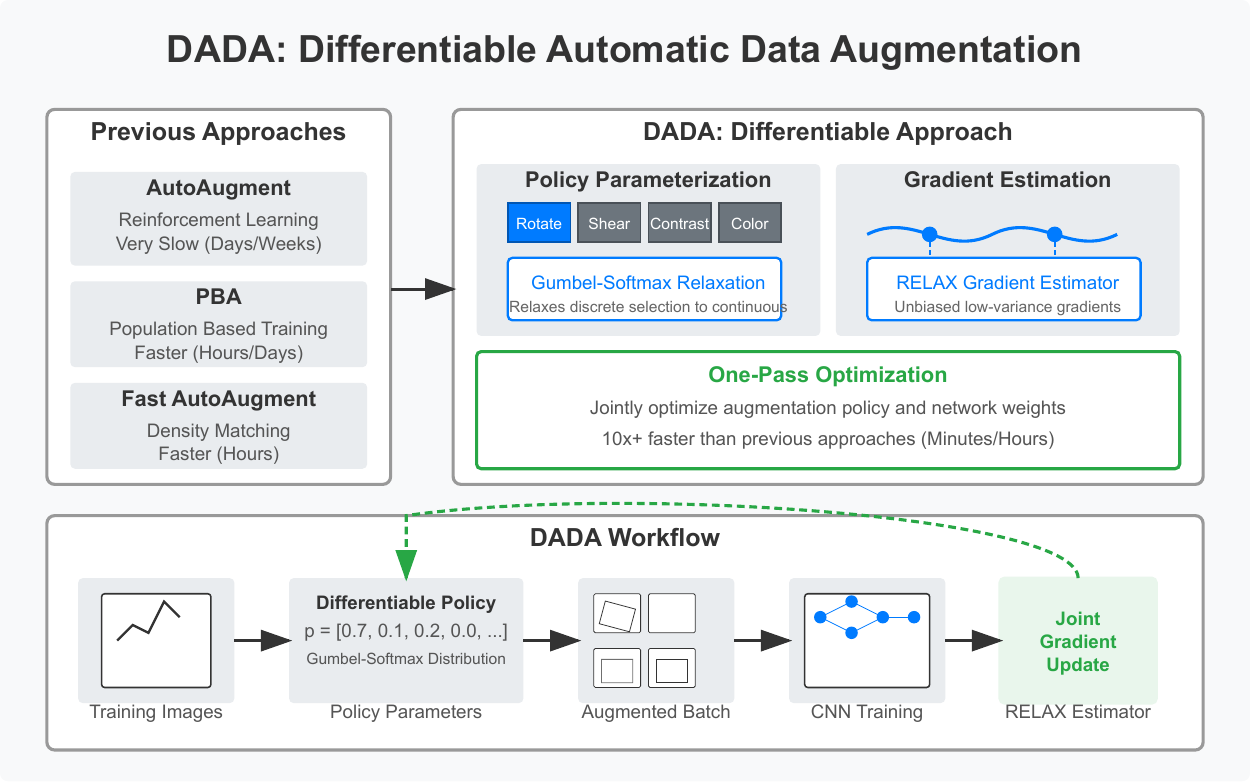}
\caption{\textit{DADA(Differentiable Automatic Data Augmentation)} is a differentiable data augmentation framework that dramatically accelerates automatic policy search by using Gumbel-Softmax relaxation and the RELAX gradient estimator for one-pass optimization.}
\label{DADA}
\end{figure}

\newpage
2-1-n.\textit{ DADA(Differentiable Automatic Data Augmentation)} \cite{li2020differentiable} introduces a novel approach to data augmentation policy optimization by formulating it as a differentiable problem using Gumbel-Softmax and an unbiased gradient estimator, RELAX, as shown in \cref{DADA}. This method significantly accelerates the search process, achieving state-of-the-art efficiency and comparable accuracy across multiple datasets, while also demonstrating its value for pre-training in downstream tasks.

\begin{figure}[H]
\centering
\includegraphics[width=\textwidth]{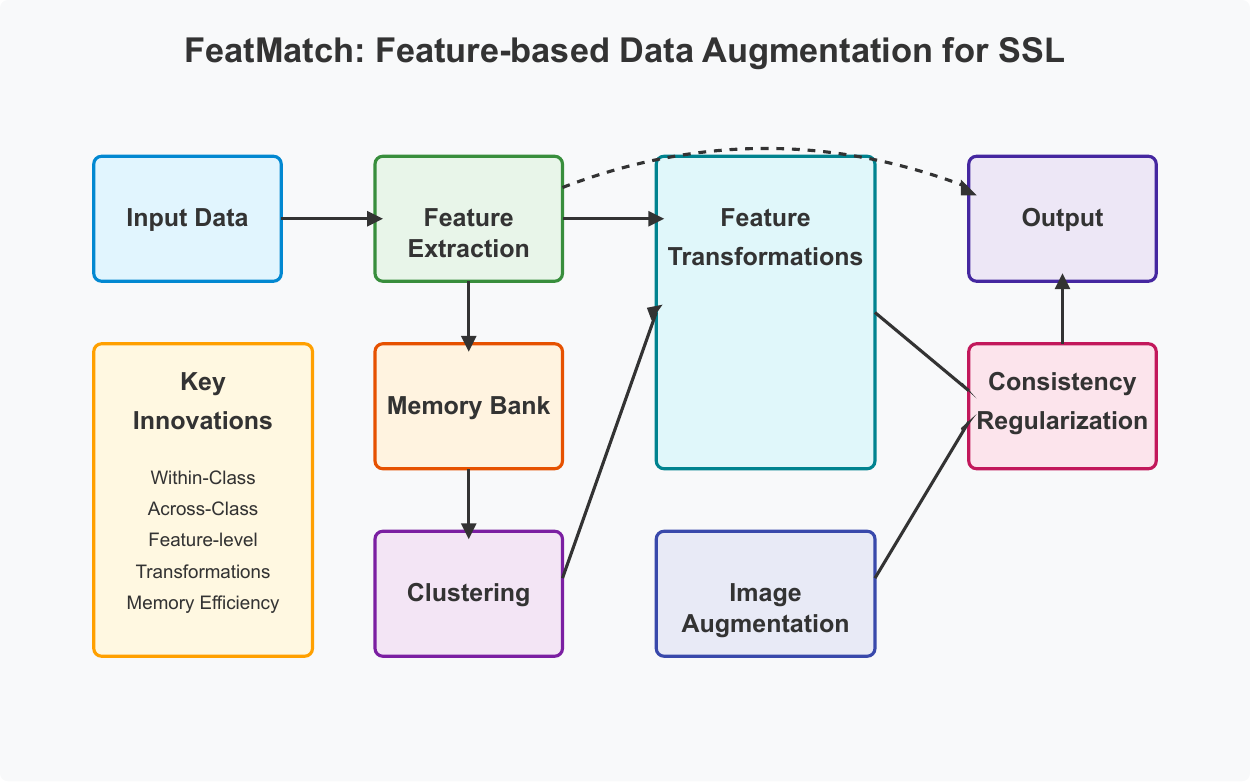}
\caption{\textit{FeatMatch} is a semi-supervised learning approach that enhances data augmentation by performing complex transformations in feature space using within-class and across-class prototypical representations.}
\label{FeatMatch}
\end{figure}

\newpage
2-2-a.\textit{ FeatMatch} \cite{kuo2020featmatch} is a novel learned feature-based refinement and augmentation method to produce a varied set of complex transformations, as shown in \cref{FeatMatch}. It also can utilize information from both within-class and across-class prototypical representations.

\begin{figure}[H]
\centering
\includegraphics[width=\textwidth]{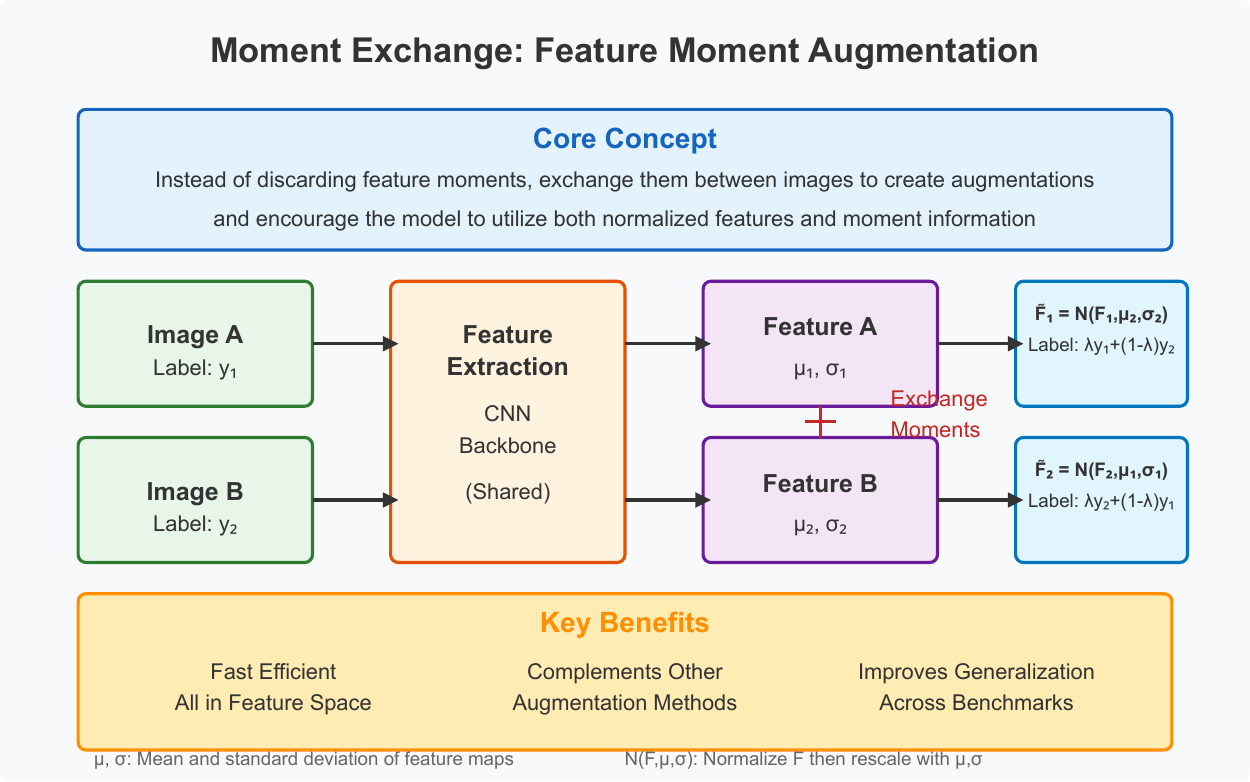}
\caption{\textit{Moment Exchange} is a novel data augmentation technique that improves recognition models by swapping feature statistics (mean and standard deviation) between images while interpolating their labels.}
\label{Moment Exchange}
\end{figure}

\newpage
2-2-b.\textit{ Moment Exchange} \cite{li2021feature} is an implicit data augmentation method that leverages the moments (mean and standard deviation) of latent features, traditionally discarded in recognition models, to enhance training by exchanging and interpolating them between training samples, as shown in \cref{Moment Exchange}. This approach improves generalization performance across various recognition benchmarks and can be seamlessly combined with existing augmentation techniques for further gains.

\begin{figure}[H]
\centering
\includegraphics[width=\textwidth]{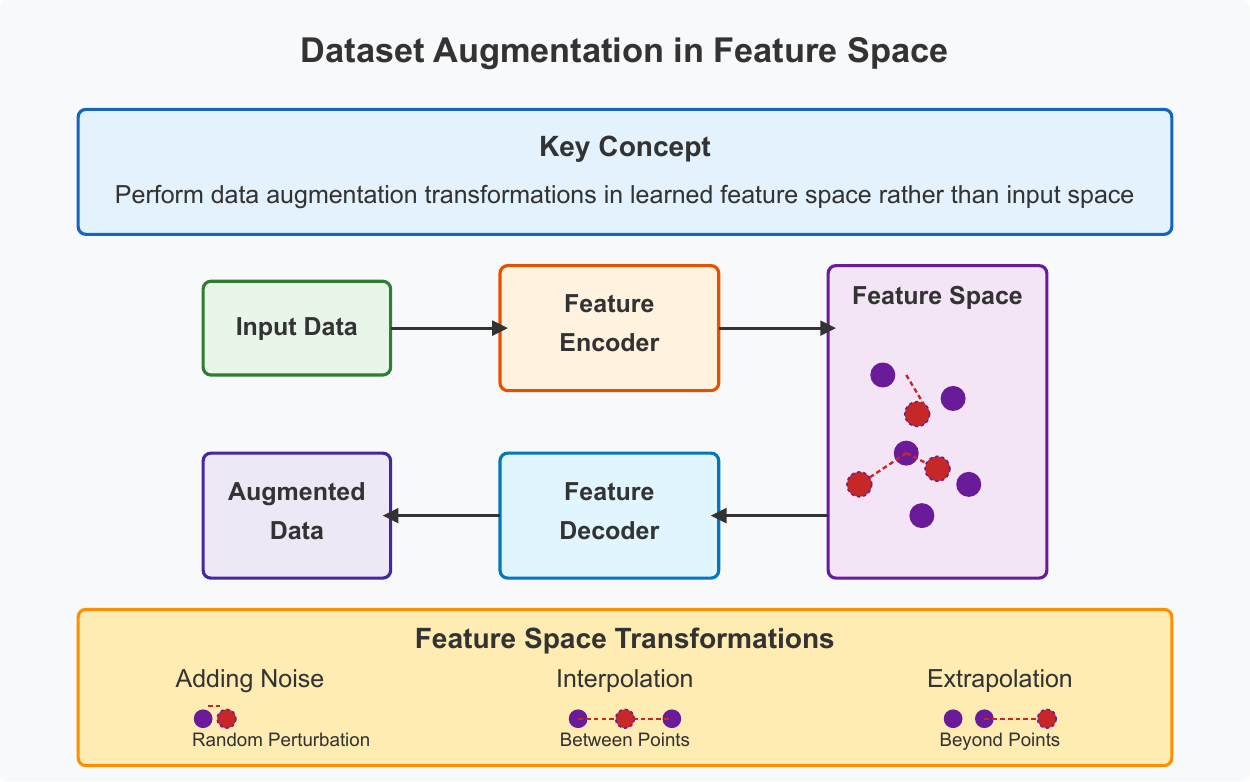}
\caption{\textit{Dataset Augmentation in Feature Space} is a domain-agnostic approach that performs simple transformations like noise addition, interpolation, and extrapolation in learned feature representations rather than input data.}
\label{Dataset Augmentation in Feature Space}
\end{figure}

\newpage
2-2-c.\textit{ Dataset Augmentation in Feature Space} \cite{devries2017dataset} introduces a domain-agnostic approach to augment training data by applying simple transformations, such as noise addition and interpolation, directly in a learned feature space rather than the input space, as shown in \cref{Dataset Augmentation in Feature Space}. This method is effective for both static and sequential data, leveraging unsupervised representation learning to enhance generalization across diverse tasks.

\begin{figure}[H]
\centering
\includegraphics[width=\textwidth]{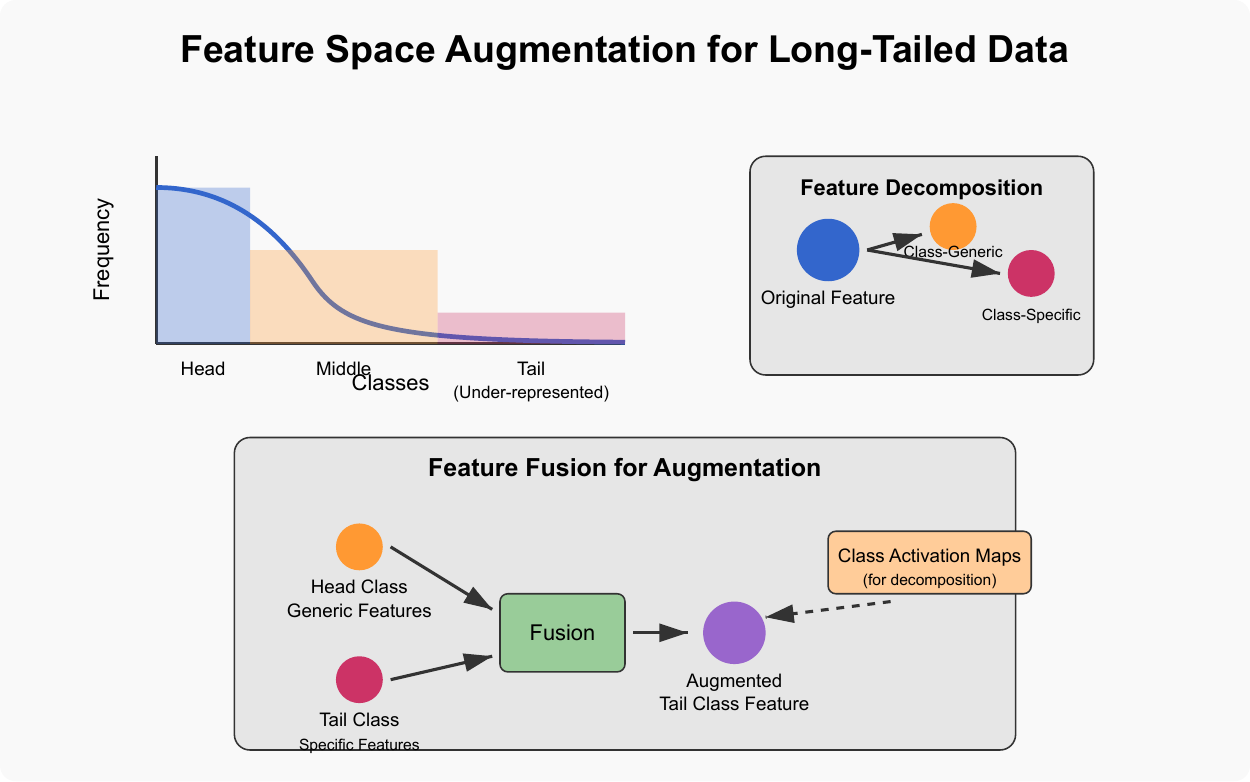}
\caption{\textit{Feature Space Augmentation for Long-Tailed Data} decomposes features into class-generic and class-specific components, then generates synthetic samples by fusing class-generic features from data-rich classes with class-specific features from under-represented classes.}
\label{Feature Space Augmentation for Long-Tailed Data}
\end{figure}

\newpage
2-2-d.\textit{ Feature Space Augmentation for Long-Tailed Data} \cite{chu2020feature} addresses class imbalance by generating novel samples for under-represented classes in the feature space, leveraging class-generic and class-specific components extracted via class activation maps, as shown in \cref{Feature Space Augmentation for Long-Tailed Data}. This approach achieves state-of-the-art performance across multiple long-tailed datasets, effectively enhancing representation for minority classes.

\begin{figure}[H]
\centering
\includegraphics[width=\textwidth]{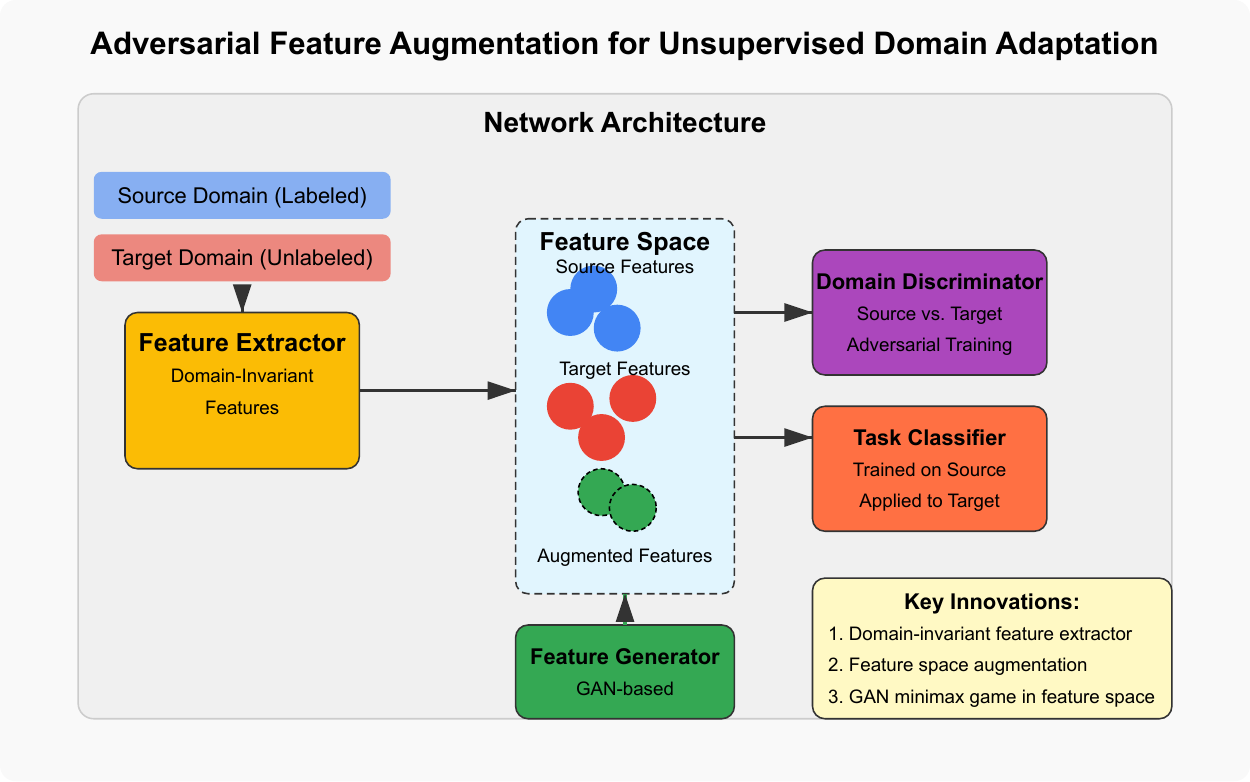}
\caption{\textit{Adversarial Feature Augmentation for Unsupervised Domain Adaptation} achieves unsupervised domain adaptation by combining a domain-invariant feature extractor with GAN-based feature space augmentation to bridge the gap between labelled source and unlabeled target domains.}
\label{Adversarial Feature Augmentation for Unsupervised Domain Adaptation}
\end{figure}

\newpage
2-2-e.textit{ Adversarial Feature Augmentation for Unsupervised Domain Adaptation} \cite{volpi2018adversarial} introduces a novel approach that utilizes a GAN-based framework to perform feature augmentation, enhancing representations for under-represented classes while enforcing domain invariance, as shown in \cref{Adversarial Feature Augmentation for Unsupervised Domain Adaptation}. This method achieves superior or comparable results to state-of-the-art techniques across various unsupervised domain adaptation benchmarks.

\begin{figure}[H]
\centering
\includegraphics[width=\textwidth]{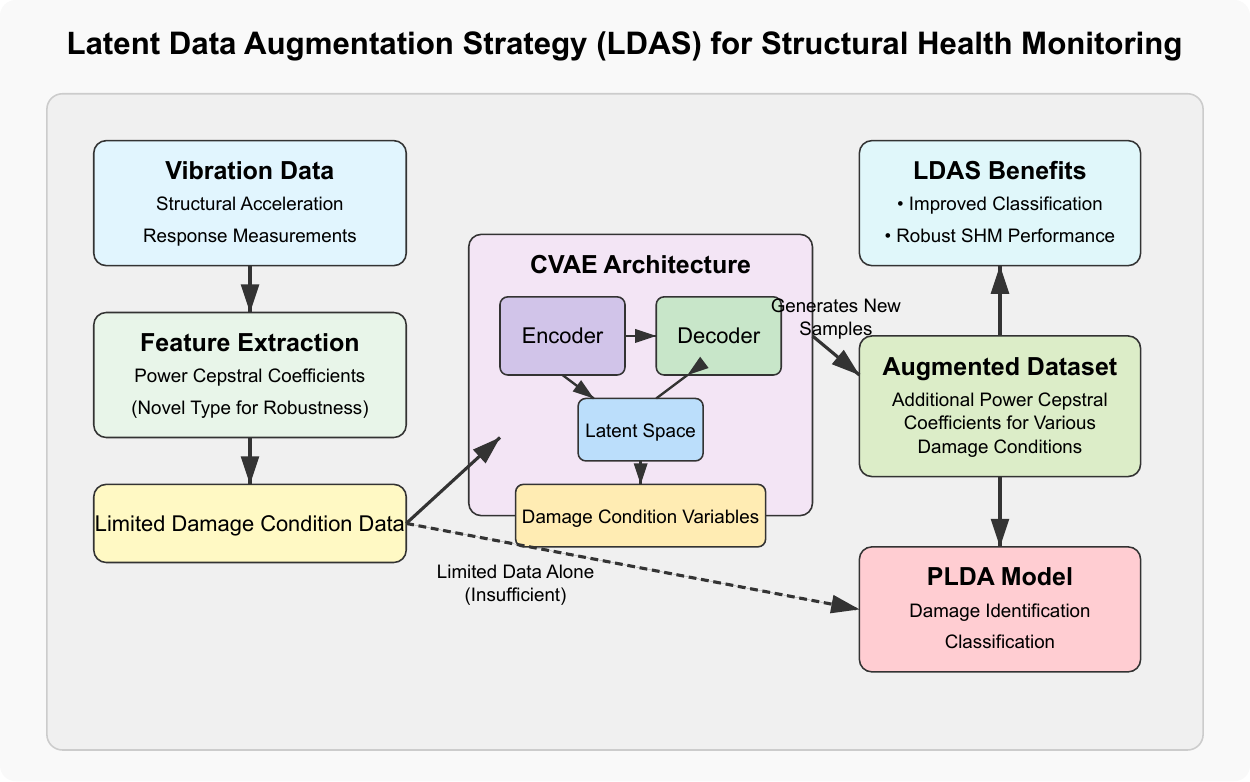}
\caption{\textit{LDAS(Latent Data Augmentation Strategy)} uses a conditional variational autoencoder to generate synthetic power cepstral coefficients representing various structural damage conditions, enabling robust damage classification despite limited real-world training data. }
\label{LDAS}
\end{figure}

\newpage
2-2-f.\textit{ LDAS(Latent Data Augmentation Strategy)} \cite{li2023machine} introduces a novel approach for generating additional training data in structural health monitoring (SHM) by leveraging a conditional variational autoencoder (CVAE) to model and augment the statistical distributions of power cepstral coefficients under various damage conditions, as shown in \cref{LDAS}. This method enhances the performance and robustness of damage classification tasks, as demonstrated through numerical simulations and experimental validations.

\begin{figure}[H]
\centering
\includegraphics[width=\textwidth]{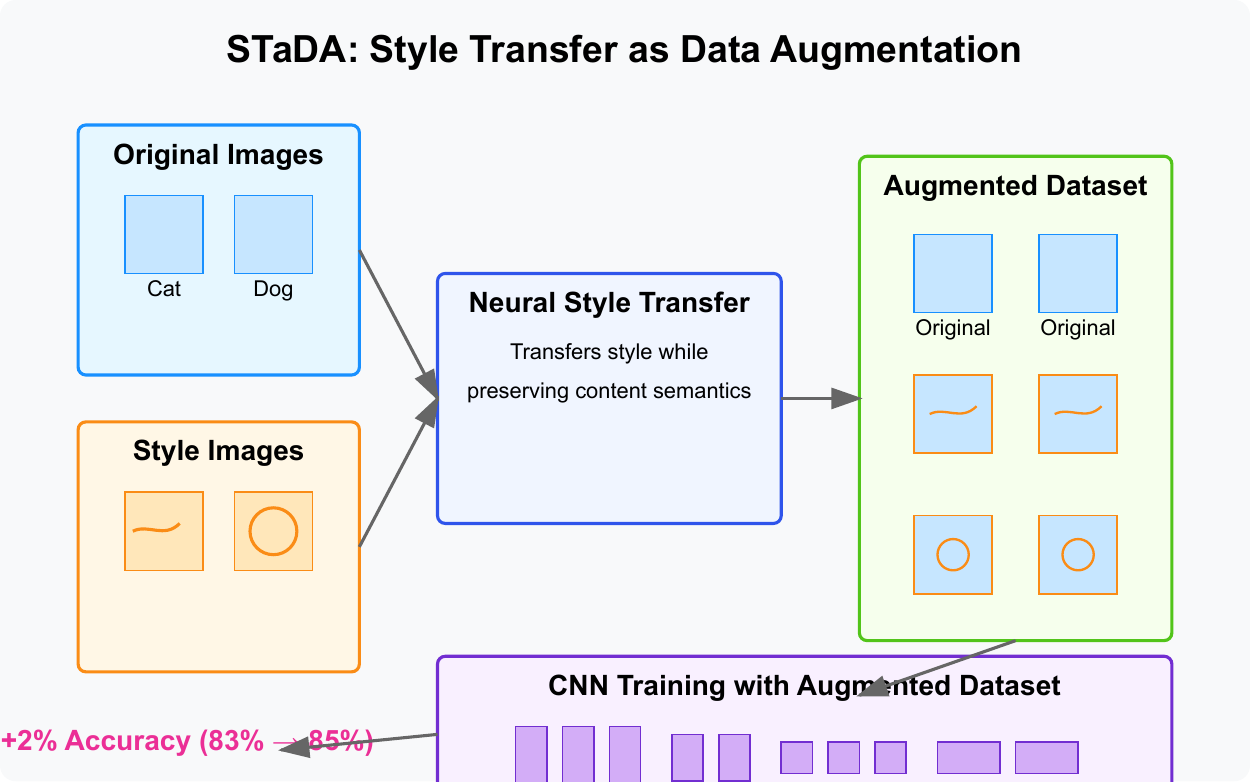}
\caption{\textit{STaDA(Style Transfer as Data Augmentation)} improves image classification accuracy by applying neural style transfer techniques to expand training datasets while preserving semantic content.}
\label{STaDA}
\end{figure}

\newpage
2-2-g.\textit{ STaDA(Style Transfer as Data Augmentation)} \cite{zheng2019stada} leverages neural style transfer techniques to enrich training datasets by applying artistic styles to images while preserving their semantic content, enhancing variation for image classification tasks, as shown in \cref{STaDA}. Experimental results on Caltech 101 and Caltech 256 demonstrate that STaDA, combined with traditional augmentation methods, improves classification accuracy and reduces reliance on large labelled datasets.

\begin{figure}[H]
\centering
\includegraphics[width=\textwidth]{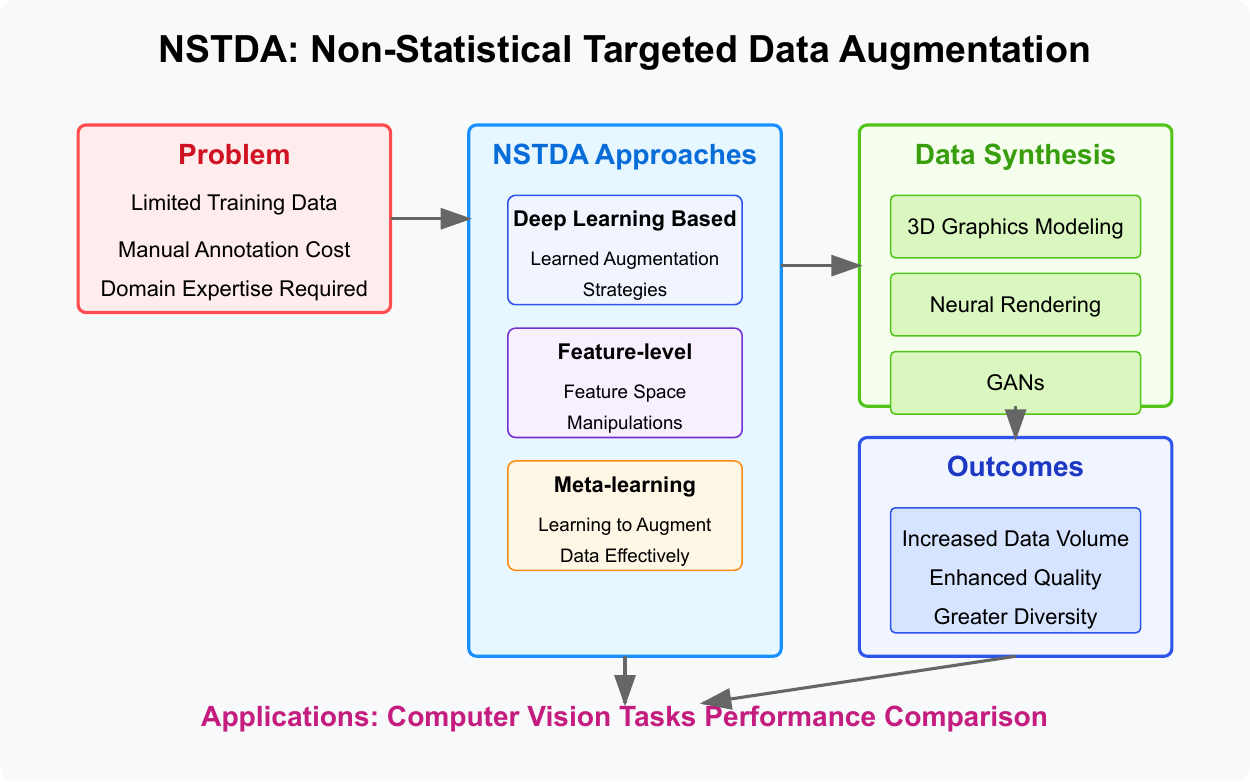}
\caption{\textit{NSTDA(Non-Statistical Targeted Data Augmentation)} improves machine learning models by employing targeted data augmentation and synthesis techniques to overcome limited training data challenges in computer vision applications.}
\label{NSTDA}
\end{figure}

\newpage
2-2-h.\textit{ NSTDA(Non-Statistical Targeted Data Augmentation)} \cite{mumuni2022data} addresses the challenge of insufficient training data by employing advanced augmentation techniques that enhance the volume, quality, and diversity of datasets in a task-specific manner, as shown in \cref{NSTDA}. This approach integrates modern methods such as neural rendering, 3D graphics modelling, and generative adversarial networks, demonstrating improved performance across various computer vision tasks and datasets.

\begin{figure}[H]
\centering
\includegraphics[width=\textwidth]{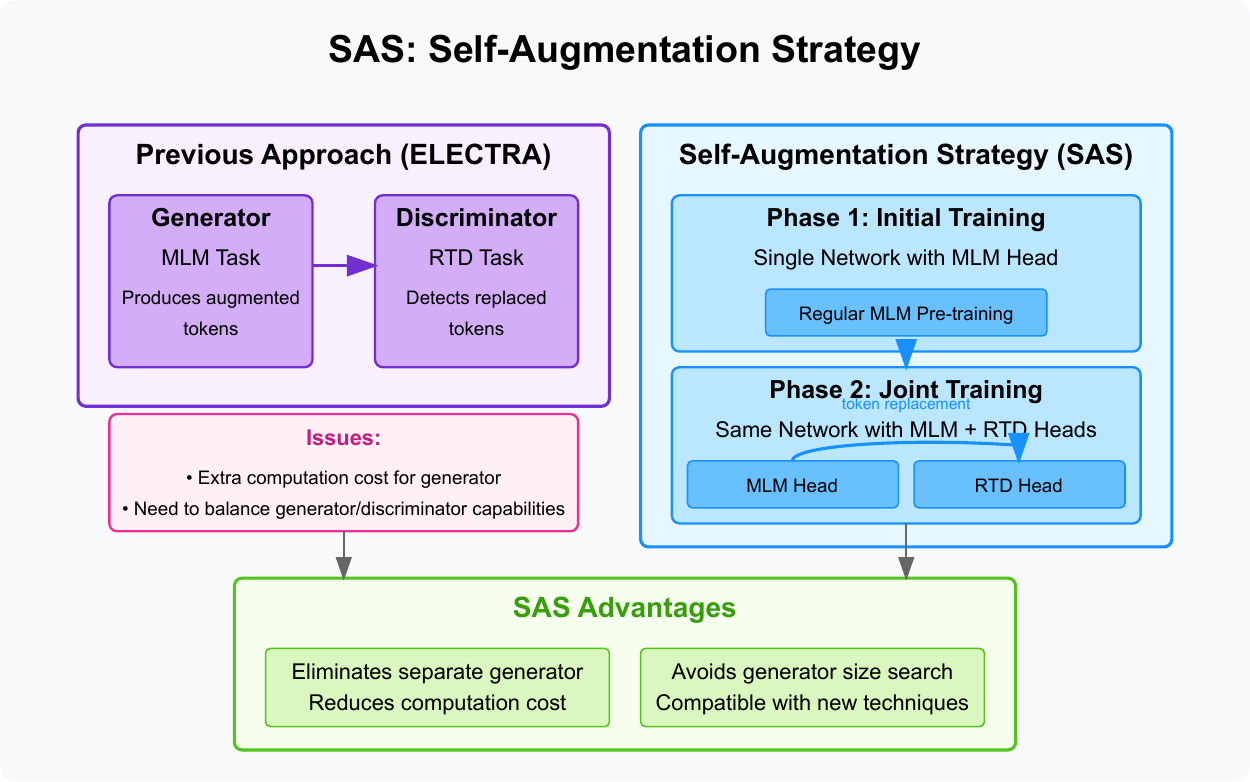}
\caption{\textit{SAS(Self-Augmentation Strategy)} improves language model pre-training by using a single network that evolves from MLM-only to joint MLM-RTD training, eliminating the separate generator network needed in previous approaches like ELECTRA.}
\label{SAS}
\end{figure}

\newpage
2-2-i.\textit{ SAS(Self-Augmentation Strategy)} \cite{xu2022sas} is a novel approach for pre-training language models that utilizes a single network to perform both regular pre-training and contextualized data augmentation, eliminating the need for a separate generator and reducing computational complexity, as shown in \cref{SAS}. By jointly optimizing Masked Language Modeling (MLM) and Replaced Token Detection (RTD) tasks, SAS simplifies the training paradigm, addresses the challenge of balancing generator-discriminator capacities, and achieves state-of-the-art performance on GLUE benchmarks with comparable or reduced computational costs.

\begin{figure}[H]
\centering
\includegraphics[width=\textwidth]{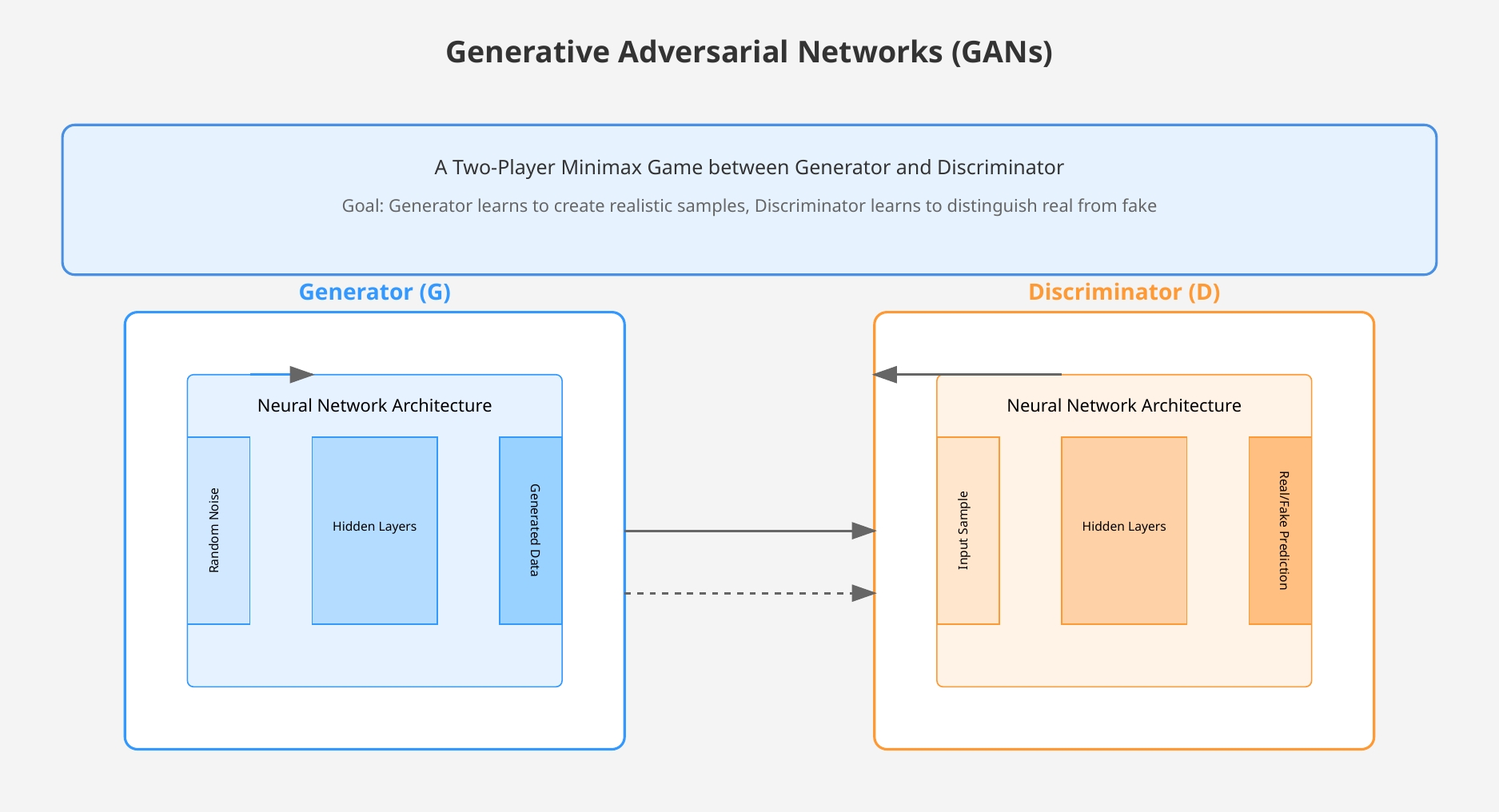}
\caption{\textit{Generative adversarial networks(GANs)} are a machine learning framework where two neural networks—a generator and a discriminator—compete to create increasingly realistic synthetic data.}
\label{GANs}
\end{figure}

\newpage
2-3-a.\textit{ Generative adversarial networks(GANs)} \cite{goodfellow2014generative} introduce a novel framework for estimating generative models by simultaneously training a generative model G, which captures the data distribution, and a discriminative model D, which distinguishes between real and generated samples, as shown in \cref{GANs}. This adversarial training process, formulated as a minimax two-player game, enables the generative model to approximate the training data distribution and the discriminative model to reach an optimal state, with the entire system trainable via backpropagation without requiring Markov chains or approximate inference networks.

\begin{figure}[H]
\centering
\includegraphics[width=\textwidth]{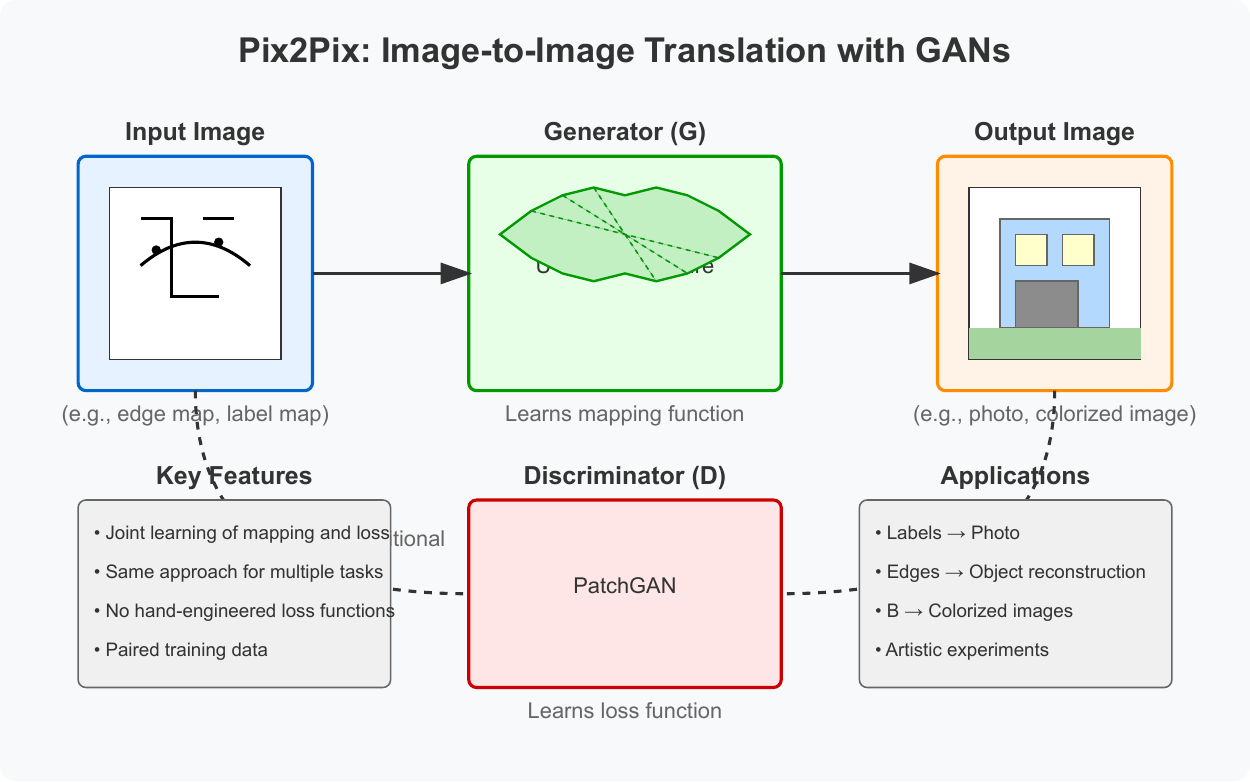}
\caption{\textit{Pix2Pix} is a conditional adversarial network framework for image-to-image translation that simultaneously learns both the mapping function and loss function, enabling diverse applications from edge maps to photos, label maps to scenes, and black/white to colourized images.}
\label{Pix2Pix}
\end{figure}

\newpage
2-3-b.\textit{ Pix2Pix} \cite{isola2017image} introduces conditional adversarial networks as a general-purpose framework for image-to-image translation, enabling the model to simultaneously learn both the mapping from input to output images and the loss function required to train this mapping, as shown in \cref{Pix2Pix}. This approach demonstrates versatility across diverse tasks, such as photo synthesis from label maps, object reconstruction from edge maps, and image colourization, eliminating the need for task-specific loss function engineering. 

\begin{figure}[H]
\centering
\includegraphics[width=\textwidth]{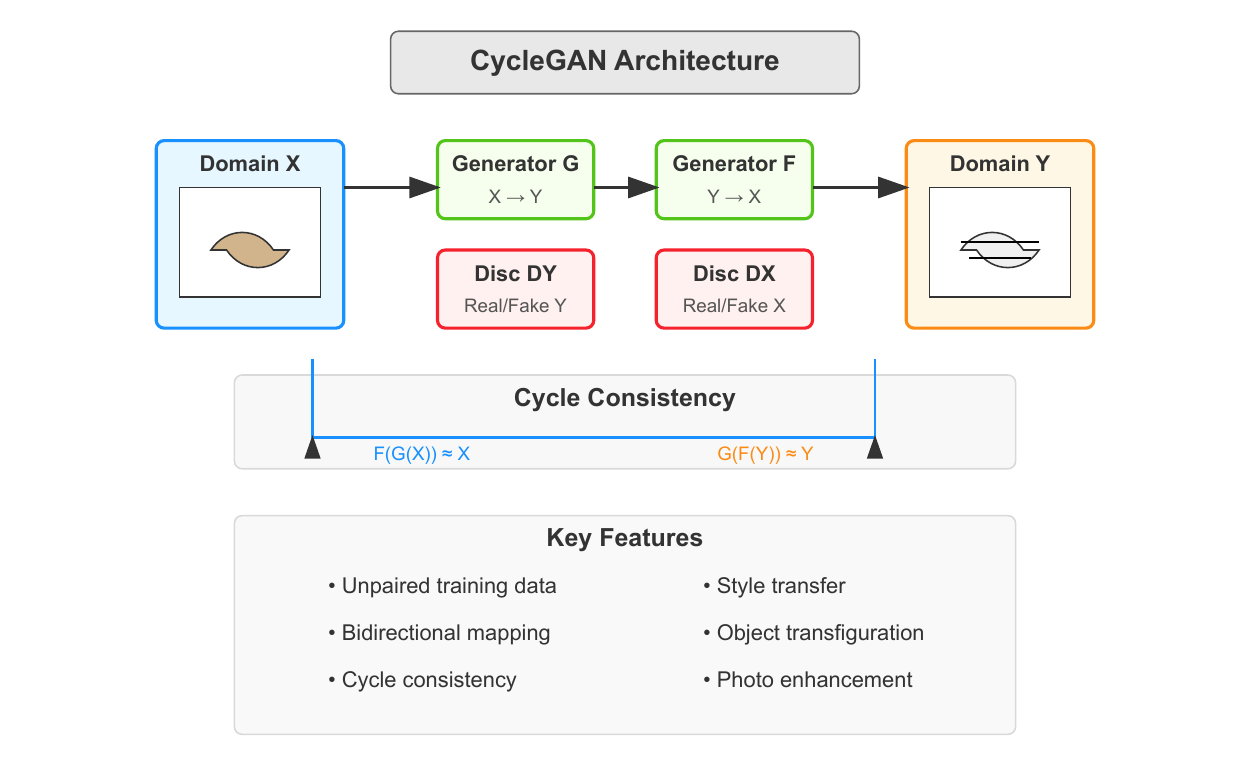}
\caption{\textit{CycleGAN} enables unpaired image-to-image translation through dual generators and cycle consistency, allowing applications like style transfer without requiring matched image pairs.}
\label{CycleGAN}
\end{figure}

\newpage
2-3-c.\textit{ CycleGAN} \cite{zhu2017unpaired} is a framework for image-to-image translation that tackles the challenge of learning mappings between a source domain and a target domain without paired training data, as shown in \cref{CycleGAN}. It combines adversarial loss to align image distributions with cycle consistency loss to ensure that translating an image to the target domain and back to the source domain reconstructs the original image. This approach achieves impressive results in tasks such as style transfer, object transfiguration, and photo enhancement, even in the absence of paired datasets.

\begin{figure}[H]
\centering
\includegraphics[width=\textwidth]{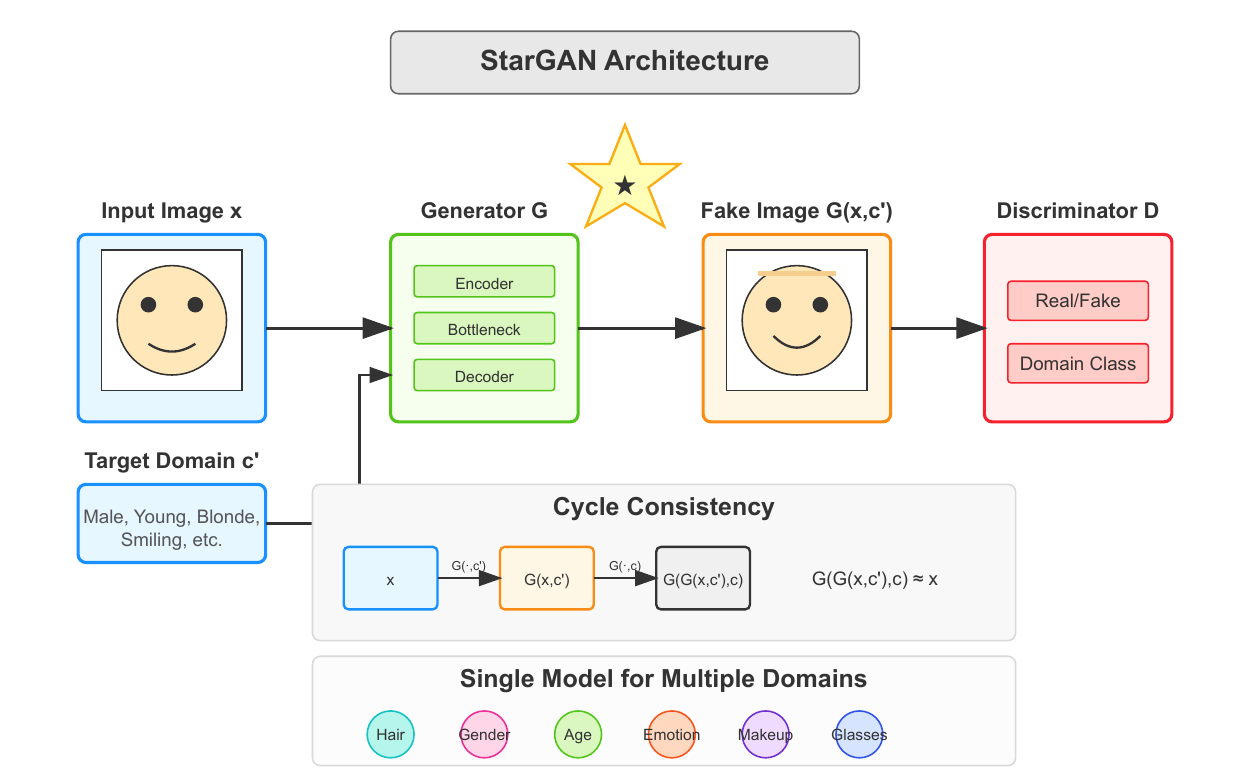}
\caption{\textit{StarGAN} achieves multi-domain image translation with a single unified model through domain labels and cycle consistency, enabling flexible facial attribute manipulation using one generator-discriminator network.}
\label{StarGAN}
\end{figure}

\newpage
2-3-d.\textit{ StarGAN} \cite{choi2018stargan} is a scalable framework for image-to-image translation that addresses the limitations of existing methods, which require separate models for each pair of image domains, as shown in \cref{StarGAN}. By utilizing a unified model architecture, StarGAN enables simultaneous training across multiple datasets and domains, achieving high-quality translations and flexible mappings to any desired target domain, as demonstrated in tasks like facial attribute transfer and expression synthesis.

\begin{figure}[H]
\centering
\includegraphics[width=\textwidth]{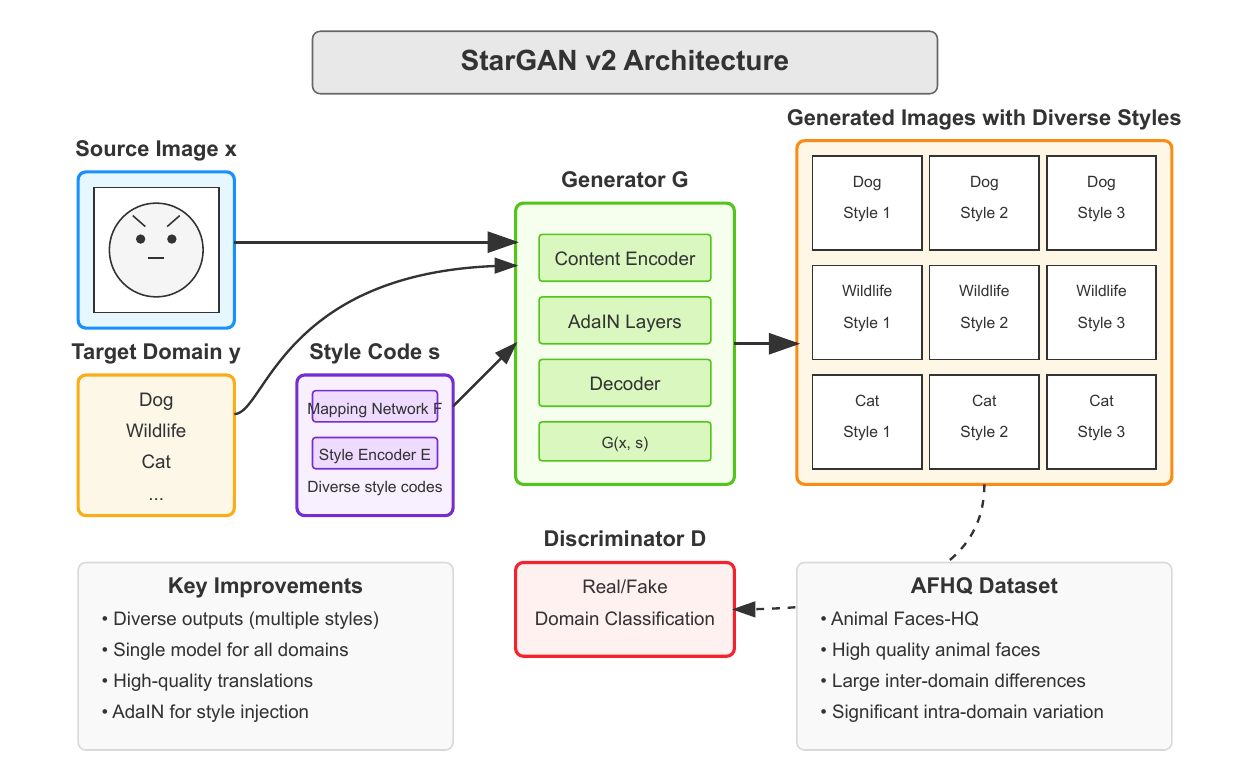}
\caption{\textit{StarGAN v2} advances image-to-image translation by combining a style-based generator with a unified multi-domain framework to produce diverse, high-quality outputs across domains using a single model.}
\label{StarGAN v2}
\end{figure}

\newpage
2-3-e.\textit{ StarGAN v2} \cite{choi2020stargan} is a unified framework for image-to-image translation that addresses both the diversity of generated images and scalability across multiple domains, overcoming the limitations of existing methods, as shown in \cref{StarGAN v2}. Through experiments on CelebA-HQ and the newly introduced AFHQ dataset, StarGAN v2 demonstrates superior performance in visual quality, diversity, and scalability, setting a new benchmark for multi-domain image translation.

\begin{figure}[H]
\centering
\includegraphics[width=\textwidth]{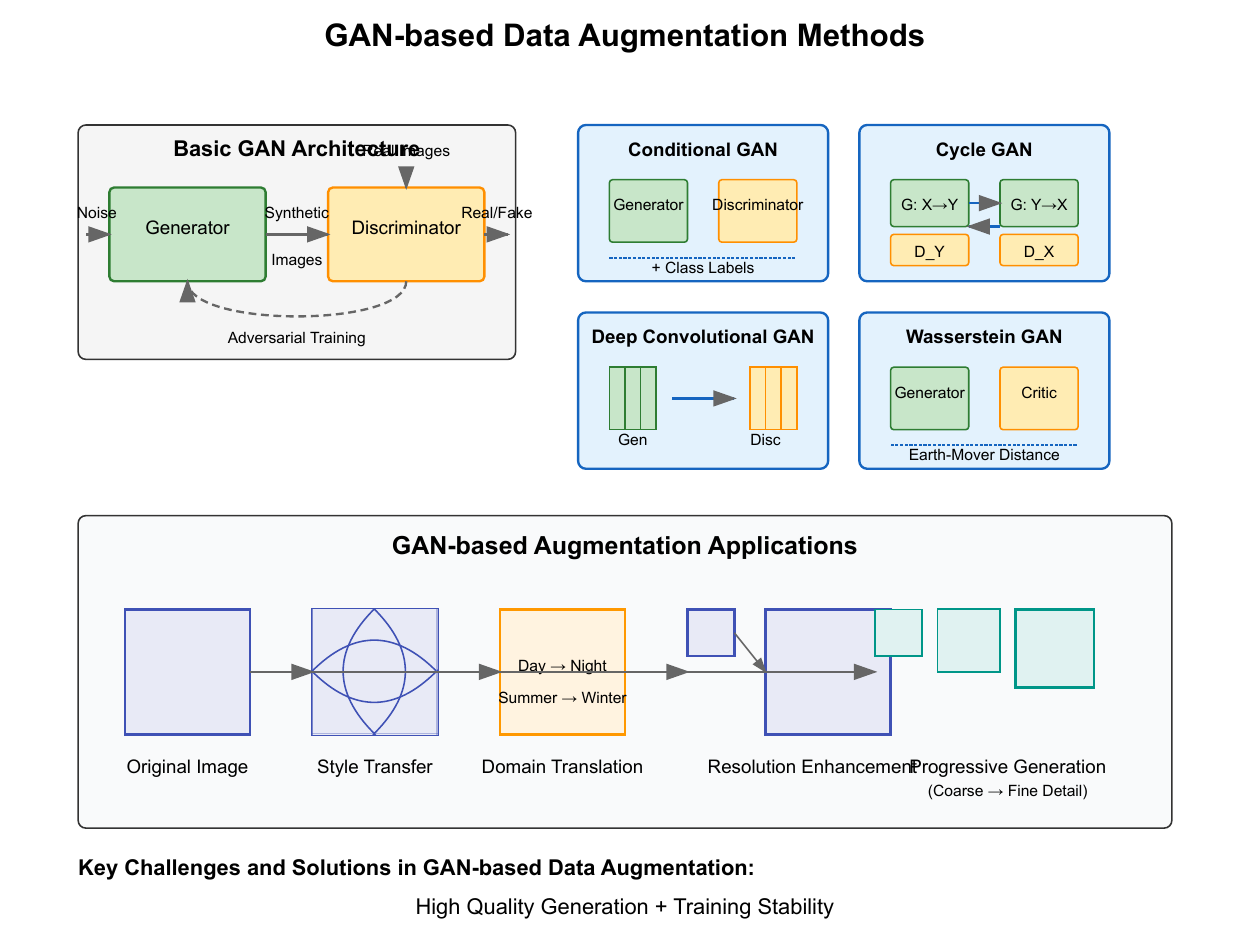}
\caption{\textit{GAN based} augmentation uses generative adversarial networks to create synthetic training images through various specialized architectures like conditional, deep convolutional, cycle, and Wasserstein GANs that address quality and stability challenges.}
\label{GAN based}
\end{figure}

\newpage
2-3-f.\textit{ GAN based} \cite{goceri2023medical} augmentation leverages Generative Adversarial Networks (GANs) to create realistic synthetic images by training a generator and discriminator in tandem, enabling diverse and high-quality data augmentation, as shown in \cref{GAN based}. Advanced GAN variants, such as Conditional GAN, Cycle GAN, and Style GAN, enhance image generation by incorporating class labels, domain translation, and fine-grained detail synthesis, addressing challenges such as quality and training stability.

\begin{figure}[H]
\centering
\includegraphics[width=\textwidth]{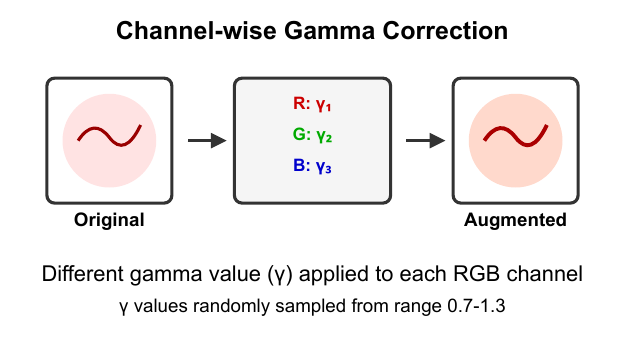}
\caption{\textit{Channel-wise gamma correction} is independent random gamma transformations applied to each RGB channel to improve retinal vessel segmentation robustness.}
\label{Channel-wise}
\end{figure}

\newpage
2-3-g.\textit{ Channel-wise gamma correction} \cite{sun2021robust} is a data augmentation technique that applies random gamma correction independently to each colour channel of a fundus image, enhancing variability in colour intensity, as shown in \cref{Channel-wise}. This method helps models learn more robust and invariant features, improving performance against global disturbances in retinal vessel segmentation tasks.

\begin{figure}[H]
\centering
\includegraphics[width=\textwidth]{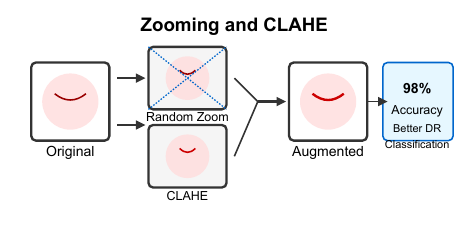}
\caption{\textit{Zooming and CLAHE} is a data augmentation approach combining random zoom transformations with contrast enhancement to achieve 98\% accuracy in diabetic retinopathy classification.}
\label{Zooming and CLAHE}
\end{figure}

\newpage
2-3-h.\textit{ Zooming and CLAHE} \cite{agustin2020implementation} augmentation are effective techniques for enhancing retinal fundus image analysis in diabetic retinopathy classification, as shown in \cref{Zooming and CLAHE}. While random zooming introduces variability by simulating different focal levels, CLAHE enhances image contrast, together improving model accuracy, sensitivity, and specificity in deep learning applications.

\begin{figure}[H]
\centering
\includegraphics[width=\textwidth]{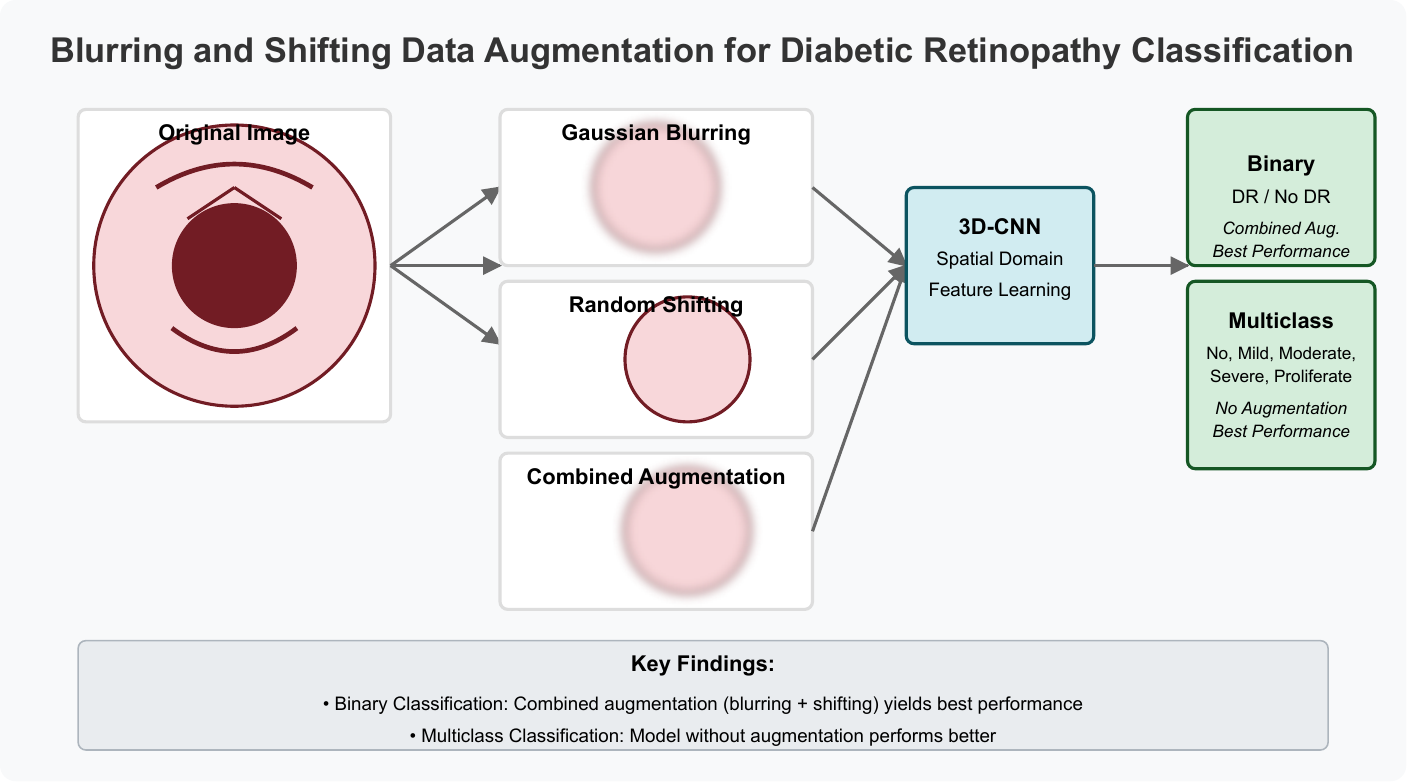}
\caption{\textit{Blurring and shifting} data augmentation techniques for diabetic retinopathy classification showing differential performance in binary versus multiclass 3D-CNN models.}
\label{Blurring and shifting}
\end{figure}

\newpage
2-3-i.\textit{ Blurring and shifting} augmentation are data enhancement techniques used to improve the robustness of deep learning models in diabetic retinopathy classification, as shown in \cref{Blurring and shifting}. Random weak Gaussian blurring reduces image details to simulate real-world noise, while random shifting alters spatial positioning, collectively aiding in training models to handle spatial and visual variability.\cite{tufail2021diagnosis} 

\begin{figure}[H]
\centering
\includegraphics[width=\textwidth]{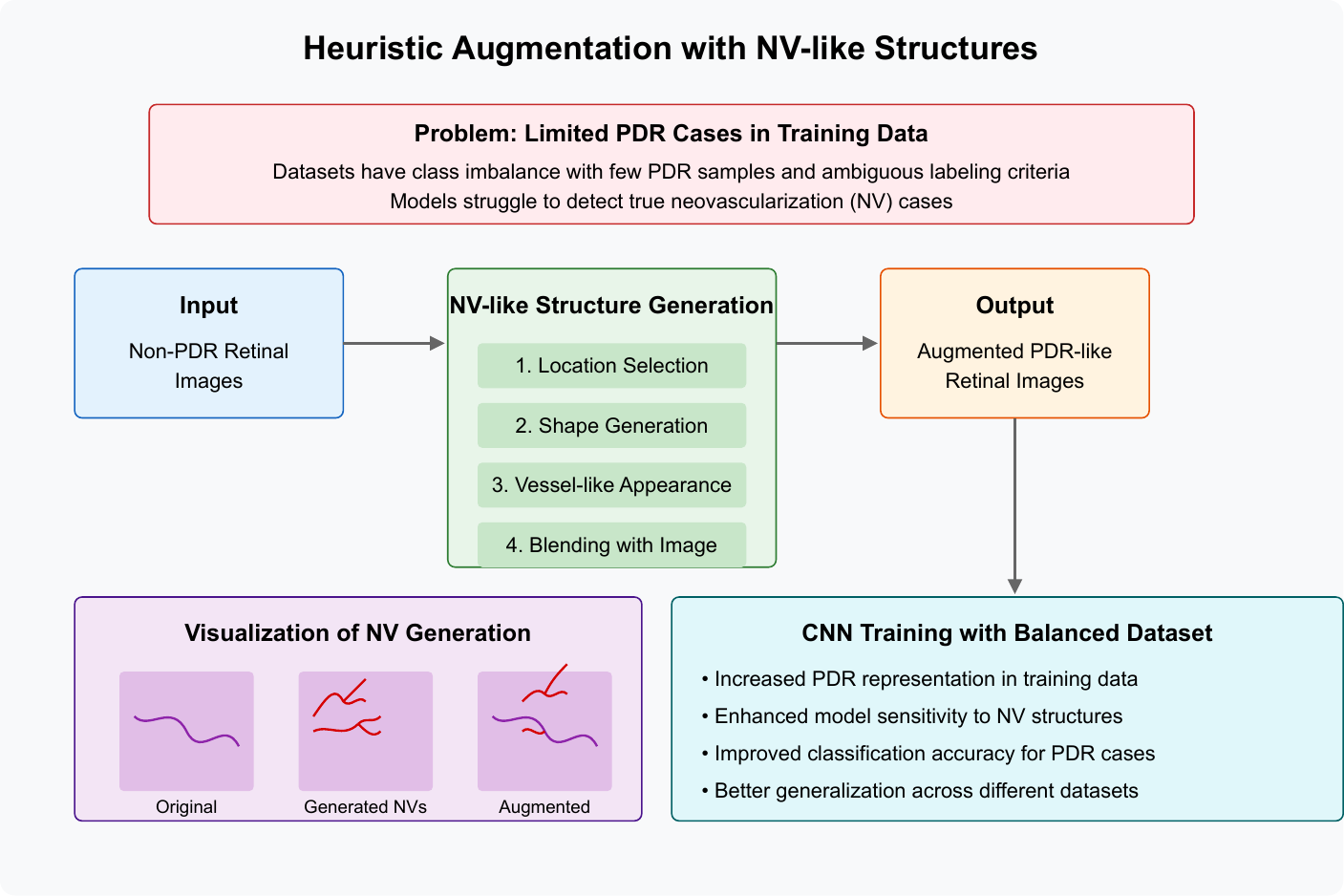}
\caption{\textit{Heuristic augmentation with NV-like structures} is a data augmentation approach that synthetically generates neovascularization patterns on retinal images to address the underrepresentation of PDR cases in training datasets.}
\label{NV-like}
\end{figure}

\newpage
2-3-j.\textit{ Heuristic augmentation with NV-like structures} \cite{araujo2020data} is a data augmentation technique designed to address the scarcity of proliferative diabetic retinopathy (PDR) cases in retinal image datasets, as shown in \cref{NV-like}. By synthesizing neovessel-like structures based on their common locations and shapes, this method enhances dataset variability and improves the ability of deep-learning models to detect PDR-related features.

\begin{figure}[H]
\centering
\includegraphics[width=\textwidth]{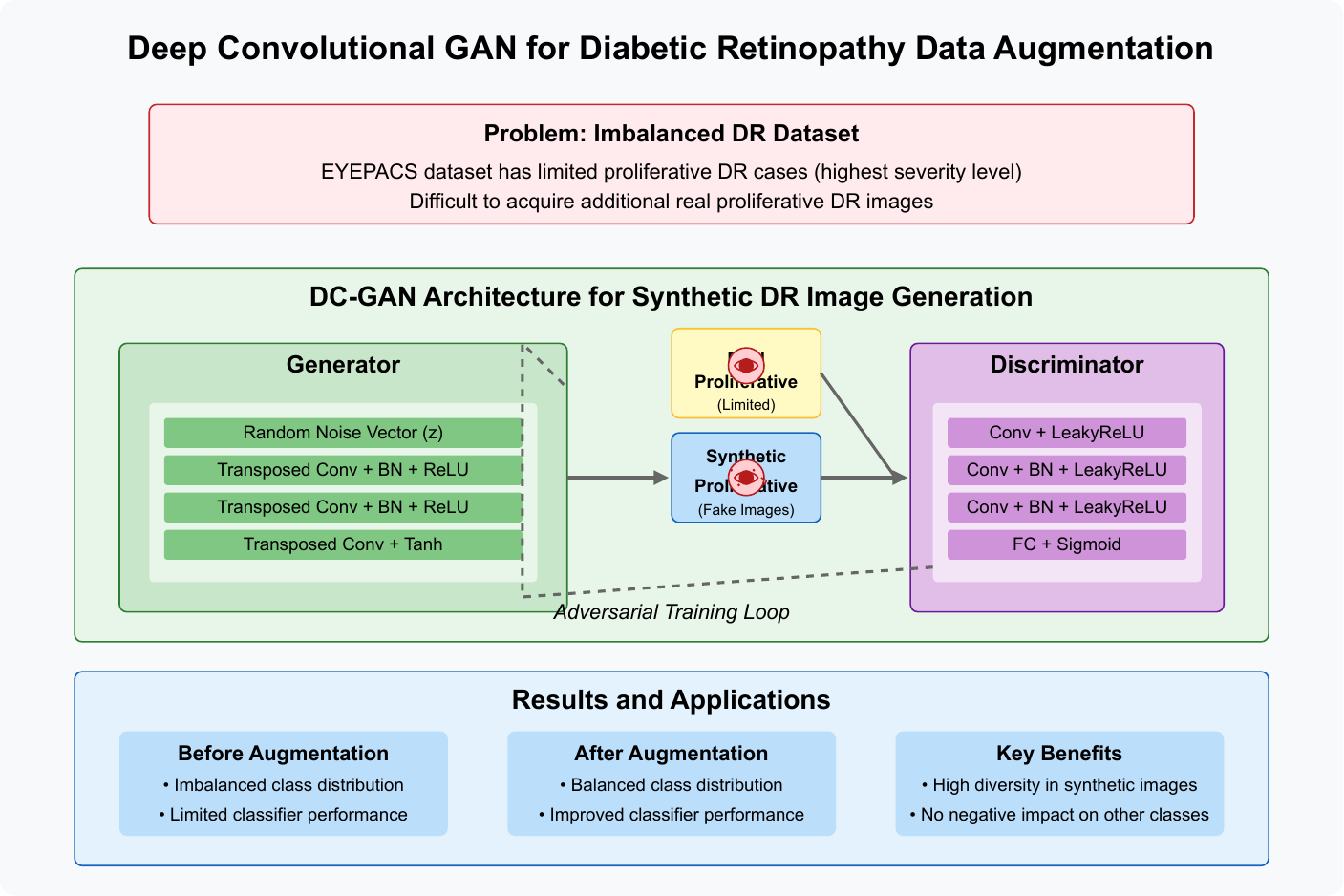}
\caption{\textit{Deep convolutional GAN} is a generative adversarial network architecture that synthesizes diverse proliferative diabetic retinopathy images to address class imbalance in training datasets, enabling improved classification performance.}
\label{Deep convolutional GAN}
\end{figure}

\newpage
2-3-k.\textit{ Deep convolutional GAN} \cite{balasubramanian2020analysis} augmentation leverages generative adversarial networks to synthesize diverse and realistic images for underrepresented classes, such as proliferative diabetic retinopathy, in imbalanced datasets, as shown in \cref{Deep convolutional GAN}. This approach enhances data availability and improves classification performance by providing high-quality synthetic samples without affecting other class distributions.

\begin{figure}[H]
\centering
\includegraphics[width=\textwidth]{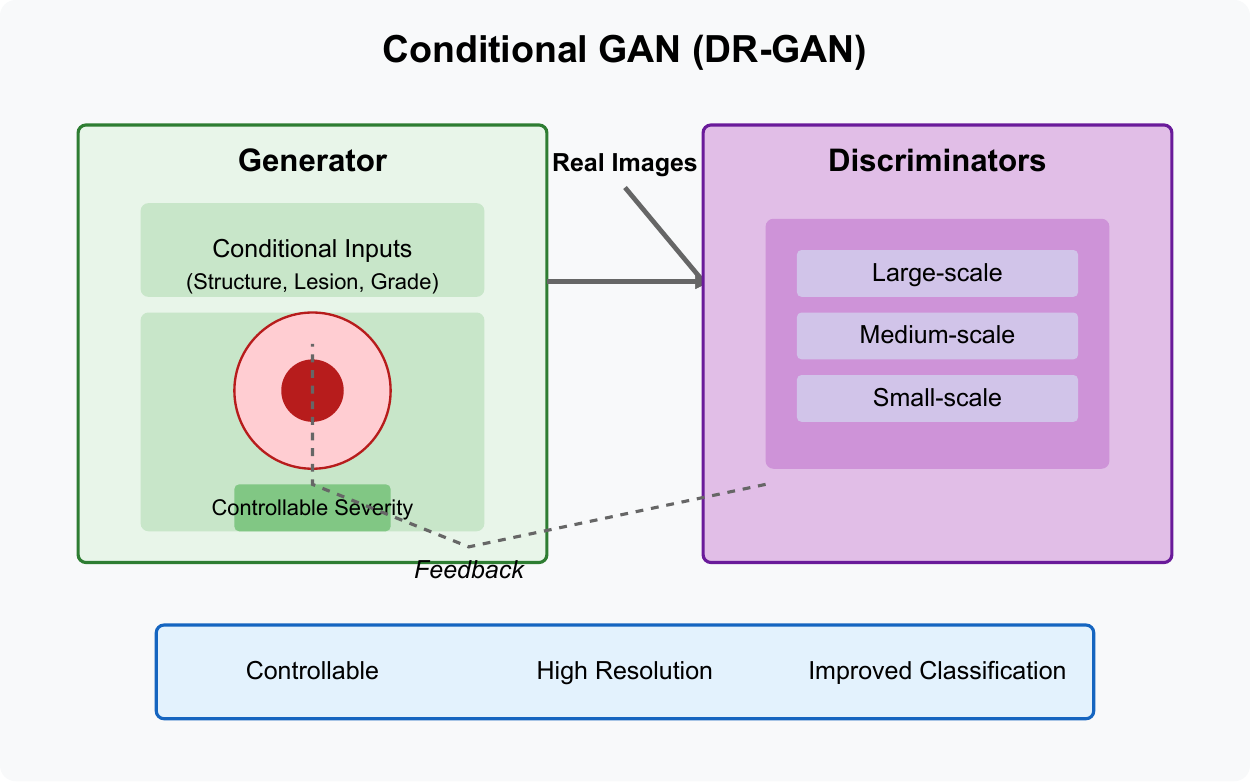}
\caption{\textit{Conditional GAN} is a generative model that synthesizes high-resolution diabetic retinopathy images with controllable severity grades by incorporating structural masks, lesion patterns, and grading vectors as input conditions.}
\label{Conditional GAN}
\end{figure}

\newpage
2-3-l.\textit{ Conditional GAN} \cite{zhou2020dr} augmentation utilizes generative adversarial networks to synthesize high-resolution fundus images conditioned on grading severity and lesion information, enabling targeted data generation for diabetic retinopathy classification, as shown in \cref{Conditional GAN}. By incorporating structural masks and adaptive grading vectors, this approach enhances data diversity and improves model performance on grading and lesion segmentation tasks.

\begin{figure}[H]
\centering
\includegraphics[width=\textwidth]{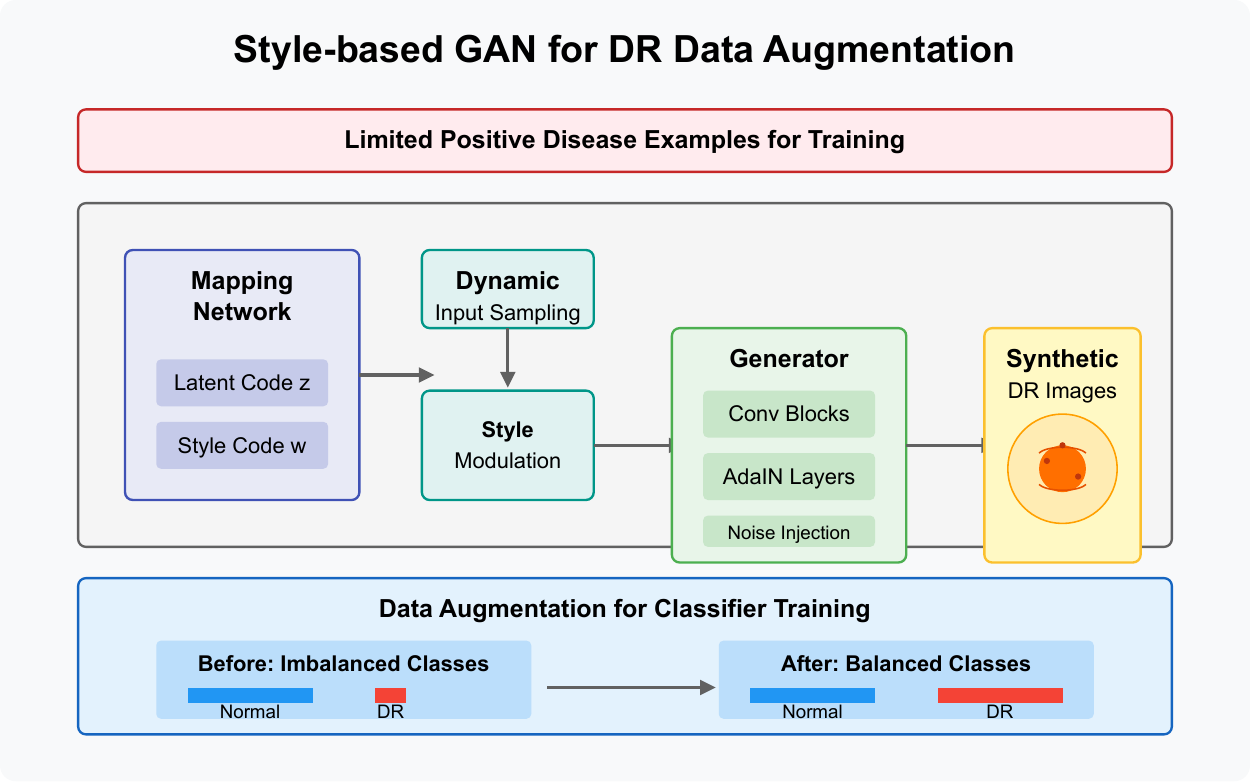}
\caption{\textit{Style based GAN} is a generative approach that leverages style modulation and dynamic input sampling to create realistic diabetic retinopathy images, addressing class imbalance by synthesizing rare disease cases for improved classifier training.}
\label{Style based GAN}
\end{figure}

\newpage
2-3-m.\textit{ Style based GAN} \cite{lim2020generative} augmentation leverages generative adversarial networks to synthesize high-quality, class-specific images by learning and transferring fine-grained style features from real images, as shown in \cref{Style based GAN}. This approach addresses data scarcity in medical imaging, particularly for rare disease conditions, and enhances classification performance in diabetic retinopathy detection tasks.

\begin{figure}[H]
\centering
\includegraphics[width=\textwidth]{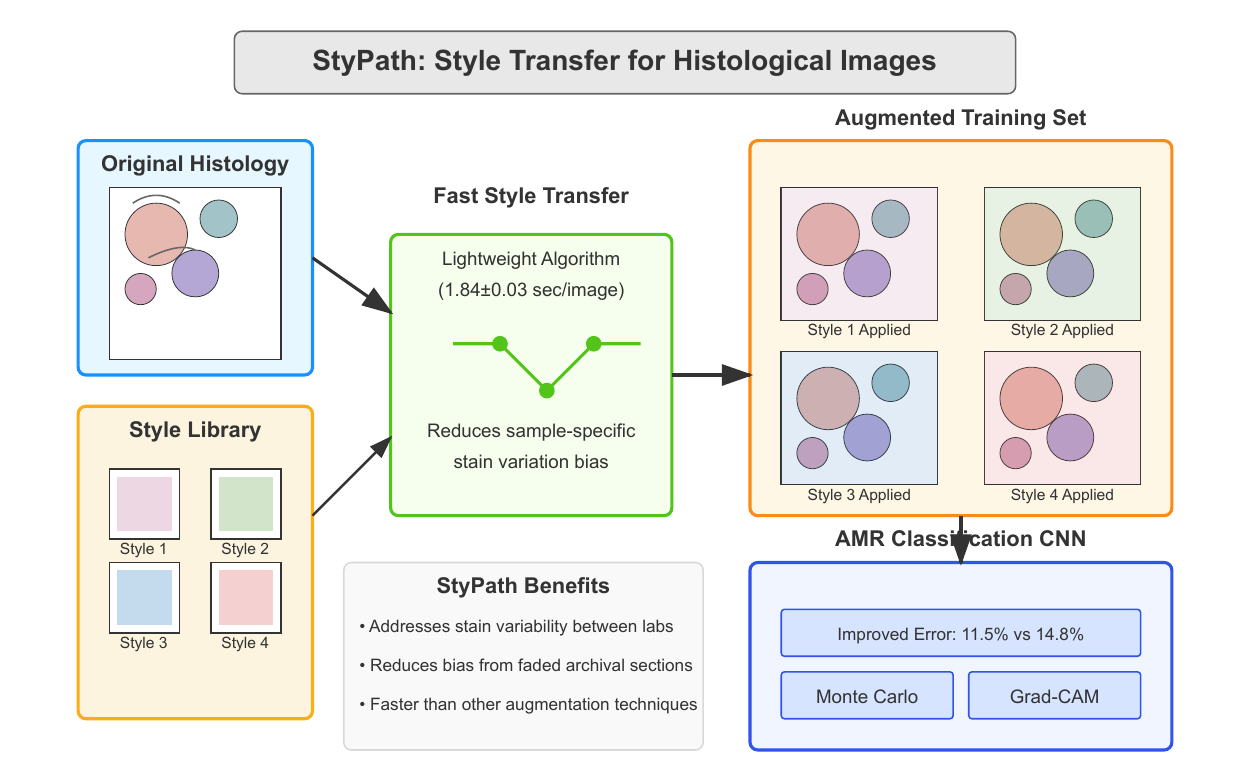}
\caption{\textit{StyPath} enhances AMR classification in kidney transplant histology through lightweight style transfer augmentation that reduces stain variation bias, improving model accuracy and generalization by simulating diverse staining conditions.}
\label{StyPath}
\end{figure}

\newpage
2-3-n.\textit{StyPath (Style Transfer Pathway)} \cite{cicalese2020stypath} is a histological data augmentation technique designed to address variability in tissue stain quality, a common challenge in kidney transplant pathology for Antibody Mediated Rejection (AMR) classification, as shown in \cref{StyPath}. By leveraging a lightweight style-transfer algorithm, StyPath reduces sample-specific bias, improves classification performance, enhances model generalization, and demonstrates faster processing compared to other augmentation methods, as validated through Bayesian performance estimates and expert qualitative analysis.

\begin{figure}[H]
\centering
\includegraphics[width=\textwidth]{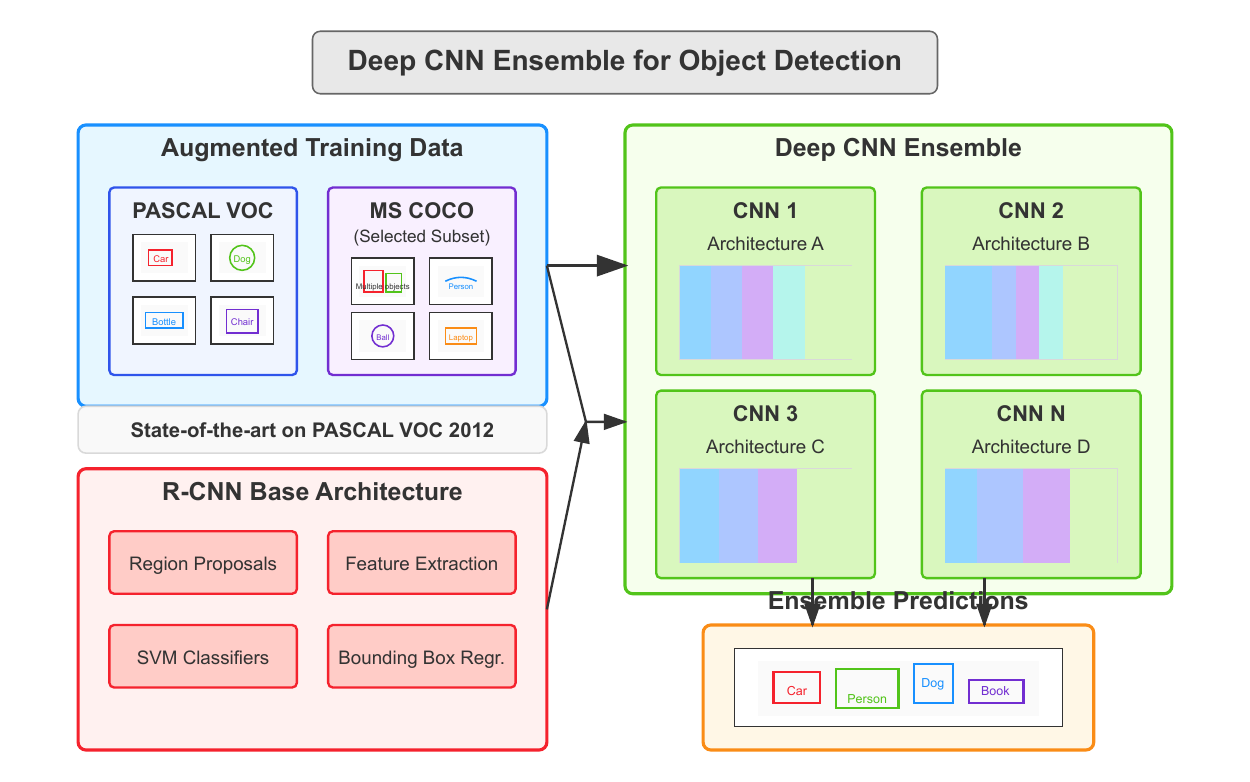}
\caption{\textit{Deep CNN Ensemble} enhances object detection performance by combining multiple complementary CNN architectures with an augmented training set that integrates selected Microsoft COCO images with PASCAL VOC data.}
\label{Deep CNN Ensemble}
\end{figure}

\newpage
2-3-o.\textit{ Deep CNN Ensemble} \cite{guo2015deep} is a variant of the R-CNN model that achieves state-of-the-art performance in object detection by combining complementary deep CNN architectures into an ensemble, as shown in \cref{Deep CNN Ensemble}. By augmenting the PASCAL VOC training set with a curated subset of Microsoft COCO images, the method significantly improves training data diversity and achieves superior results on the PASCAL VOC 2012 detection task.

\end{document}